\documentclass[review,12pt,authoryear]{elsarticle}

\usepackage{geometry}
\usepackage{fleqn}
\usepackage{graphicx}
\usepackage{caption}
\usepackage{subcaption}
\usepackage{enumerate}
\usepackage{amssymb}
\usepackage[utf8]{inputenc}
\usepackage{multicol}
\usepackage[lined,linesnumbered,boxed]{algorithm2e}
\usepackage{float}

\begin{document}
\let\today\relax
\begin{frontmatter}

\title{$\mu$-MAR: Multiplane 3D Marker based Registration for Depth-sensing Cameras}

\author[alc]{Marcelo Saval-Calvo\corref{cor2}}
\ead{msaval@dtic.ua.es}

\author[alc]{Jorge Azorín-López\corref{cor1}}
\ead{jazorin@dtic.ua.es}

\author[alc]{Andrés Fuster-Guilló\corref{cor1}}
\ead{fuster@dtic.ua.es}

\author[alc]{Higinio Mora-Mora\corref{cor1}}
\ead{hmora@dtic.ua.es}

\cortext[cor1]{Corresponding author}
\cortext[cor2]{Principal corresponding author}

\address[alc]{University of Alicante, Department of Computer Technology, \\ C/ San Vicente - Alicante s/n
03690, San Vicente del Raspeig (Alicante)}

\begin{abstract}
Many applications including object reconstruction, robot guidance, and scene mapping require the registration of multiple views from a scene to generate a complete geometric and appearance model of it. In real situations, transformations between views are unknown an it is necessary to apply expert inference to estimate them. In the last few years, the emergence of low-cost depth-sensing cameras has strengthened the research on this topic, motivating a plethora of new applications. Although they have enough resolution and accuracy for many applications, some situations may not be solved with general state-of-the-art registration methods due to the Signal-to-Noise ratio (SNR) and the resolution of the data provided. The problem of working with low SNR data, in general terms, may appear in any 3D system, then it is necessary to propose novel solutions in this aspect. In this paper, we propose a method, $\mu$-MAR, able to both coarse and fine register sets of 3D points provided by low-cost depth-sensing cameras, despite it is not restricted to these sensors, into a common coordinate system. The method is able to overcome the noisy data problem by means of using a model-based solution of multiplane registration. Specifically, it iteratively registers 3D markers composed by multiple planes extracted from points of multiple views of the scene. As the markers and the object of interest are static in the scenario, the transformations obtained for the markers are applied to the object in order to reconstruct it. Experiments have been performed using synthetic and real data. The synthetic data allows a qualitative and quantitative evaluation by means of visual inspection and Hausdorff distance respectively. The real data experiments show the performance of the proposal using data acquired by a Primesense Carmine RGB-D sensor. The method has been compared to several state-of-the-art methods. The results show the good performance of the $\mu$-MAR to register objects with high accuracy in presence of noisy data outperforming the existing methods.
\end{abstract}


\end{frontmatter}

\section{Introduction}
\label{sec:reg:intro}

Nowadays in the computer vision area, many applications make use of 3D data to reconstruct a geometric and colour model of a real object \citep{Ramos2012,Kramer2012,Newcombe2011,Izadi2011}. In order to obtain the full model, it is necessary to register or align different views of the scene into a common coordinate system. This registration is a critical task for the application because the rest of the process will rely on the quality of the alignment. In ideal situations, the viewpoints used to acquire the data from the scene are known. Hence, the transformations (i.e. rotations and translations) able to align the acquired data could be easily calculated and applied to each corresponding view in order to obtain the complete model reconstruction. However in most situations, the different views are obtained at unknown positions requiring an expert system able to infer the viewpoints and to calculate the transformations making use of the acquired data and some possible prior knowledge. Many different sensors provide 3-dimensional data, such as stereo cameras, laser, and structured light devices. The emergence of the new low-cost depth-sensing cameras, also known as RGB-D, (e.g. Primesense Carmine, Microsoft Kinect) has strengthened the research on this topic. These kind of cameras have low resolution and a considerable error in depth \citep{Khoshelham2012,Wilson2010,Smisek2011}. For example, Microsoft Kinect depth sensor provides a 320x240 matrix of real depth data and a field of view of 4,6mx2,8m at 1,5m far from the sensor. It means that the scenario is sampled in horizontal each 1,43cm and 1,17cm in vertical. This resolution is enough accurate for body parts recognition, but not for a detailed object or scenario reconstruction with small details. Many different techniques have been proposed for 3D data registration but they do not provide proper results in presence of noisy data, such as the obtained by these new low-cost cameras. Therefore, multiple views registration still remains a challenging problem in this case. In this paper, a novel method to deal with this problem is presented.

The most common algorithms for registering multiple views from a scene are based on \textit{Iterative Closest Point} (ICP)~\citep{Rusinkiewicz2001} and RANSAC \citep{Fischler1981}. ICP iteratively registers in a fine way two point clouds using the closest point matching to evaluate the correspondences between them and it is still used in recent application \citep{Costa2014, Whelan2015}. Many variants have been proposed for enhancing the result adding normal information or knowledge about other kind of constraints (e.g. borders, colour, etc.). However, this method needs an initial transformation to avoid convergence in a local minima. On the other hand, RANSAC-based methods evaluate matches (commonly estimated using 2D or 3D features) and remove the wrong correspondences between points of a view and the reference view. RANSAC is often used to extract the initial transformation for the ICP. Those methods, despite the fact that are noise-resistant, only can deal with a certain level of noise. Specifically, there exist various works dealing with RGB-D low-cost sensors. For example, \cite{Morell2014} reviewed several methods of registration using RGB-D sensor data in mapping and object reconstruction. Also, it is interesting to note the works of \cite{Han2013} and \cite{Shao2014} reviewing algorithms and applications using Kinect-like devices.

In general, a robust way of approaching the noise problem in the computer vision methods is to use models calculated from data. In the literature is common to find registration of planar models (e.g. patches) in building or urban reconstruction such as in \citep{Dold2004,Dold2006,Theiler2012}. In this case, the model is calculated from a large number of points being the noise/data ratio very low in the whole scene. Moreover, the accuracy required is in order of centimetres. This accuracy is far lower than the required for reconstructing small objects. For other purposes, plane registration is common in robotics and augmented reality to estimate the position of the camera in the scenario. For example, walls, floor and roof are used to extract the plane models that define them in \citep{Pathak2009,Lee2012,Uematsu2009}. Those planes are registered along the movement of robots to estimate robot position, but not in order to faithfully reconstruct the scenario. In consequence, the accuracy could be a non critical requirement. Xiao et al.~\citep{Xiao2012,Xiao2013} proposed the alignment of planar patches for robot mapping using two initial non-parallel pair of planes. After the translation is obtained using a third pair of planes non-parallel to both of the initial planes. They evaluate in each step if the rotation and translation are plausible using the kinematic of the robot. 

The use of planes as a model of raw data also could be an important solution for reconstructing entire scenes due to planes could be found almost anywhere. In consequence, they can be found in many objects and in almost every scenes. In \citep{Pathak2009,Pathak2010b}, a planar-based registration algorithm is proposed for fast 3D mapping. It extracts plane-segments from point clouds obtained by a 3D sensor, turning out to be faster than previous point-based approaches. This approach needs a large number of planes to produce a reliable result as it uses least square techniques and consensus approach. This algorithm was also tested with coarse and noisy data from a sonar for underwater 3D mapping \citep{Pathak2010a}. Plane-based approach has also proved to be useful in the registration of buildings under construction to provide an effective project control. In this way, \citep{Bosche2012} presented a novel semi-automated plane-based registration system for coarse registration of laser scanned 3D point clouds. A real-time SLAM system that uses both points and planes for registration from 3D sensors can be found in \citep{Taguchi2013}. This mixed approach enables faster and more accurate registration than using only points. Hence, plane registration methods are successfully used for coarse registration.

A different approach has been proposed recently by \cite{Ahmed2015}. They extract Virtual Interest Points from planes corresponding to the intersection of several planes. The method registers them using the location for the translation, and the two largest normal angle found for the rotation. They need large planes to ensure a reliable planar model and hence the intersection of them, which is not necessary in our proposal.

Regarding specific object reconstruction algorithms, different methods have been developed using low-cost RGB-D sensors. \cite{Ramos2012} used a continuous rotation platform (turntable) knowing exactly the transformations between consecutive views. In this case, their method provided registration results that do not allow to distinguish some features of the shape that could be necessary in certain problems. In the same way, RGBDemo software \citep{Kramer2012,burrus2011} used 2D colour markers (ARToolkit markers) to make a initial coarse registration, then a traditional ICP is applied for a final fine registration. In \citep{Mihalyi2015}, the authors also proposed the use of this kind of 2D markers for reconstructing objects using general purpose RGB-D sensors for robotic purposes. The problem of using these markers is that the corner estimation quality relies in the colour camera resolution, which is low in RGB-D sensors. Then, a large number of markers need to be placed to compensate possible errors. Other works use expensive equipments to acquire the 3D model, such as KIT Object Model Database \citep{Kasper2012}, which provides accurate results but cannot be proposed as a general purpose devices due to the high cost.

The problems described above become a challenging problem when intricate shape and small objects have to be reconstructed using low cost depth-sensing cameras. Hence, new techniques should be conceived. In this paper, we propose a novel MUltiplane 3D MArker based Registration method ($\mu$-MAR) to deal with this problem. $\mu$-MAR uses known 3D markers around the object to be reconstructed and extend our previous registration method \textit{MBMVR} presented in \citep{Saval-Calvo2013} in order to reconstruct static objects with high accuracy. Since the marker is known, a model-based registration could be applied reducing noise effects and occlusions. The object to be reconstructed will take benefit from the marker registration transformations to properly register the corresponding views with higher accuracy than if it is reconstructed using the object raw data. Additionally, we propose the use of multiple views of the same part of the scene to increase the accuracy of the reconstructed model. Hence, the main contributions of this paper are the use of 3D markers to perform a model-based registration avoiding the problems associated to the noise and resolution of the point sets provided by low-cost depth-sensing cameras; and the proposed iteratively multi-view registration method to finely match same parts of the marker to increase accuracy. Since the method uses the same transformations to register the whole scene, the objects could be reconstructed with high accuracy. Moreover, this method is not restricted for RGB-D sensors, so that it could be applied using any source of 3D data.

The rest of the paper is divided as follows: section \ref{sec:reg:overview} presents an overview of the proposal. Section~\ref{sec:reg:ransac} describes the method to extract model of planar markers. Section~\ref{sec:reg:extMBMVR} presents the proposed registration algorithm for planes and the final reconstruction of the object. Section~\ref{sec:reg:exp} describes the experiments and shows the results of the method. Section~\ref{sec:reg:discussion} discusses the proposed method according to its contributions and limitations. Finally, section~\ref{sec:reg:conclusions} shows the extracted conclusions and proposes future research lines.

\section{Overview of the Multiplane 3D Marker based Registration method}\label{sec:reg:overview}

\begin{figure}[!ht]
 \centering
		\includegraphics[width=\textwidth]{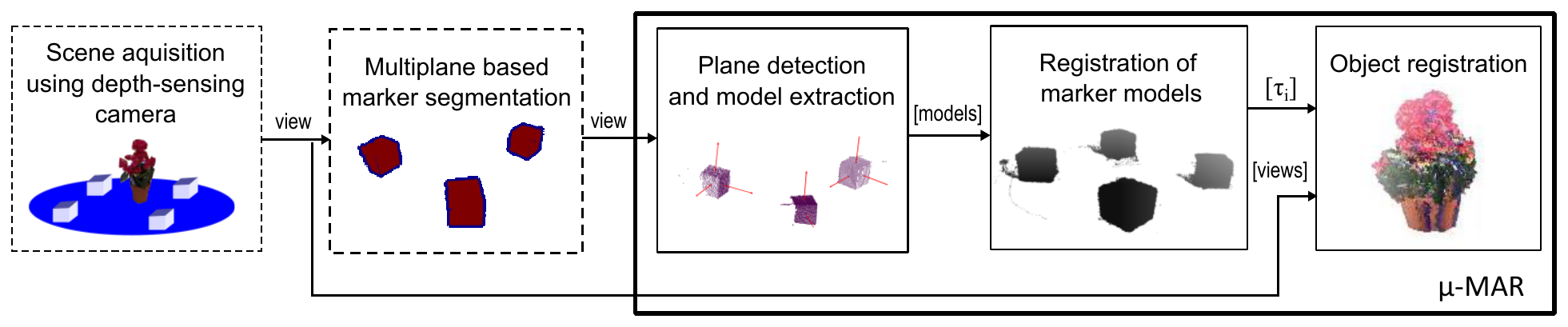}
	\caption{Overview of the proposed object registration method. Hyphened rectangles represent previous steps. The core of the method is composed by a plane detection, marker model extraction, registration of markers and, finally, object transformation.}
	\label{fig:scheme}
\end{figure}

The proposed method uses external markers to help in the registration process. Since the quality of data is not appropriate for state-of-the-art methods, here it is presented a proposal which uses 3D markers formed by planes. As the plane models reduce the noise effects, the proposed method is able to provide proper alignment with high accuracy. 

In order to reconstruct a specific object, some known objects have to be placed around it (see Fig.~\ref{fig:scheme}). The known objects (hereafter also called markers) have to be composed by multiple planes (e.g. cubes, pyramids). A multiple view acquisition of the object and the markers is carried out by a depth-sensing camera. The method assumes a large number of views close each other to ensure the acquisition of all parts, reducing the occlusion effects and the low accuracy of the camera in order to have enough information of each part of the object.  Each view acquired from the scene is aligned in a common coordinate system using the proposed  MUltiplane 3D MArker based REgistration ($\mu$-MAR) method. 

The $\mu$-MAR method for objects is divided into tree main steps (see the last three steps framed with a box Figure~\ref{fig:scheme}): plane detection and model extraction; registration of marker models; and, finally, object registration. 

The first step uses only the raw depth information of the markers, extracted from a simple colour and depth based segmentation, to detect the multiple planes that compose the markers. The planes after detected are fitted in a geometric model taking into account a prior knowledge about plane constraints in the marker (an explanation of this step can be found in Sect.~\ref{sec:reg:ransac}). 

The second step is the core of the $\mu$-MAR. It uses the geometric models of the markers as input to provide a fine registration of planar models using a iteratively multi-view technique that is presented in detail in Sect.~\ref{sec:reg:extMBMVR}. This method is an extension of our previous \textit{MBMVR} proposal \citep{Saval-Calvo2013}.

Finally, the method registers the specific object using the same transformations (i.e. rotations and translations) calculated for the markers that are applied to all views acquired by the camera. In other words, the object is transformed in the same way as the markers are done.

In order to ensure the best alignment of the planes different aspects have to be considered. The number of markers is a critical aspect, many of them will help the accuracy but could slow down the processing time. If few markers are place around the object and, hence, few planes appear in the scene, an ICP could be applied during the registration but taking into account some aspects: ICP should be applied in each loop of the multi-view process (see Sect.~\ref{sec:reg:extMBMVR}) after the registration of planes (normals alignment and centroids approximation), and is highly recommended to use a worst match rejection variation of ICP with a high percentage of rejection. Moreover, the markers should be placed in the way that at least three non-coplanar planes appear in the scene. Otherwise, the registration could end with undesired results.

\section{Plane detection and marker model extraction}\label{sec:reg:ransac}

This step is able to detect and extract a model of the marker from the segmented view. The model of a plane is established as the normal vector and a point belonging to the plane. Since, we are interested in a specific area of a plane (i.e. faces of the object), the point is the centroid of the region. In order to calculate this step, our method called Multiplane Model Estimation proposed in \citep{Saval-Calvo2015} is used. For the sake of completeness of the paper, a summarize is presented.

The proposal is a variant of the well-known RANSAC method incorporating prior knowledge of the markers to accurately estimate the model of the planes. This prior knowledge is understood as constraints about geometrical features of the markers, without taking into account size or position. Therefore, the method allows the free use and location of different markers (e.g. cubes, pyramids...). These constraints enforce the search of the planes that fit the points of each face of the marker. For example, if the marker is a cube, the constraints could set three as the maximum number of planes visible in a single view and 90 degrees as the angles among the faces that conform the cube. 


\begin{algorithm}[h]
 
 
 
 \KwData{$p=\{x,y,z\}$}
 \KwResult{$\{norm_i,cent_i\}$  }
 $(d_1,d_2,...,d_m) = clustering(p,normals(p),n);$ \label{alg:cluster} \\
 
 \While{not Constraint}{
	$set_{ini1}$ = select randomly n points from $d_1$\;
	$\{norm_1, cent_1\}$ = extractModel($set_{ini1}$)\;
	...
	Repeat for all $d_i$ \\
	
	\If{Planes fit the constraints}{
		$Constraint = OK$\;
	}
 }
 \For{every point in $d_1$ not in $set_{ini1}$}{
	$\{norm_{aux}, cent_{aux}\}$ = extractModel($set_{ini1} + point$)\\
	\If{Planes fit the constraints}{
		add point to $set_{ini1}$\;
	}
 }
 ...
 Repeat for all $d_i$
 \caption{Modified RANSAC for model plane estimation} \label{alg:ransac}

\end{algorithm}

Formally, the algorithm is described in Alg.~\ref{alg:ransac}. The result is a set of planar models (normals and centroids) that fit 3D points (input \textit{Data}) of the marker. Initially, the algorithm cluster the point cloud into $m$ regions using points and normals as input of the k-means algorithm ( Alg.~\ref{alg:ransac},  line~\ref{alg:cluster}). The parameter $m$ is estimated depending on the marker. For example, if we have a cube, three faces at most will be visible in a single view. Therefore, $m$ will be 3 plus a variable percentage to handle the noise (e.g. if 40\% is used, round($3 + 3*0.4$) = 4). In the clustering step, there are also some sets that are joint and rejected to finally return the best clusters that could fit the input data.

Once the clusters belonging to faces are defined, the actual models of the planes are estimated. In order to do this, a variant of RANSAC is used. This RANSAC variant first estimates the plane models (centroids and normals)  with a subset of points of each cluster, $d_i$ (Alg.~\ref{alg:ransac}, lines 3 and 4). With these models, the algorithm takes into account the constraints of the object (e.g. angles between planes) and evaluates if the planes fit these constraints (line 6 in Alg.~\ref{alg:ransac}). If the planes fit the constraints, the procedure continues evaluating the number of points (not in the initial subset) that belong to the plane models (Alg.~\ref{alg:ransac}, lines 10 to 16). In this final step the constraints are also evaluated to ensure an accurate result.

\section{Registration of marker models}\label{sec:reg:extMBMVR}

Planar models previously estimated are used by $\mu$-MAR to register the marker, and hence the scene. It is not necessary to use the previous technique explained in Sect.~\ref{sec:reg:ransac} since the registration method only needs planes whatever the source is, however the planes estimated with the method in Sect.~\ref{sec:reg:ransac} ensure accurate models and then improve the registration results. The better the planes fit the 3D data, the better the result will be. This independence in the input data makes the method able to work with any sensor as long as it is possible to obtain a 3D point cloud and estimate the planes from it.

A particular instance of the registration method presented in this paper was used previously to reconstruct a single object composed by planes in \citep{Saval-Calvo2013}. In the present paper, the method is applied to register markers and extended to take into account several planar-based objects that are in the scene. The extension solves different issues consequence of registering several markers: the resolution and accuracy of the markers data are different depending on the distance to the camera, occlusions could appear, etcetera. 

All markers are used together to obtain a full scene registration. The proposed method is based on a multi-view registration technique. It was initially proposed by Pulli in \citep{Pulli1999}. The main idea is to register iteratively each view on the rest. The advantage is that the noise or error in registration are minimized as the target of registration is not a single view but a group of them. Multi-view technique can be applied using the whole group of views or a subset of them. Imagine a group of views ($V_{1..n}$) to be registered, we mark a view as Data ($D = V_i$) and use the rest of views ($S = V_{1..n \neq i}$) as a target Scene. \textit{S} can be used as it is, a concatenation of views, or as a model that represents the views. For instance, the model could be the mean values of the views, a B-spline curve, a voxelization of the views, a kernel in which a probability density function is estimated from the views, among others. Once the \textit{S} is obtained, the registration process is applied to estimate the transformation to align \textit{D} to the Scene. Iteratively \textit{D} changes in the different views of \textit{V} until all of them have been registered. This process could be applied iteratively until convergence. 

\begin{figure}[h!]
  \begin{center}
  	\begin{subfigure}[t]{0.2\textwidth}
               \includegraphics[width=\textwidth]{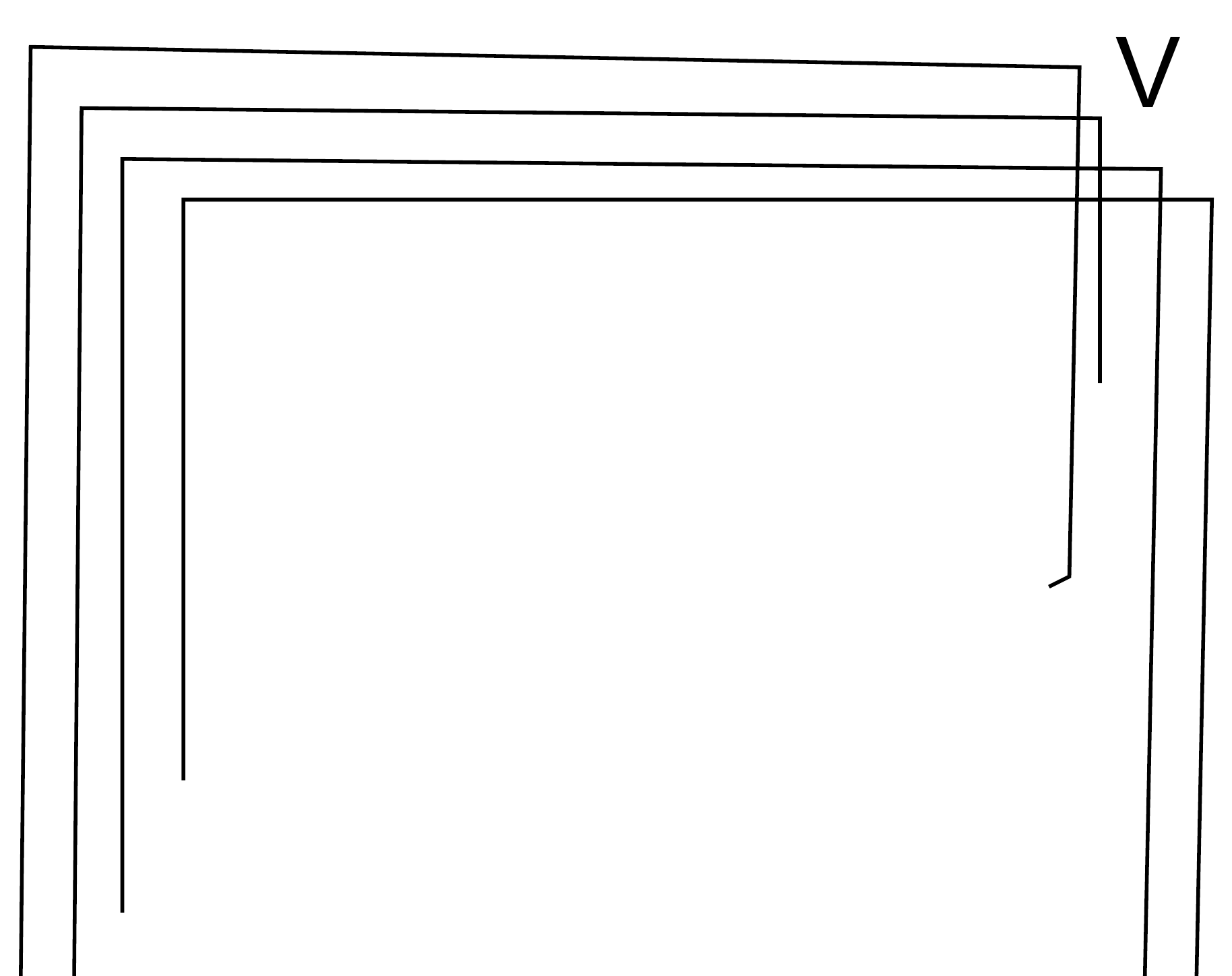}
                \caption{}
                \label{fig:reg:MVScheme:a}
        \end{subfigure}
        \begin{subfigure}[t]{0.2\textwidth}
               \includegraphics[width=\textwidth]{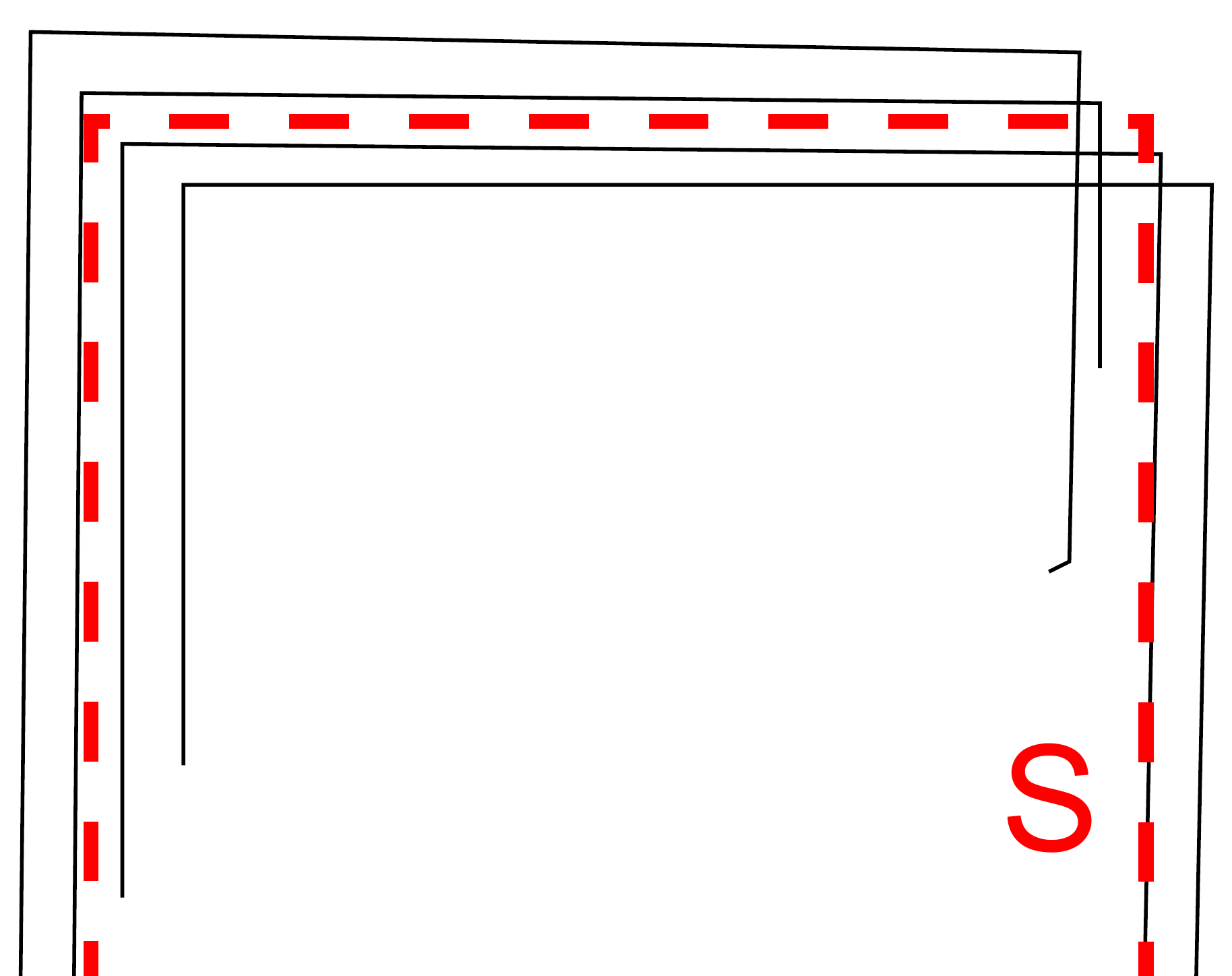}
                \caption{}
                \label{fig:reg:MVScheme:b}
        \end{subfigure}
        \begin{subfigure}[t]{0.2\textwidth}
 				\includegraphics[width=\textwidth]{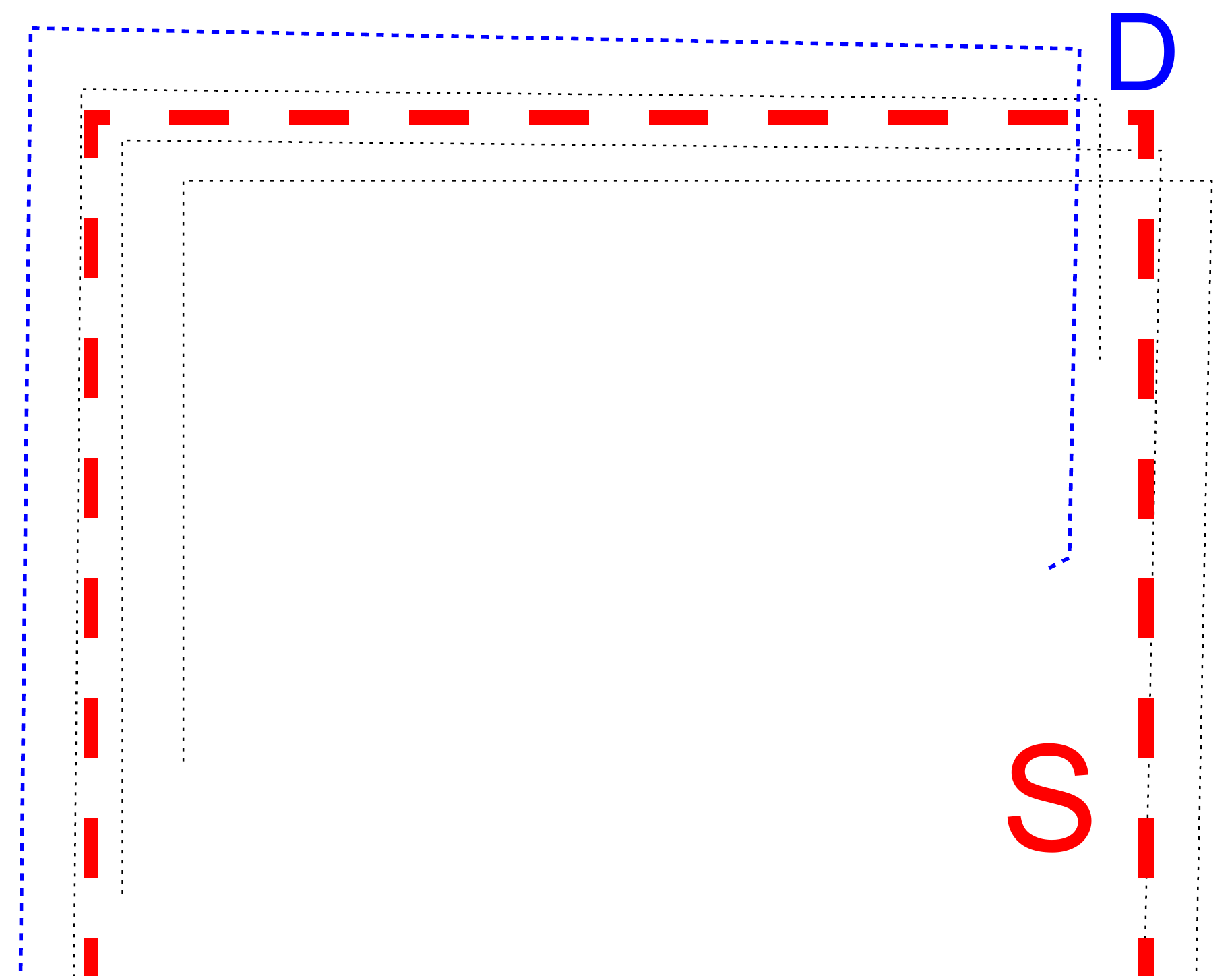}                				
 				\caption{}
                \label{fig:reg:MVScheme:c}
        \end{subfigure}
        \begin{subfigure}[t]{0.2\textwidth}
 \includegraphics[width=\textwidth]{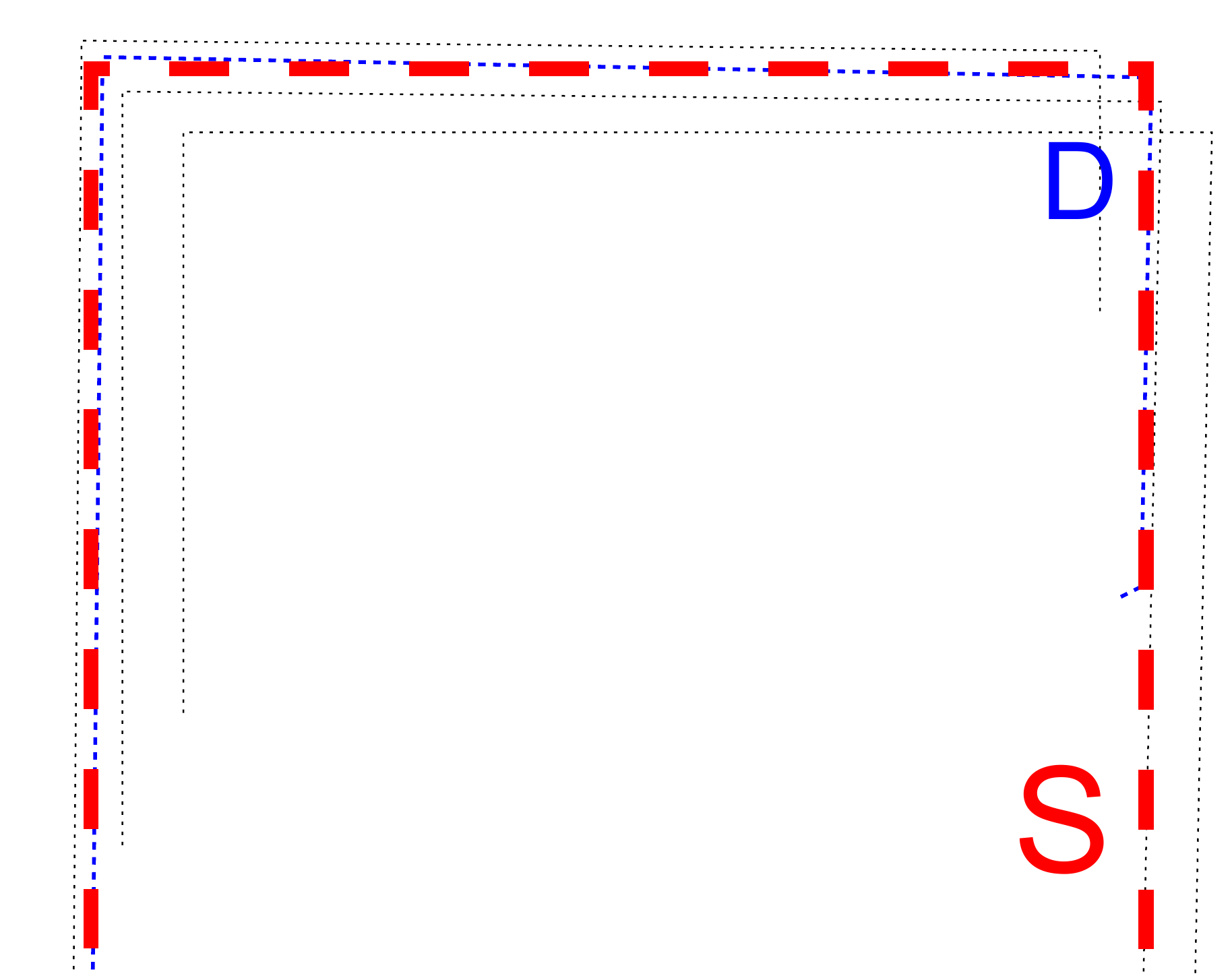}
 \caption{}
                \label{fig:reg:MVScheme:d}
        \end{subfigure}
 
 	\caption{General scheme of a multi-view registration process. Set of views (\textit{V}) (\ref{fig:reg:MVScheme:a}). Scene (\textit{S}) represented by a hyphened red line (\ref{fig:reg:MVScheme:b}). Data (\textit{D}) plotted in a blue dotted line (\ref{fig:reg:MVScheme:c}). Registration result of \textit{D} onto \textit{S} ((\ref{fig:reg:MVScheme:d}).}
 	\label{fig:reg:MVScheme}
  \end{center}
\end{figure}

Figure~\ref{fig:reg:MVScheme} shows an example of a multi-view process. In \ref{fig:reg:MVScheme:a} image a set of views of a square side is represented. Initially $D = V_1$ and $S = mean(V_{2..4})$, shown in Fig. \ref{fig:reg:MVScheme:b} and \ref{fig:reg:MVScheme:c} images. After registering the Data with Scene the result is as the plotted in the  Figure~\ref{fig:reg:MVScheme:d}. Once $V_1$ has been registered, $D$ is assigned to $V_2$ and $S = mean(V_{1,3,4})$, and continue until all views are registered.

\begin{figure}
	\begin{center}
		\includegraphics[width=1\textwidth]{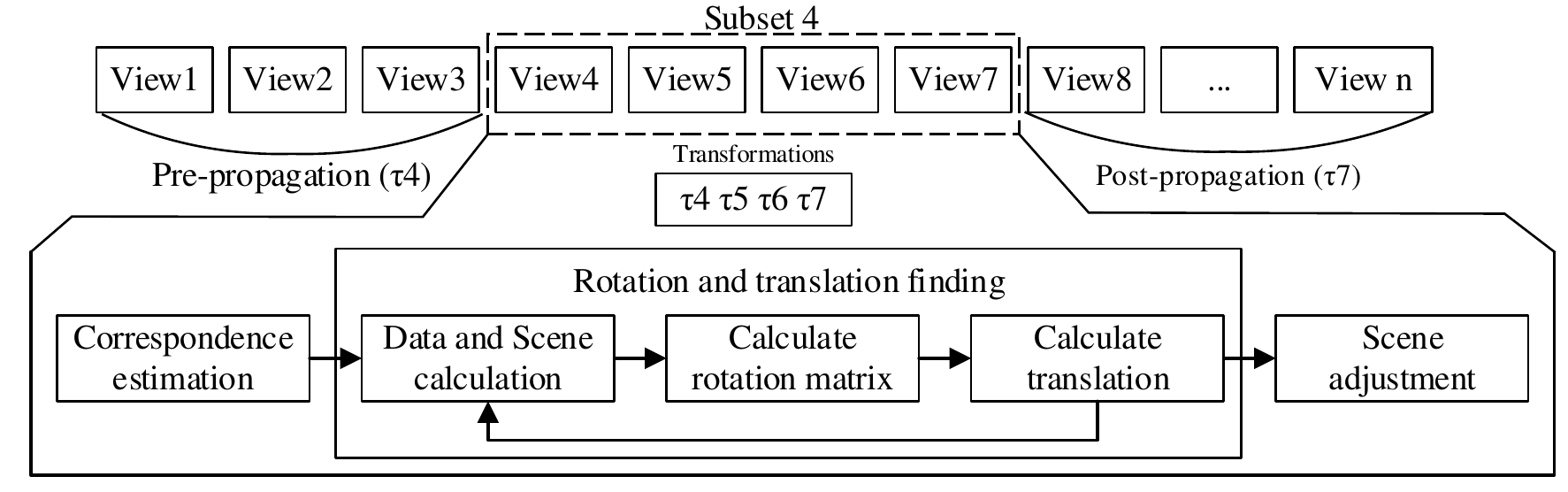}
		\caption{General scheme of registration process. The view $V_4..7$ are registered in order to find $\tau 4..7$ that aligns all views together. The registration is composed by correspondence estimation of the planes, $D$ and $S$ selection, rotation and translation ($\tau$) estimation and scene adjustment. $\tau 4$ is pre-propagated and $\tau 7$ post-propagated in order to maintain the coherence of the previous and posterior views out of the subset. }
 	\label{fig:reg:RegScheme}
	\end{center}
\end{figure}

As it is said before, multi-view registration could be applied to either the whole set of views or only a subset of them. In the case of this work, a subset is used because of the large number of views. Therefore, the general process takes iteratively subsets of \textit{n} consecutive views and applies a multi-view registration composed by three main steps: estimation of model correspondence in the different views (Sect.~\ref{sec:correspondence}); rotation and translation finding (Sect.~\ref{sec:transformation}); and complete scene adjustment (Sect.~\ref{sec:adjustement}). A complete scheme of the proposed registration is shown in Figure~\ref{fig:reg:RegScheme}.

With this strategy, the set of \textit{n} views is iteratively registered using the plane models (normals and centroids) belonging to different markers from one view as set to be registered, \textit{D}, and the mean of the corresponding plane models of the remaining views in the subset as a static target, \textit{S}. This process is repeated until all views in the subset are registered to the remaining views. Finished the registration of a subset, a new subset is selected until the method finishes to register all captured views. In the example of the Figure \ref{fig:reg:RegScheme} the size $n$ is four, then the first corresponds to the views $V_{1..4}$. Once the multi-view is finished, the subset changes to the $V_{2..5}$, and so on until the subset $V_{n-3..n}$. In order to keep the previous and post registrations, a propagation of the calculated transformations is performed. The transformation of the first view of the subset is applied to all previous views (in the Figure~\ref{fig:reg:RegScheme}, pre-propagation $\tau 4$). Similarly, a post-propagation (Figure~\ref{fig:reg:RegScheme}, post-propagation $\tau 7$) is applied to all views after the subset using the last view in the subset. This is done for all subsets, with the exceptions of the first and last, where they do not have pre-propagation and post-propagation respectively. 

Moreover, if the views are not close each other, we suggest to apply a initial pairwise registration of the views in the subset. The reason is that when the views are separated and the \textit{Scene} is calculated as the mean value of the views, the deviation produces wrong normals and centroids, what drives the whole algorithm to a misalignment.

\subsection{Correspondence estimation}\label{sec:correspondence}

Correspondence estimation is a key step in every registration method. ICP uses closest point, featured-based techniques use similarity between descriptors. We use the similarity between plane models in the different views using only 3D information avoiding any other information that restrict the use of any marker as, for example, colour or texture. Various techniques can be applied to estimate correspondences between subsets of views groups of data (Fig.~\ref{fig:reg:regsteps:1}). In this case, we propose the use of similarity between normal directions and centroid positions due to the views are close each other (Fig.~\ref{fig:reg:regsteps:2}). To evaluate the similarity a k-nearest neighbours technique is used with both centroids and normals. 

\begin{figure}[h]
	\centering
	\includegraphics[width=0.4\textwidth]{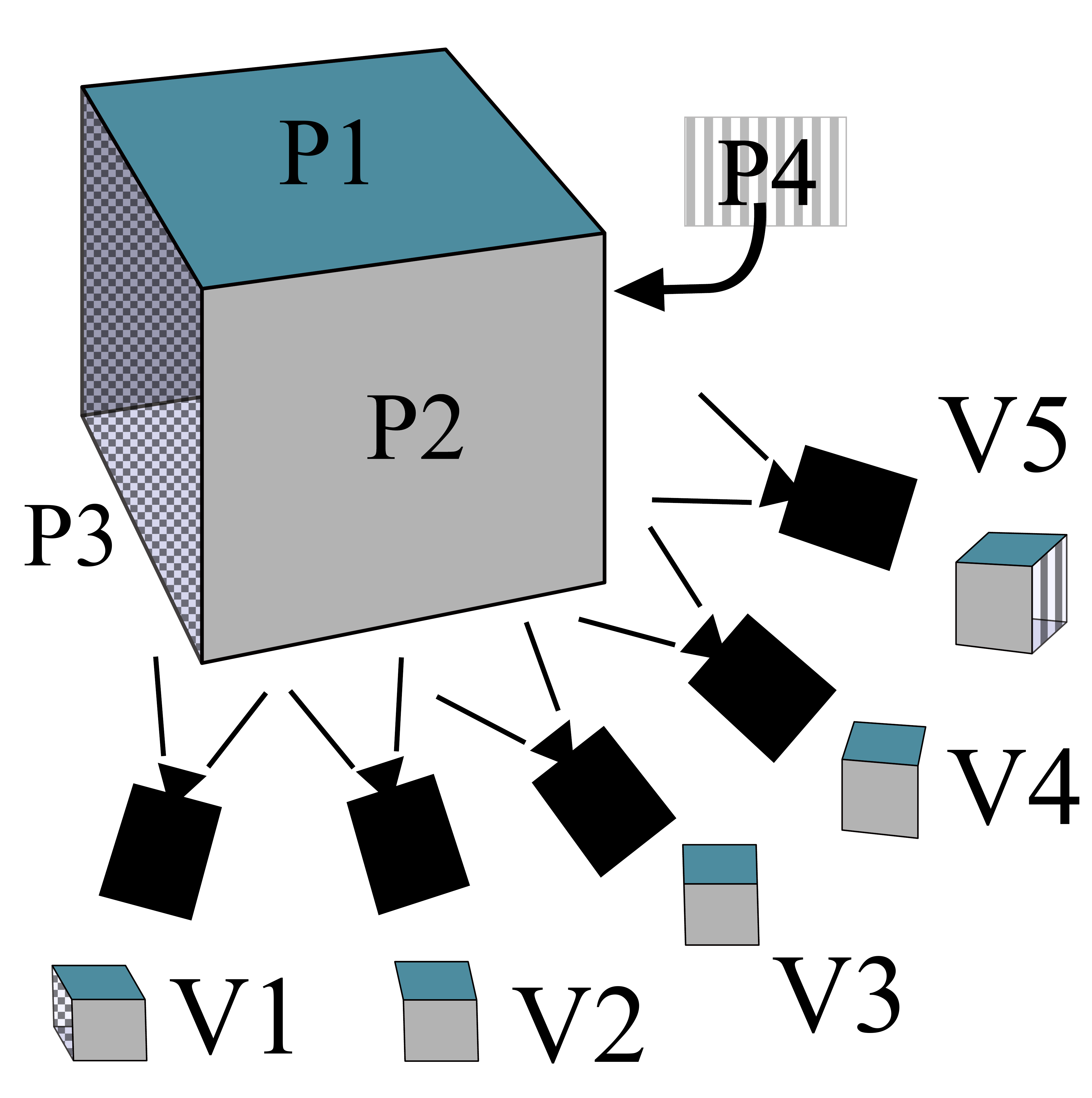}
	\caption{Image representing the different views of a cube. This is a graphical representation of the situation shown in Table \ref{tab:reg:matcorr}.}
	\label{fig:reg:modelcubeviews}
\end{figure}

When the point of view changes, some parts of objects (planes in this case) appear or disappear. Within the \textit{n} views in the subset of the multi-view this problem happens, so a robust correspondence technique is used to handle it. We propose to use a structure composed by \textit{n} columns (one per view) and as many rows as total number of planes in the subset. This structure is created dynamically while the correspondences are estimated. The process calculates the similarity in pairs as $V_1 \Leftrightarrow V_2$ to $V_{n-1} \Leftrightarrow V_{n}$. For example, imagine a situation in which one marker is registered. In this example, five views are in the subset of the multi-view (Figure~\ref{fig:reg:modelcubeviews}). In the first view, three planes are visible, so we have three rows. Then, in the second view, one of the sides disappear because of the new point of view (in the figure the chessboard side), so in the second column one row will be empty.  It continue with only two planes for the second, third and forth view. Finally, in the fifth a new plane appears (in Figure~\ref{fig:reg:modelcubeviews} the strips plane), then a new row is added and all columns but the fifth will be empty in this row. So eventually, the structure will be like appears in Table~\ref{tab:reg:matcorr}.

In order to be provide a more robust solution, the correspondences are also evaluated using the immediate previous and two previous views. Then, we choose the correspondences between the view that has more planes in common. With this, we avoid situations of planes that disappear because an exceptional problem (e.g. an unexpected brightness, or a surface which is almost parallel to the point of view of the camera that is difficult to acquire and sometimes appears and disappears). 

\begin{table}[h]
\centering
\scalebox{1}{

\begin{tabular}{|c|c|c|c|c|c|}
\hline
 & $V_1$ & $V_2$ & $V_3$ &$V_4$  &$V_5$ \\ 
\hline
P1 & $<C_{1,1},N_{1,1}>$  & $<C_{2,1},N_{2,1}>$ & $<C_{3,1},N_{3,1}>$ & $<C_{4,1},N_{4,1}>$ & $<C_{5,1},N_{5,1}>$ \\
P2 & $<C_{1,2},N_{1,2}>$  & $<C_{2,2},N_{2,2}>$ & $<C_{3,2},N_{3,2}>$ & $<C_{4,2},N_{4,2}>$ & $<C_{5,2},N_{5,2}>$ \\
P3 & $<C_{1,3},N_{1,3}>$  & Empty               & Empty               & Empty               &  Empty              \\
P4 & Empty                & Empty               & Empty               & Empty               & $<C_{5,3},N_{5,3}>$ \\
\hline
\end{tabular}}
\caption{Example of structure of correspondences. The table shows five view correspondences where each column is a view and row is a plane. $C_{i,j}$ means Centroid and $N_{i,j}$ Normal of the plane $j$ seen from the view $i$.}
\label{tab:reg:matcorr}
\end{table}

Once we know the correspondences, the subset of \textit{n} views is divided into two groups of plane models (centroids and normals): \textit{Data}, which is the set of models corresponding to a single view, and \textit{Scene}, which is the plane model defined as the mean value of centroids and normals of the corresponding planes of the rest of views. Referring to Figure~\ref{fig:reg:regsteps:3} and Figure~\ref{fig:reg:regsteps:4}, \textit{Data, D} is the first view in thin pink and \textit{Scene, S} is the mean value of the rest of view in the subset, the thick green.

\subsection{Rotation and translation finding}\label{sec:transformation}

\begin{figure}[H]
        \centering
        \begin{subfigure}[t]{0.4\textwidth}
                \includegraphics[width=\textwidth]{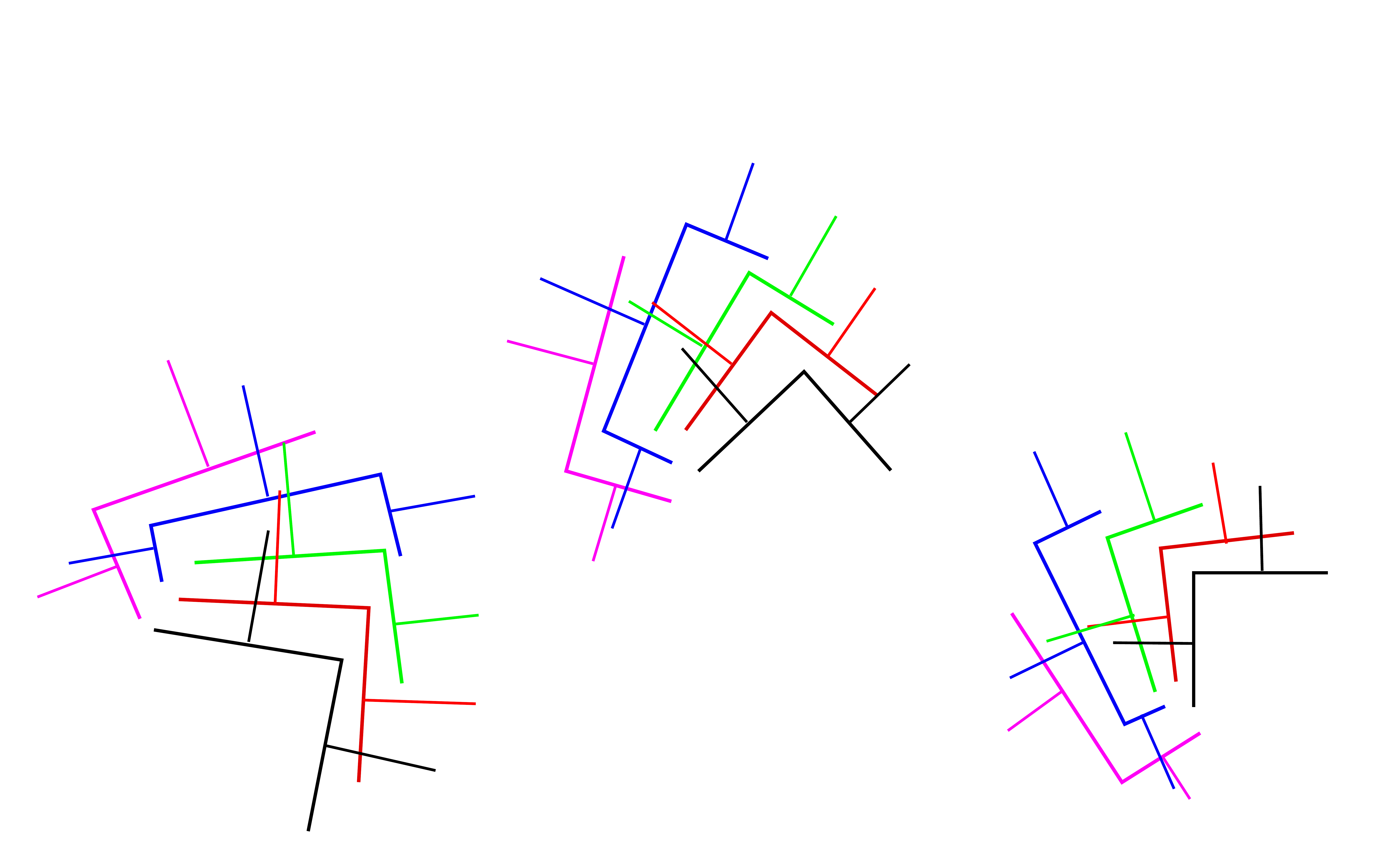}
                \caption{Subset of views of the multi-view.}
                \label{fig:reg:regsteps:1}
        \end{subfigure}
        \qquad \quad 
        ~ 
        \begin{subfigure}[t]{0.4\textwidth}
                \includegraphics[width=\textwidth]{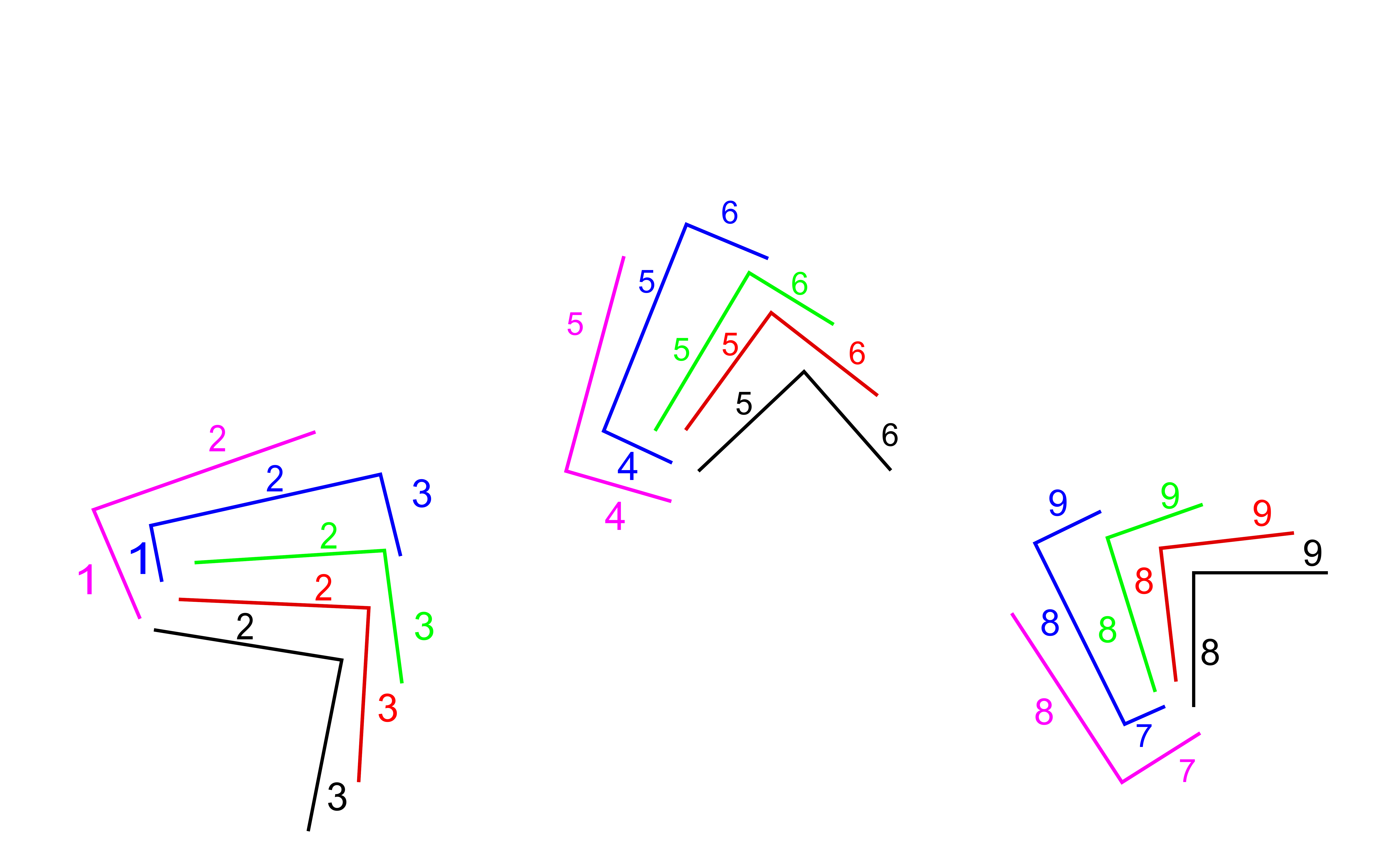}
                \caption{Correspondences}
                \label{fig:reg:regsteps:2}
        \end{subfigure}
        \\
        \begin{subfigure}[t]{0.4\textwidth}
                \includegraphics[width=\textwidth]{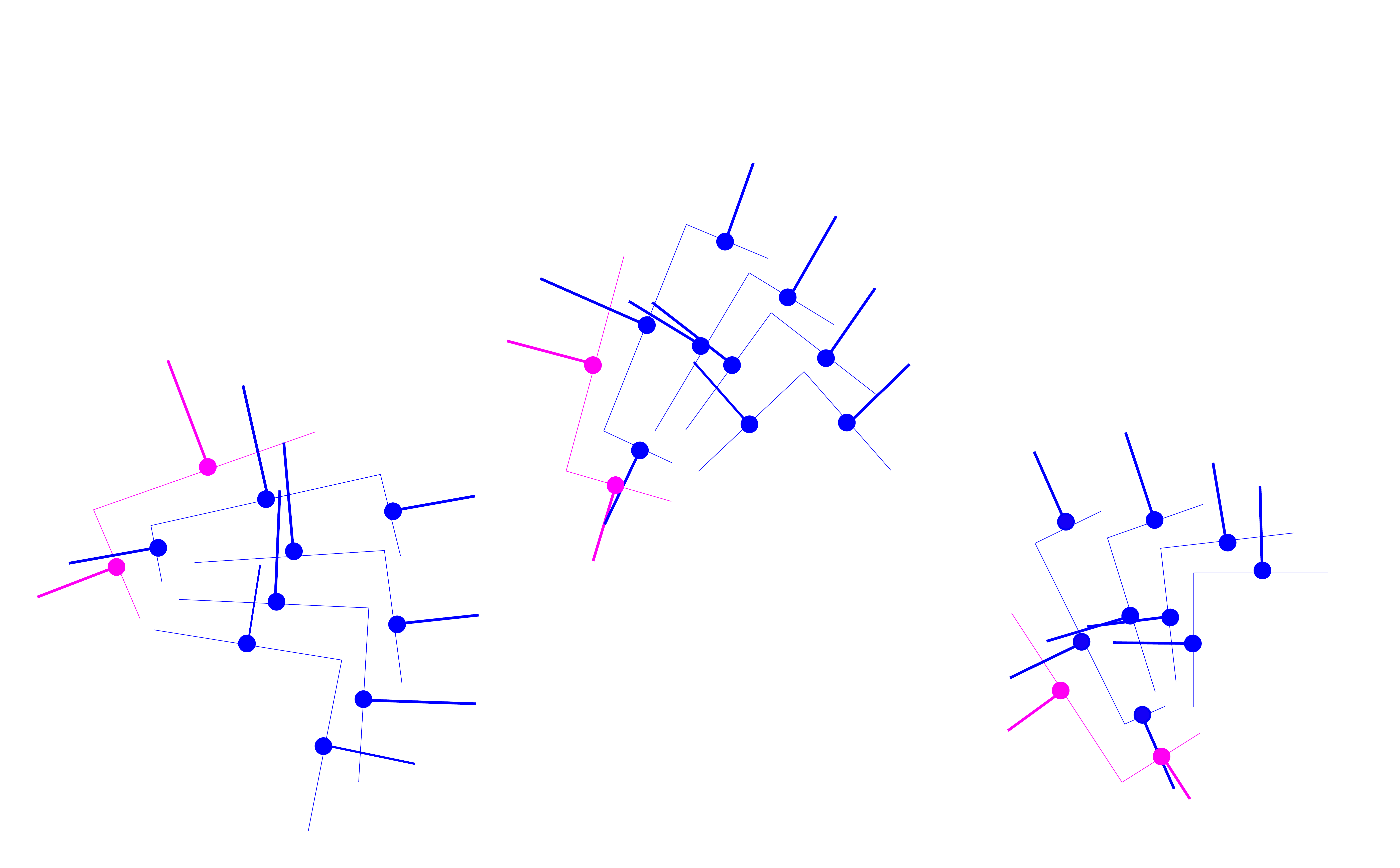}
                \caption{$D = V_1$ in pink and $S = V_{2..5}$}
                \label{fig:reg:regsteps:3}
        \end{subfigure}
        \qquad \quad 
      	\begin{subfigure}[t]{0.4\textwidth}
                \includegraphics[width=\textwidth]{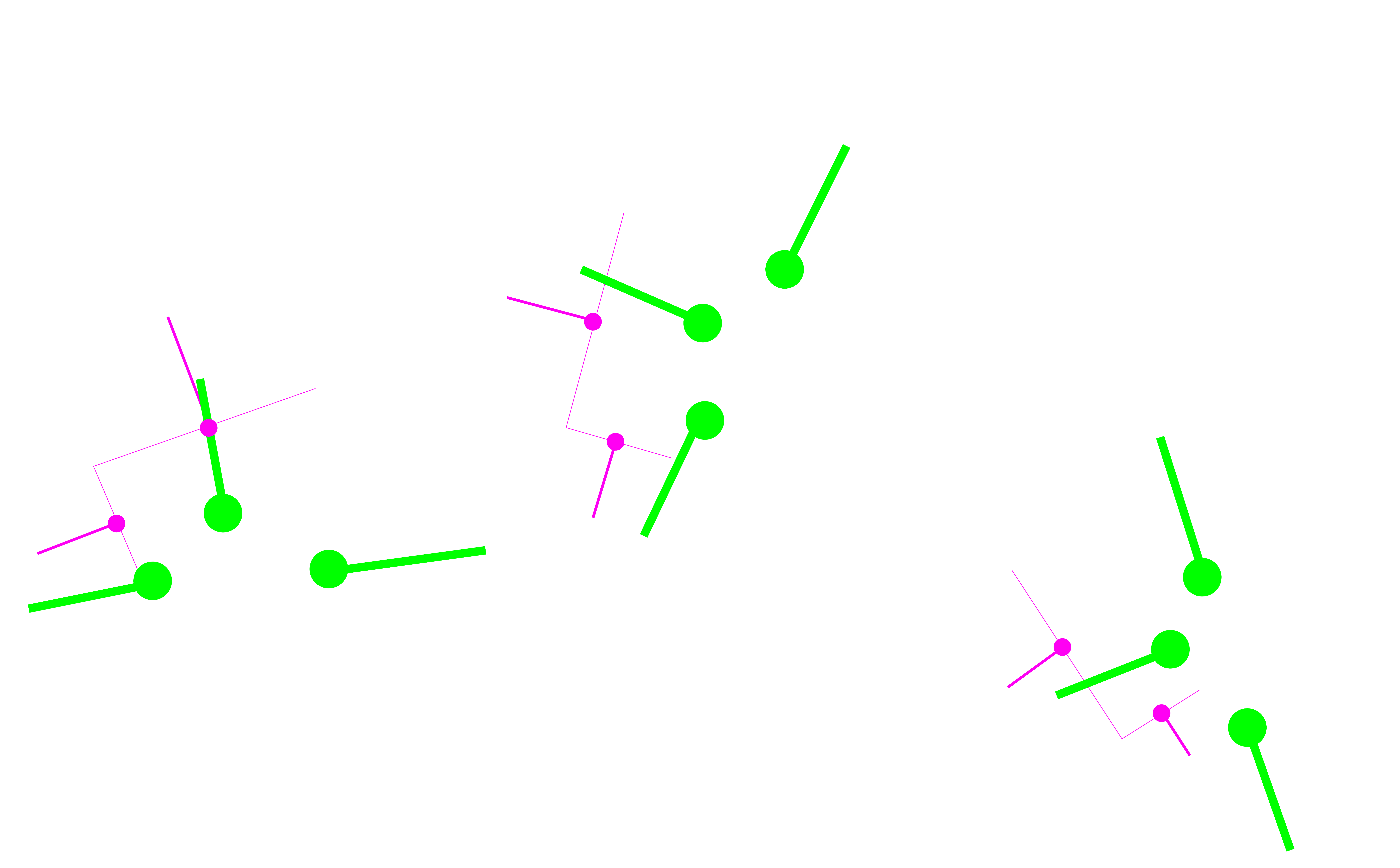}
                \caption{$D = V_1$ in thin pink and $S = V_{2..5}$ in thick green}
                \label{fig:reg:regsteps:4}
        \end{subfigure}
		\\
		\vfill		
		\begin{subfigure}[t]{0.4\textwidth}
                \includegraphics[width=\textwidth]{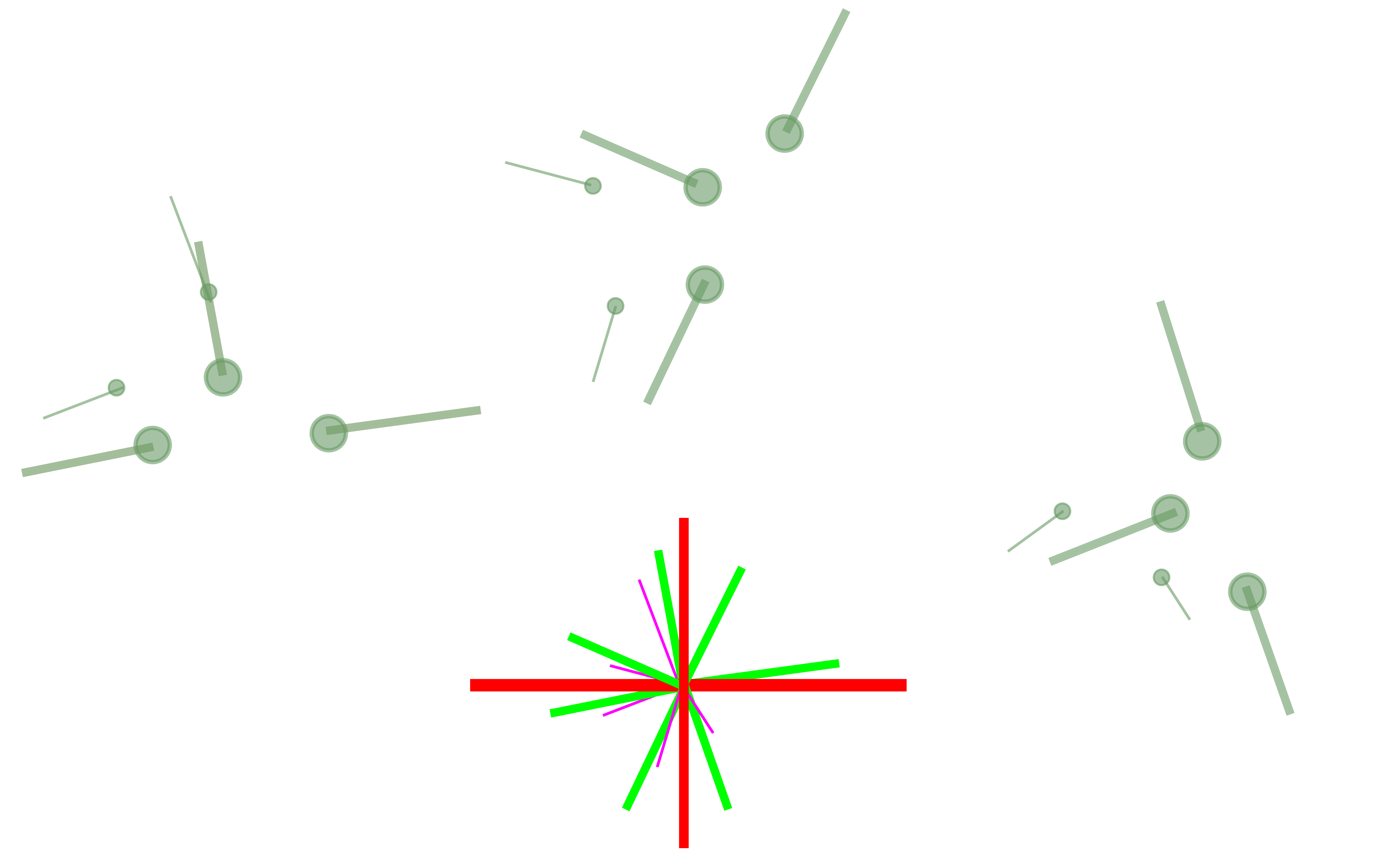}
                \caption{Rotation estimation using normals}
                \label{fig:reg:regsteps:5}
        \end{subfigure}
        \qquad \quad 
      	\begin{subfigure}[t]{0.4\textwidth}
                \includegraphics[width=\textwidth]{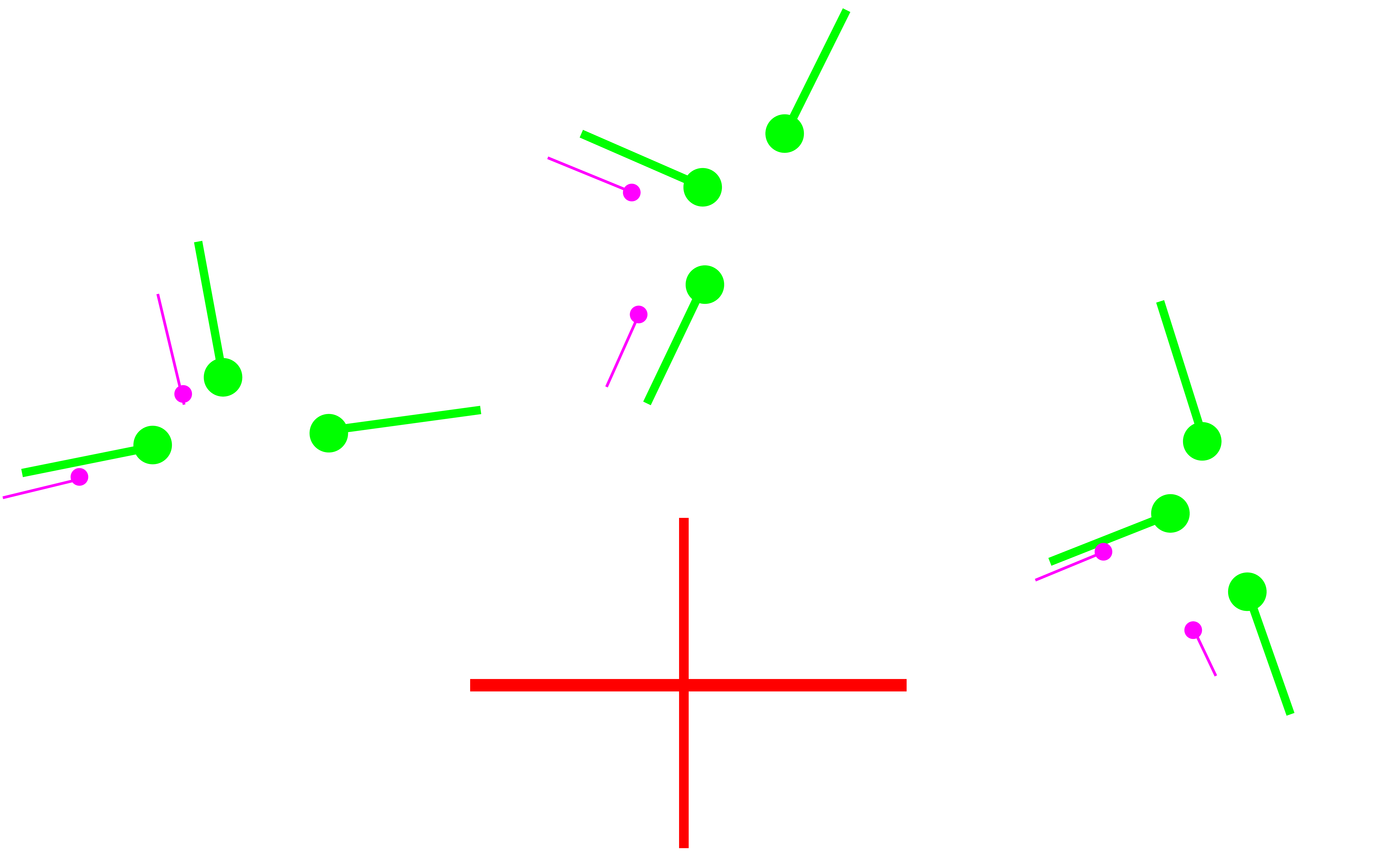}
                \caption{Alignment}
                \label{fig:reg:regsteps:6}
        \end{subfigure}
		\\
		\vfill		
		\begin{subfigure}[t]{0.4\textwidth}
                \includegraphics[width=\textwidth]{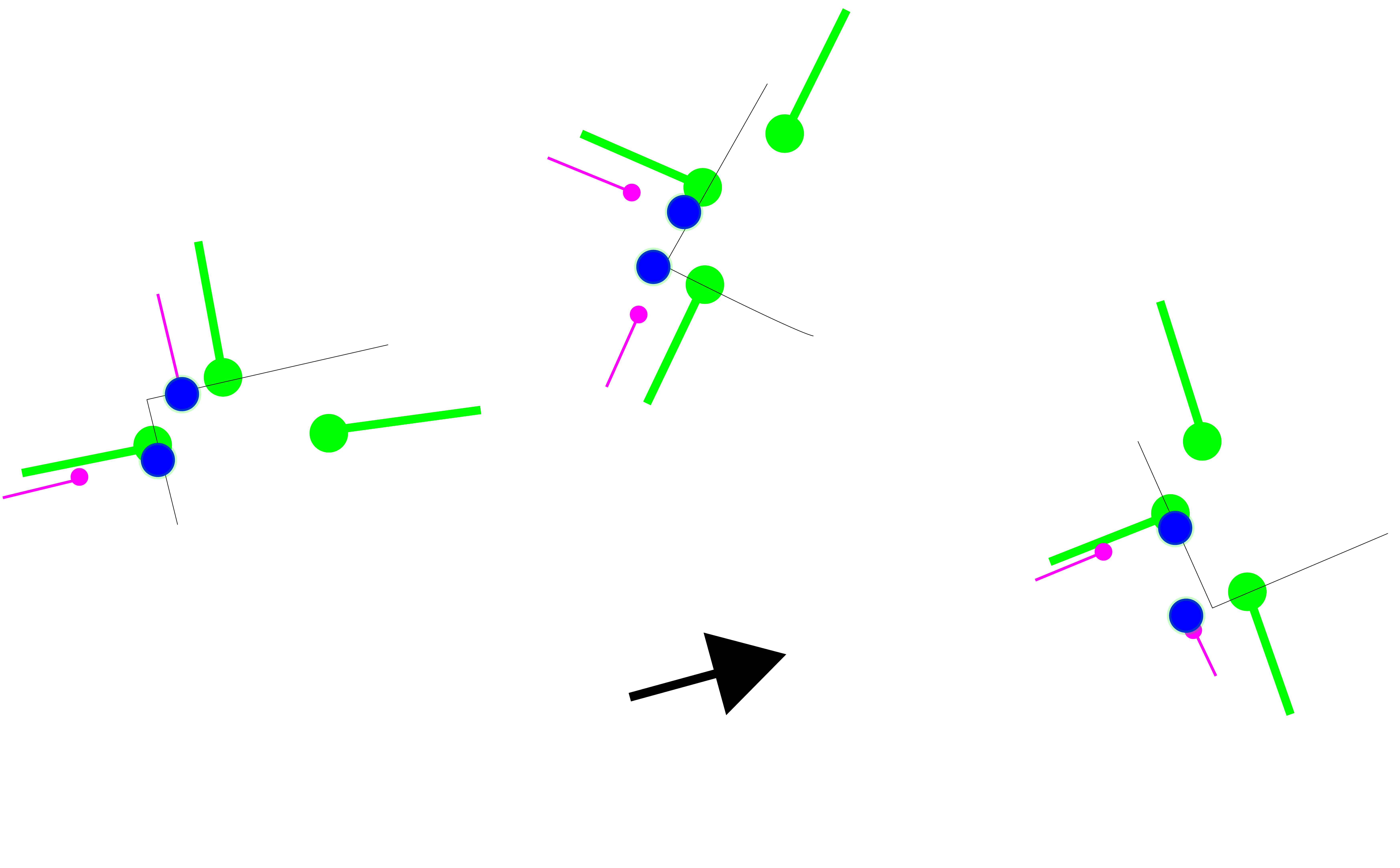}
                \caption{Projections estimation}
                \label{fig:reg:regsteps:7}
        \end{subfigure}
        \qquad \quad 
      	\begin{subfigure}[t]{0.4\textwidth}
                \includegraphics[width=\textwidth]{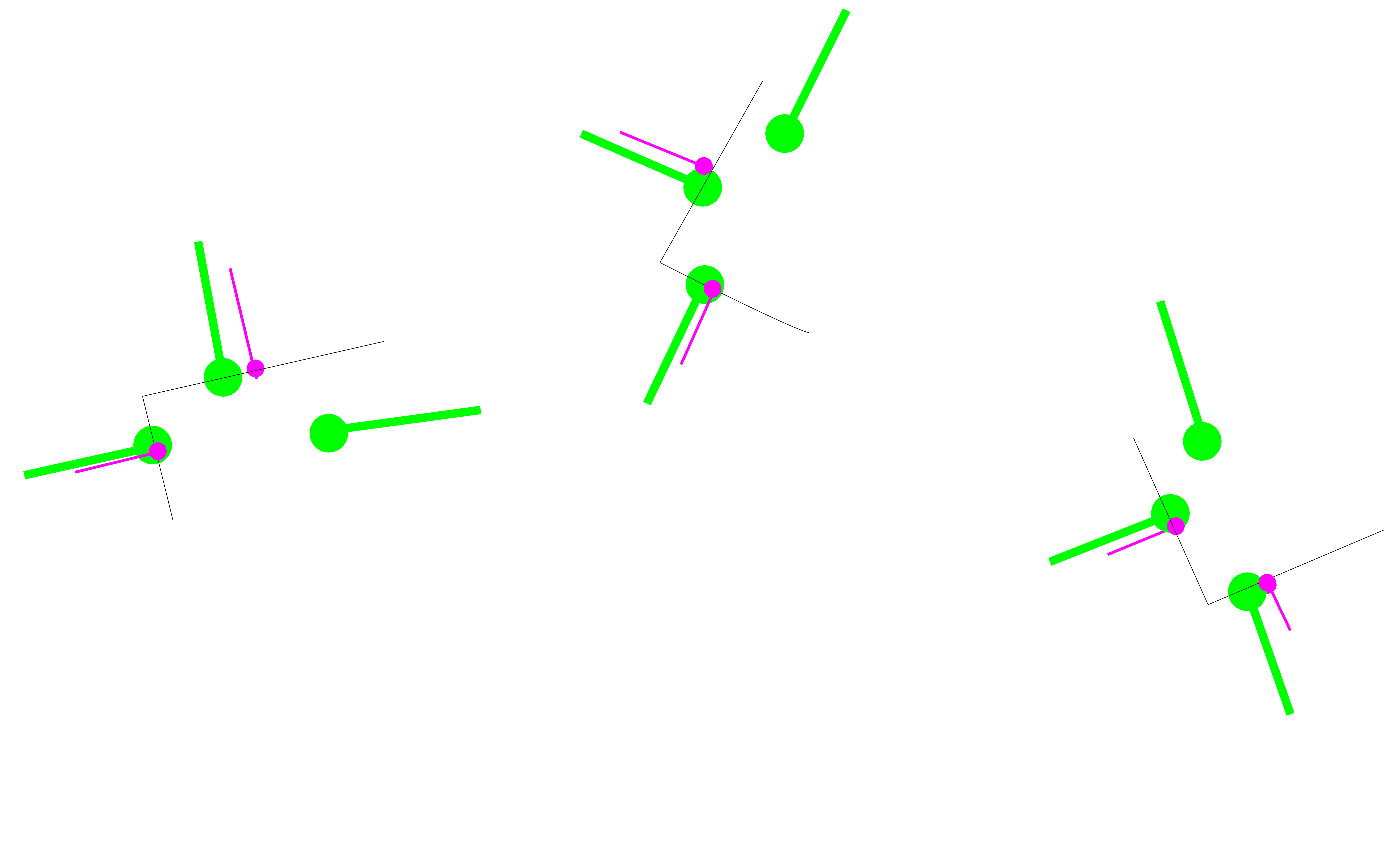}
                \caption{Translation}
                \label{fig:reg:regsteps:8}
        \end{subfigure}
        \caption{General registration workflow}\label{fig:reg:regsteps}
\end{figure}

The second step finds the best rotation and translation to align the \textit{Data} to the \textit{Scene}. The method estimates the rotation as the angle in each axis necessary to align $D$ and $S$. A well-know method to estimate transformations is \textit{Singular Value Decomposition}(\citep{Besl1992}~\citep{Muller2004}), of which we use the rotation matrix for this step. As we cannot ensure the relative position of centroids of each plane in $D$ and $S$ (we explain this better in the next paragraph), only normals are used for this purpose. As shown in Figure~\ref{fig:reg:regsteps:5} all normals are located in the origin of coordinates to calculate the rotation. With the result of SVD to the normals, both normals and centroids of $Data$ are rotated (Figure~\ref{fig:reg:regsteps:6}). 

The centroids are used to find the translation that minimizes the distance between each pair of \textit{Scene}-\textit{Data} planes. Centroid positions are relative to the points obtained from the sensor, their location could not correspond to the same absolute location of the same plane in two different views. For example, when a plane is appearing in the view only a part is visible. Hence, the centroid is calculated just for this part, while in a different view this same plane is complete and the centroid will be in the center of the whole face. Therefore, the projection of the \textit{Data} centroids on the \textit{Scene} planes is used  as the translation target, as it is shown in Figure~\ref{fig:reg:regsteps:7} with blue dots. Thereby, the translation is computed as the mean value of distances between each \textit{Data} centroid and its projection for each axis (see Figure \ref{fig:reg:regsteps:7} black arrow). Finally, Figure~\ref{fig:reg:regsteps:8} represents the translation result and hence the final registration. 

Both rotation matrix and translation vector conform a tranformation matrix (one for each view as all views are registered in the multi-view) which minimizes the distance between the Data and the Scene. After the registration of a single iteration of the multi-view, an error still remains in the alignment. This error is calculated in two dimensions, one the sum of the angles between each normal of the planes in D and the correspondents in S (as in multi-view each view is Data once, this error is averaged for all views). Second, the addition of distances between each centroind and the projection of them in the Scene (as with the normals the error is the average of all view errors). This error is expressed in Equation~\ref{eq:reg:error}. Then, the registration algorithm iterates until the error converges. That means, once an iteration of the multi-view finishes the views might not be finely aligned, it is, an error in the registration is still present. Then, more iterations of the whole subset is performed until convergence (i.e. the error is under a threshold).  expresses the considered error that takes into account rotation and translation error.


\begin{equation}\label{eq:reg:error}
error=\sum^n_{i=1} \sum^{numplanes}_{k=1} \left [angle(nl_{i,k},\mu(nl_{<1..n> \neq i, k})); dist(cn_{i,k},\mu(cn_{<1..n> \neq i, k}))\right ] / n
\end{equation}


where $nl_{i,k}$ means the normal of the view i and plane k (the Data normal) and $cn_{i,k}$ is the centroid of the view i and plane k (the Data centroid). The Scene (S) is represented as the normal and centroid of all views except the Data, i.e. $\forall V \neq i$. The function $\mu$ represents the mean value which is the Scene (S) used for the registration. Finally, $dist$ is the Euclidean distance.  


%


\subsection{Scene adjustment}\label{sec:adjustement}

The multi-view variant in this work uses a subset of views instead of the whole group of them. This is due to the large number of views not initially registered makes very difficult to converge. Therefore, a specific plane can be seen in a set of views, then disappear and reappear later. As the registration is performed only in the subset, this plane will not be registered in those views before and after reappearing. In order to handle this problem, we propose to have a general structure of the whole scene (see Table~\ref{tab:reg:scenecorr}), dynamically created when new planes appear. With this, not only a plane that was visible before can be readjusted, but also the cumulative error is minimized. 

This general structure stores the model (normal and centroid) of each plane in the scene. This structure groups per plane instead of per view, but it also has the number of view in which this plane was visible. As we assume an increasing subset views registration, we can ensure that every already registered view is well aligned with all posterior views. With this assumption, we add to the general structure the planes of the first view in the subset because this one is not going to be registered any more. To find the correspondence the same principle of the multi-view registration is used (i.e. normals and centroids are used with a k-nearest neighbour technique to find the similarity). As all views are aligned, the correspondences are perfectly detected, and when a new plane appears it is added as a new row. Table~\ref{tab:reg:scenecorr} shows and example of this general structure. Each column represents a plane in the scene. Hence, all plane models from different views that corresponds to each plane are stored there. For example plane $P3$ (see Figure~\ref{fig:reg:modelcubeviews}) is in $V_1$ but then disappears and is not visible until view $V_{10}$. As this structure is incrementally filled, the third column would be only with one value until the tenth view is already registered, and then added as to the general structure.

The purpose of this structure is to refine the registration, then before adding planes of the new view to this structure, a registration of the whole subset is performed to this general group of planes. In order to do this, a model of the subset is extracted as it is done in the registration part (mean of centroids and normals). Then, this subset model is used as Data $D$, and the mean value of all view for each plane in the general structure is used as Scene $S$. Then a registration process is applied as explained before. 

\begin{table}[h]
\centering
\scalebox{1}{

\begin{tabular}{|c|c|c|c|c|c|}
\hline
 & $P1$ & $P2$ & $P3$ & $P4$  &$P5$ \\ 
\hline
 & $<C_1,N_1,V_1>$  & $<C_2,N_2,V_1>$ & $<C_3,N_3,V_1>$  & $<C_3,N_3,V_5>$ & $<C_3,N_3,V_8>$ \\
 & $<C_1,N_1,V_2>$  & $<C_2,N_2,V_2>$ & $<C_3,N_3,V_{10}>$ & $<C_3,N_3,V_6>$ & $<C_3,N_3,V_9>$ \\
 & $<C_1,N_1,V_3>$  & $<C_2,N_2,V_3>$ & $<C_3,N_3,V_{11}>$ & $<C_3,N_3,V_7>$ & $<C_3,N_3,V_{10}>$ \\
 & $<C_1,N_1,V_4>$  & $<C_2,N_2,V_4>$ & $<C_3,N_3,V_{12}>$ & $<C_2,N_2,V_8>$ & $<C_3,N_3,V_{11}>$ \\
 & $\cdots$         & $\cdots$        & $\cdots$         & $\cdots$        & $\cdots$         \\
\hline
\end{tabular}}
\caption{General structure of planes for the Scene adjustment. Each column represents a plane with all normals ($C_a$), centroids ($C_b$, and view ($V_c$). As it is possible to appreciate, the plane $P3$ is present in view $V_1$ and then until view $V_{10}$ it is not visible.}
\label{tab:reg:scenecorr}
\end{table}

\section{Experimentation}
\label{sec:reg:exp}

The proposed method $\mu$-MAR has been tested in two different situations to validate its performance: firstly, theoretical aspects (i.e. to comprobe whether the methodology works properly) using synthetic data; secondly, in real environment with objects acquired by a RGB-D sensor. 

In order to evaluate the methodology proposed for $\mu$-MAR, a simulation of a turntable has been carried out (Section~\ref{sec:reg:exp:synthetic}). The scenario was created using a plug-in of the graphics software Blender, called Blensor \citep{Gschwandtner2011}, which simulates different sensors included a Microsoft Kinect. In the scene four cubes as markers and the target object are placed in front of the sensor and a 360 degrees turn is performed with 60 captures (6 degrees between frames). A representation of the scene is shown in Figure~\ref{fig:exp:scene} with four markers and a double pyramid in the center.

\begin{figure}[H]
\centering

\includegraphics[width=0.8\textwidth]{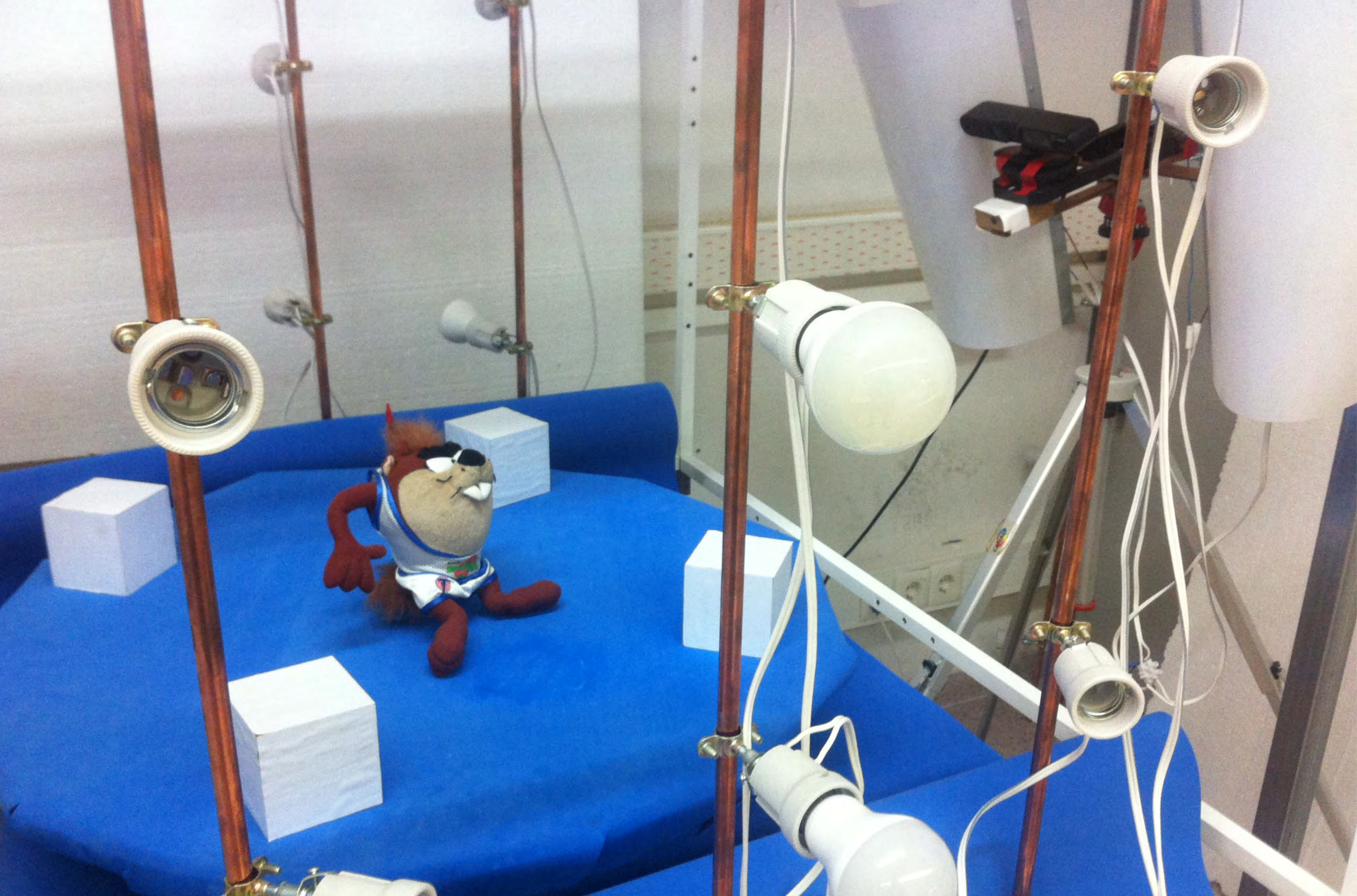}
\caption{Real turntable set up for the real data experimentation. The markers are white cubes around the target object (a Taz toy).}
\label{fig:reg:exp:turntable}
\end{figure}

For real data experiments (Section~\ref{sec:reg:exp:real}), four different objects were acquired by a Primesense Carmine RGB-D camera. This experimentation uses a real turntable and a scenario where the objects are acquired in a controlled environment. The 3D markers and target objects are placed in a similar position than in the synthetic set up (see Figure~\ref{fig:reg:exp:turntable}). In this case, the 360 degrees turn is performed with 64 captures (5.625 degrees between frames). In the evaluation, the same objects are registered with state-of-the-art methods and our proposal to compare visually the results. 

\subsection{Synthetic data}\label{sec:reg:exp:synthetic}

\begin{figure}[H]
\centering

\includegraphics[width=0.8\textwidth]{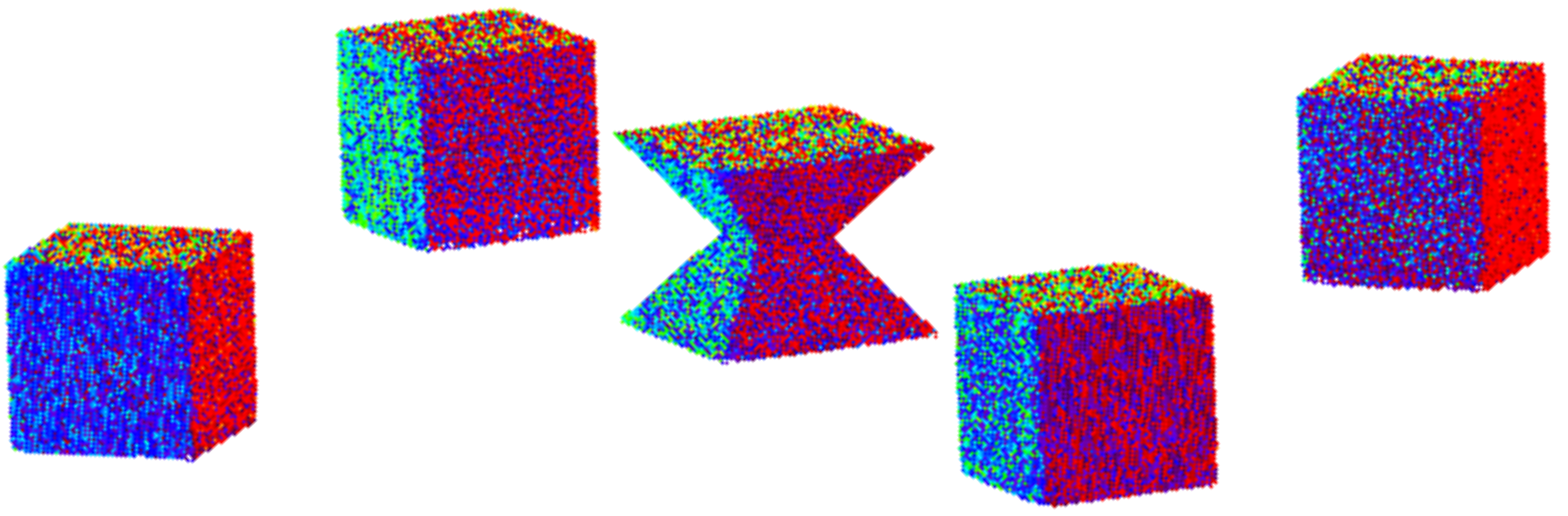}
\caption{Scene representative of the set up. The markers are cubes around the target object (a double pyramid).}
\label{fig:exp:scene}
\end{figure}

\begin{figure}[!h]
\centering

\includegraphics[height = 3cm]{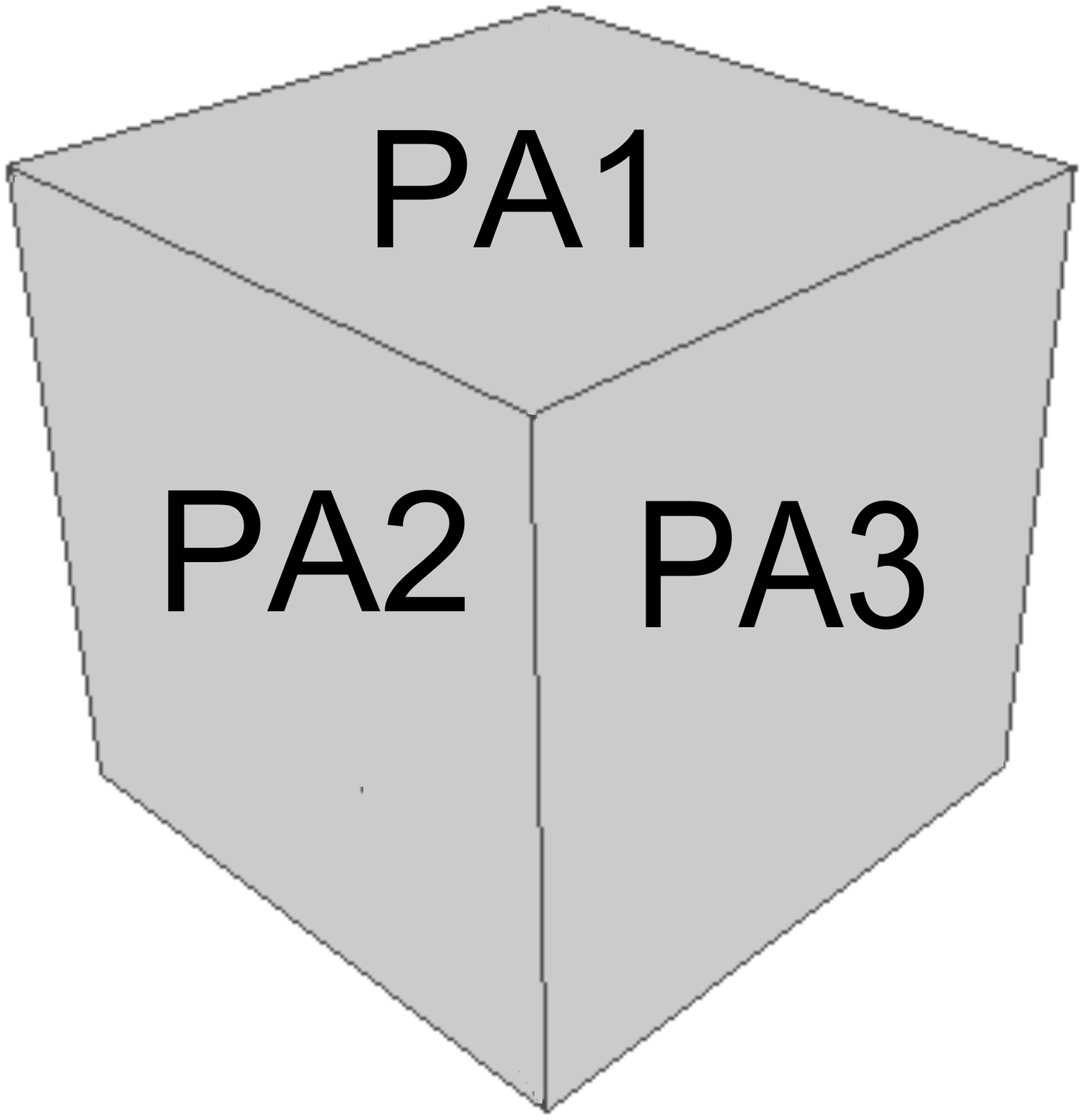} \includegraphics[height = 3cm]{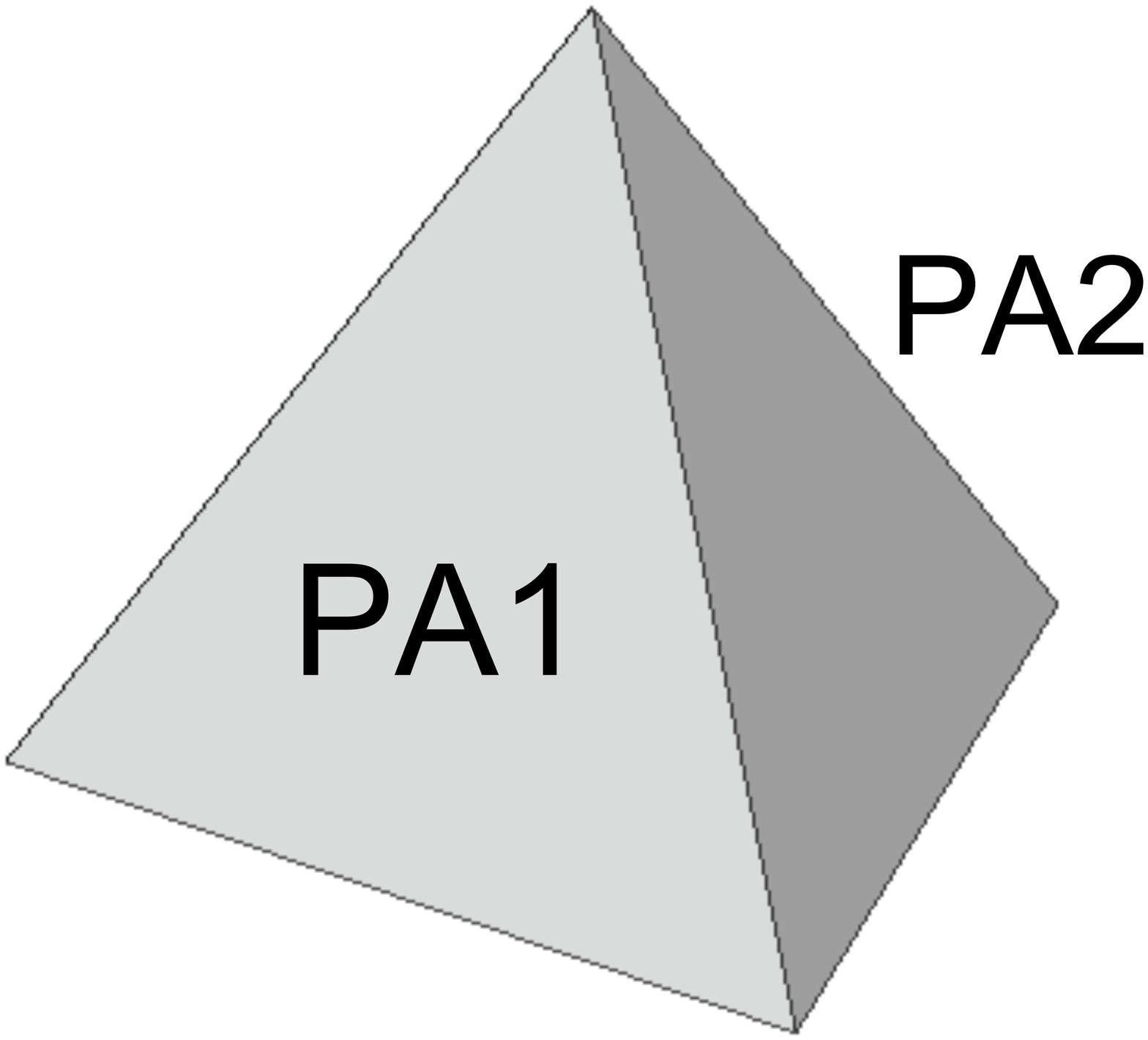} \includegraphics[height = 3cm]{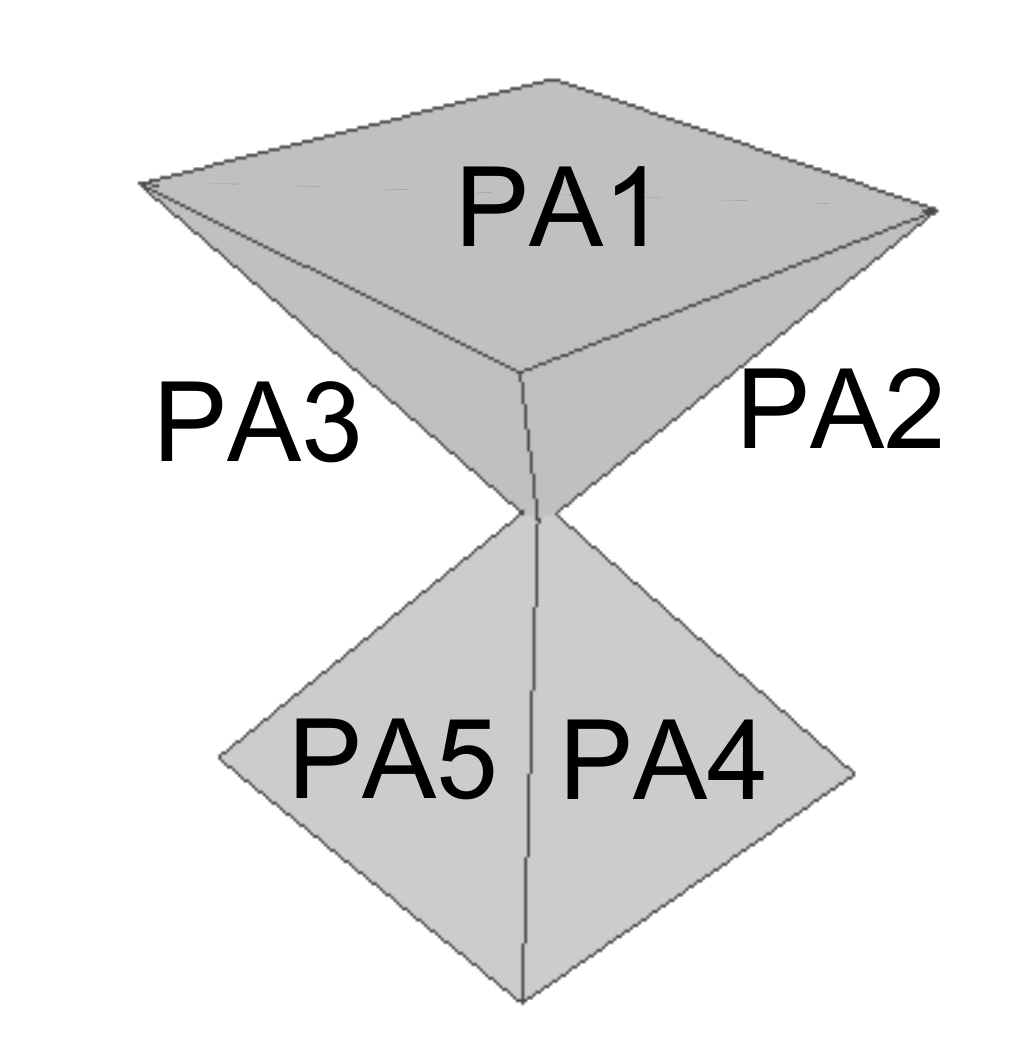}
\caption{Object models created in Blender a target objects. A cube, pyramid, and double pyramid.}
\label{fig:exp:model}

\end{figure}

As we said before, the synthetic set up has been created  using the Blender graphic software that allows to generate noise free objects. To simulate the RGB-D sensor, the Blensor \citep{Gschwandtner2011} plug-in was used. Since the original scene is known, it is going to be used as ground truth. Figure~\ref{fig:exp:scene} shows one of the scenes create in Blender with four cube markers and a double pyramid as target object. Specifically, in this experimentation four cubes are used as markers for all the scenes. Three different shapes are used as target objects to be reconstructed: a cube, a pyramid, and a double pyramid (this shape has one pyramid inverted joined with another pyramid by their tips). The models are presented in Figure~\ref{fig:exp:model}. To evaluate the method against noise effects, Gaussian noise has been added to the points.

\begin{figure}[H]
\centering

%

\begin{tabular}{c c c c}
A & B & C & D
\\
\includegraphics[width=0.2\textwidth]{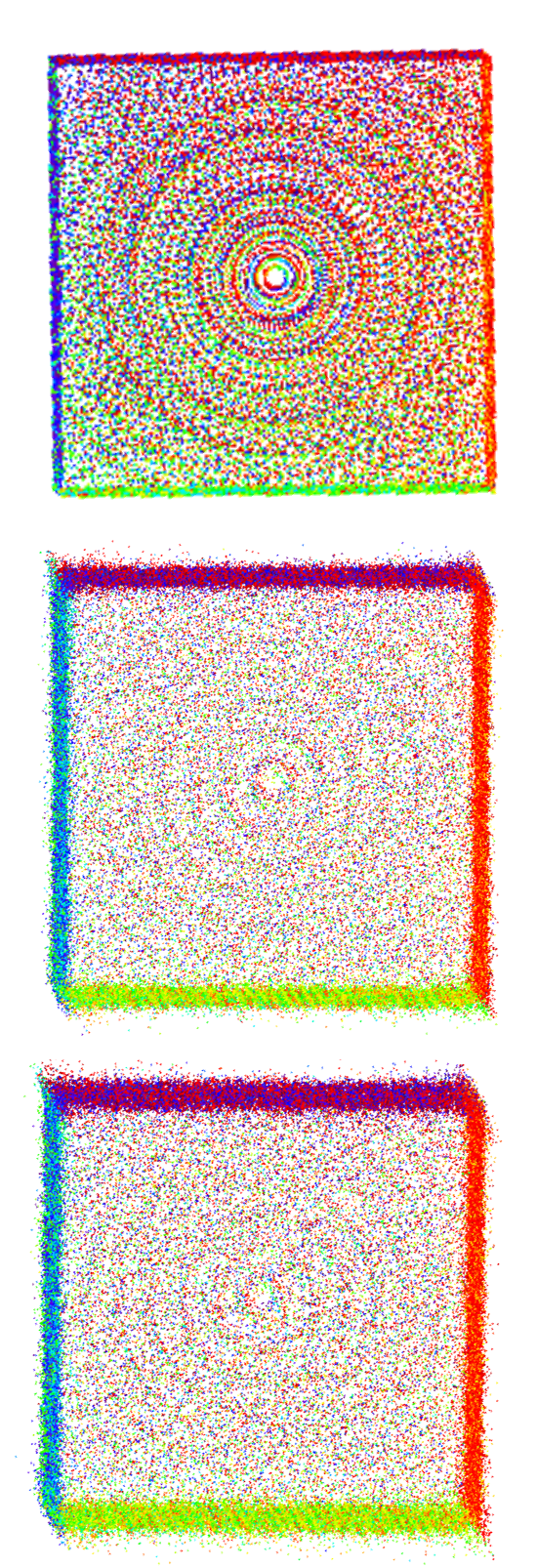} 
&
\includegraphics[width=0.2\textwidth]{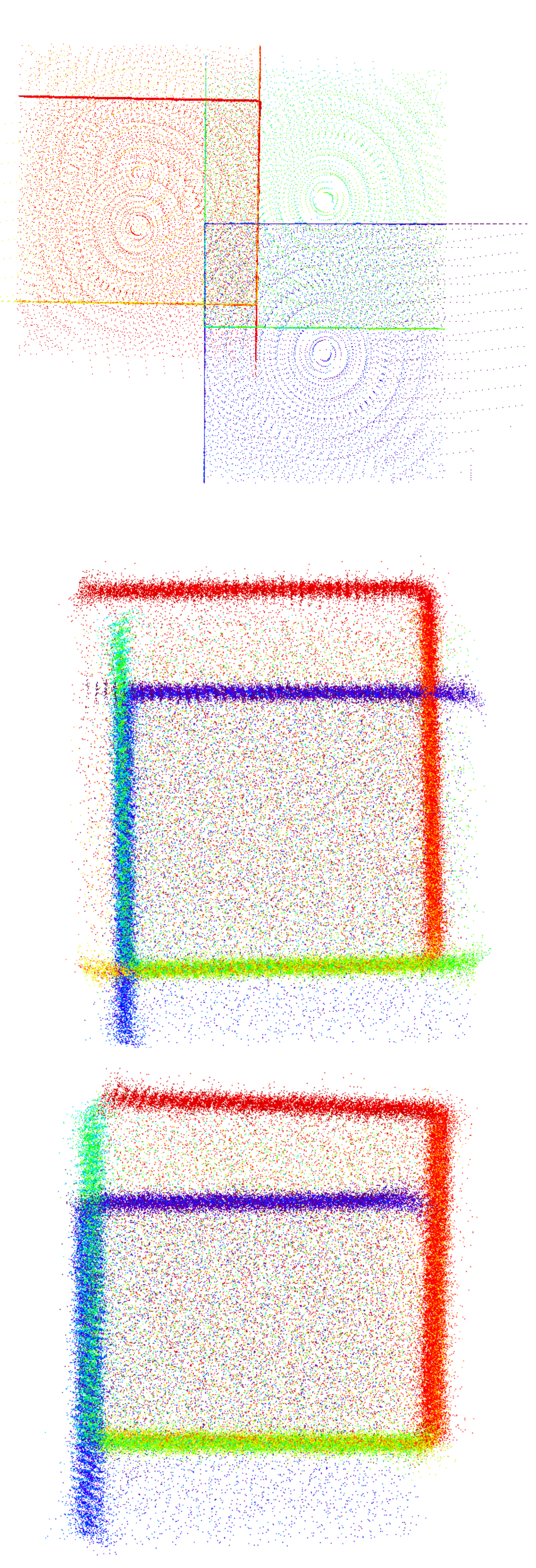}
&
\includegraphics[width=0.2\textwidth]{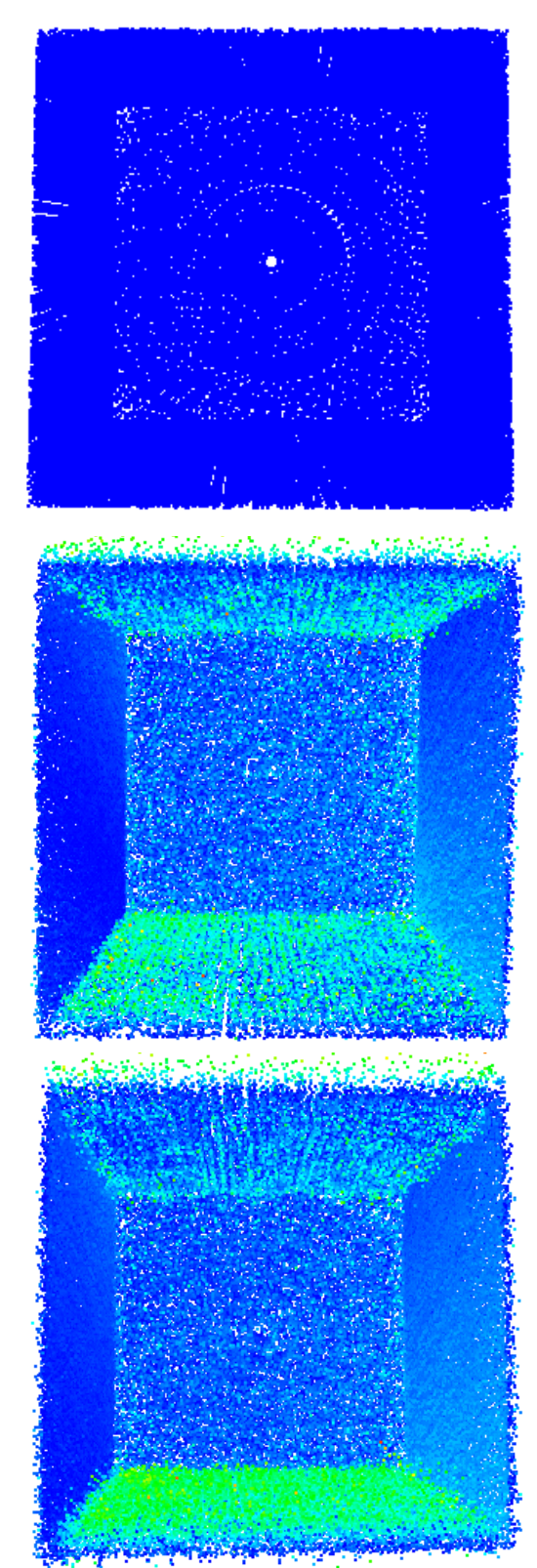}
&
\includegraphics[width=0.2\textwidth]{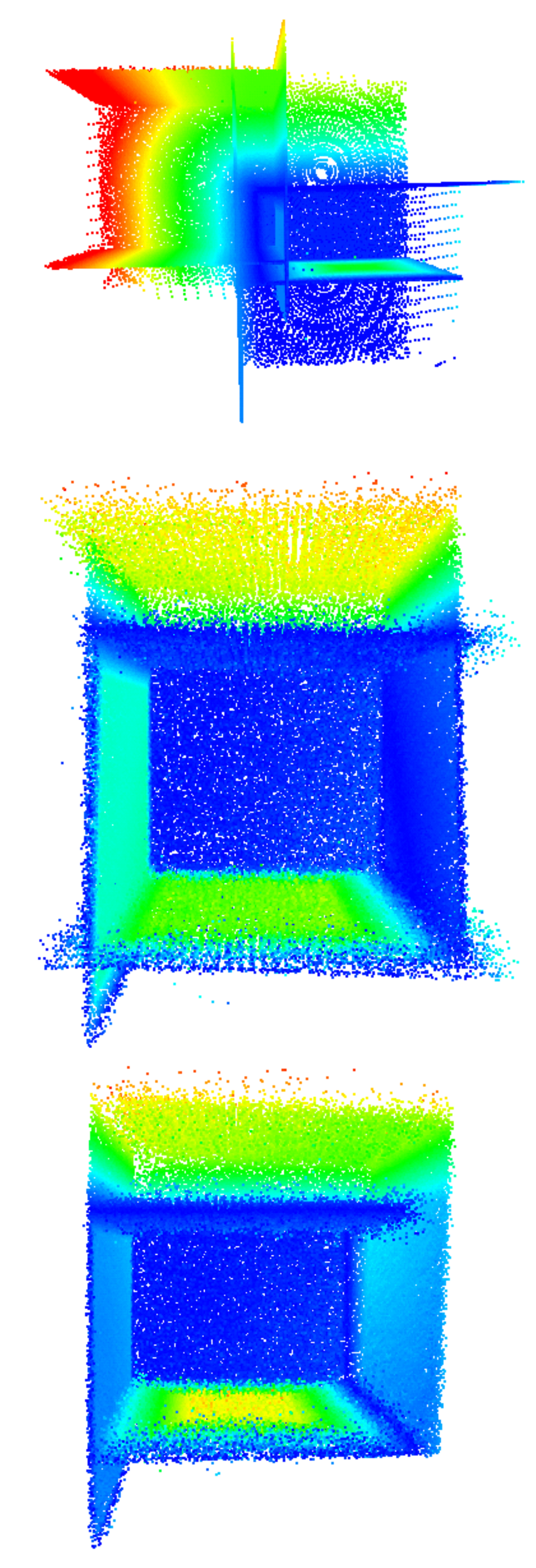}
\end{tabular}
\caption{Cube results. The first column shows the proposed method, and the second the ICP. The rows have 0, $4\cdot10^{-6}$, and $6\cdot10^{-6}$ levels of noise ($\sigma$).}
\label{fig:exp:res_cube}
\end{figure}

\begin{figure}[H]

\centering

%

\begin{tabular}{c c c c}
A & B & C & D
\\
\includegraphics[width=0.2\textwidth]{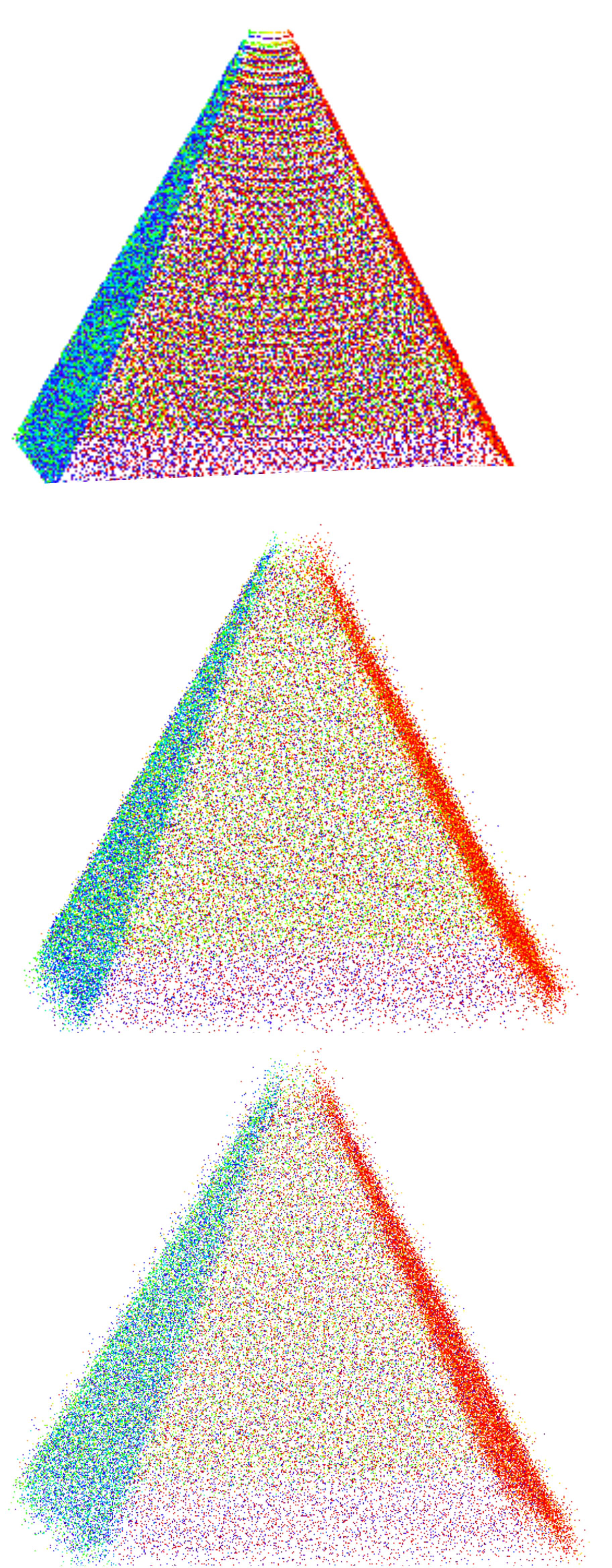} 
&
\includegraphics[width=0.2\textwidth]{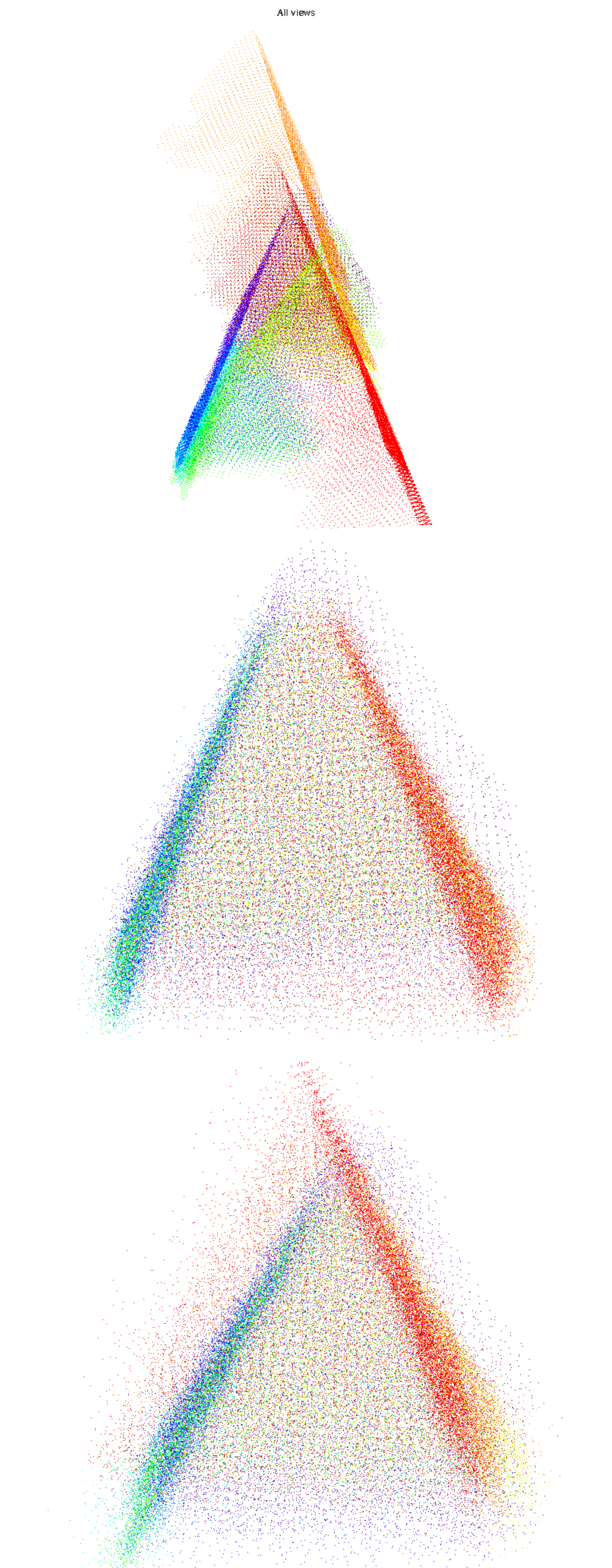}
&
\includegraphics[width=0.2\textwidth]{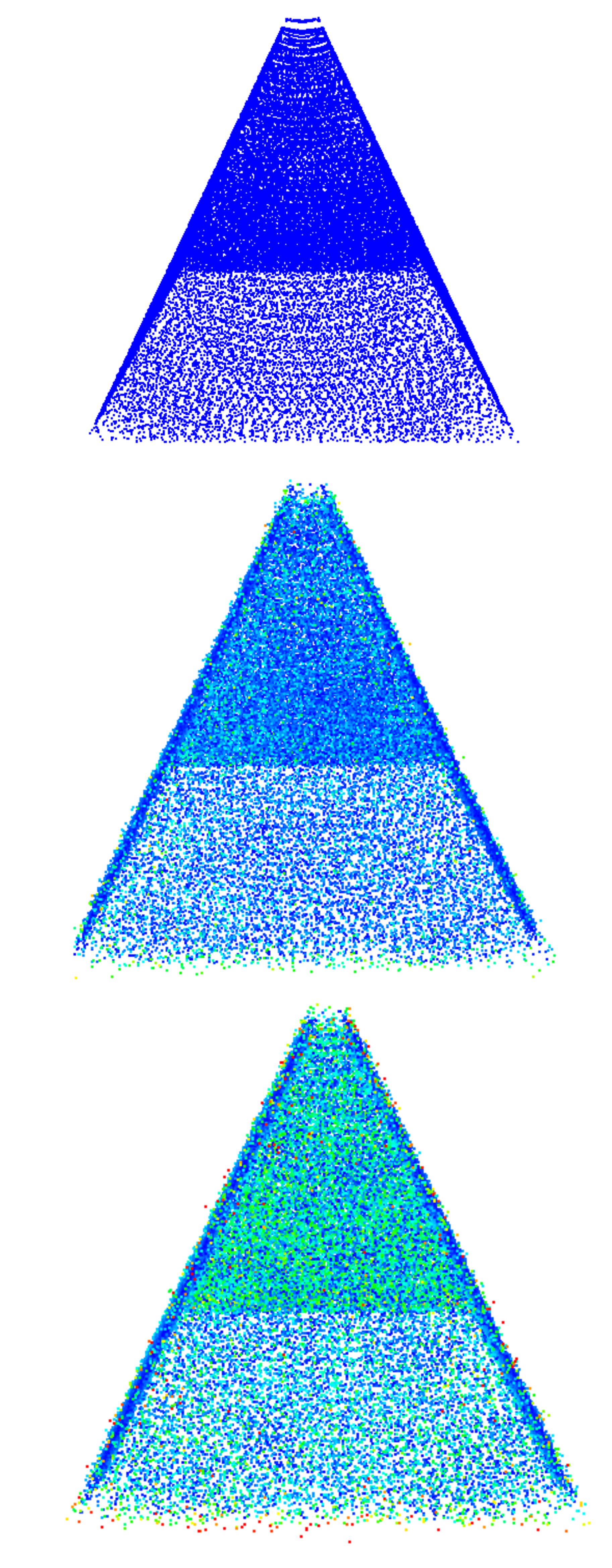} 
&
\includegraphics[width=0.2\textwidth]{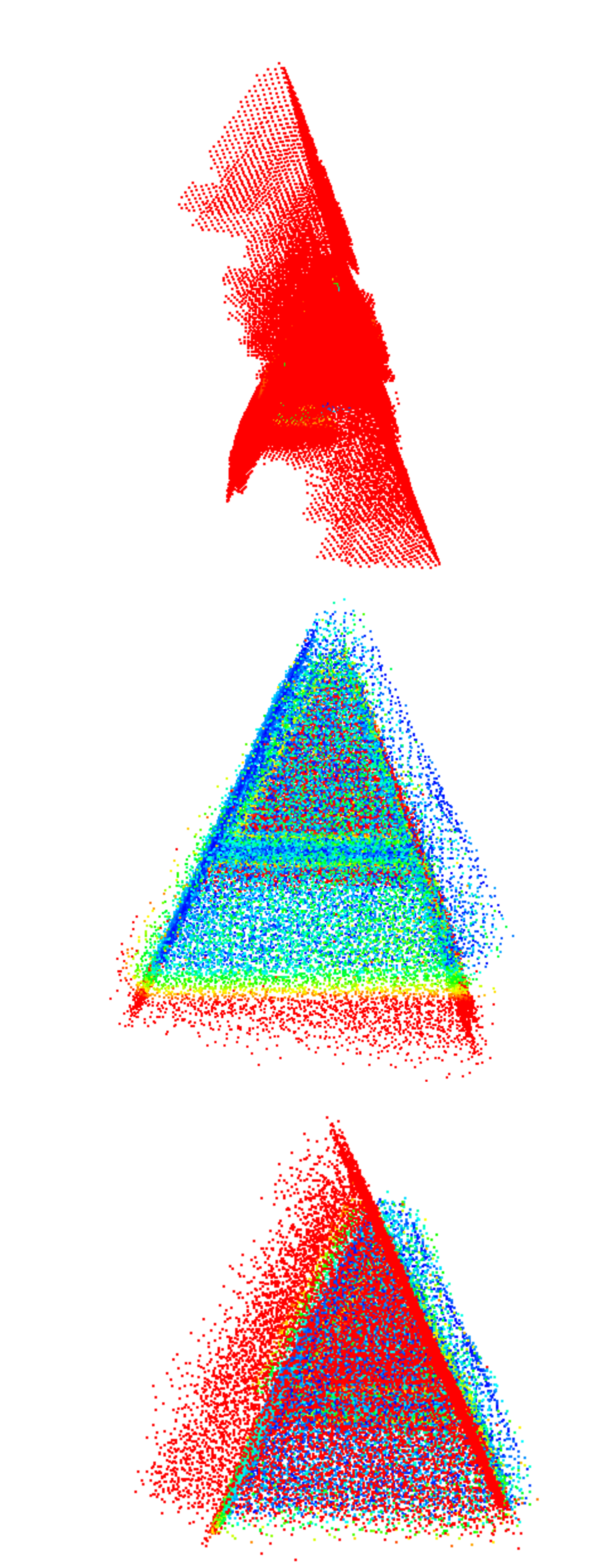}
\end{tabular}
\caption{Pyramid results. The first column shows the proposed method, and the second the ICP. The rows have 0, $4\cdot10^{-6}$, and $6\cdot10^{-6}$ levels of noise ($\sigma$).}
\label{fig:exp:res_pyr}
\end{figure}

\begin{figure}[!ht]
\centering

%

\begin{tabular}{c c c c}
A & B & C & D
\\
\includegraphics[width=0.2\textwidth , height = 10cm]{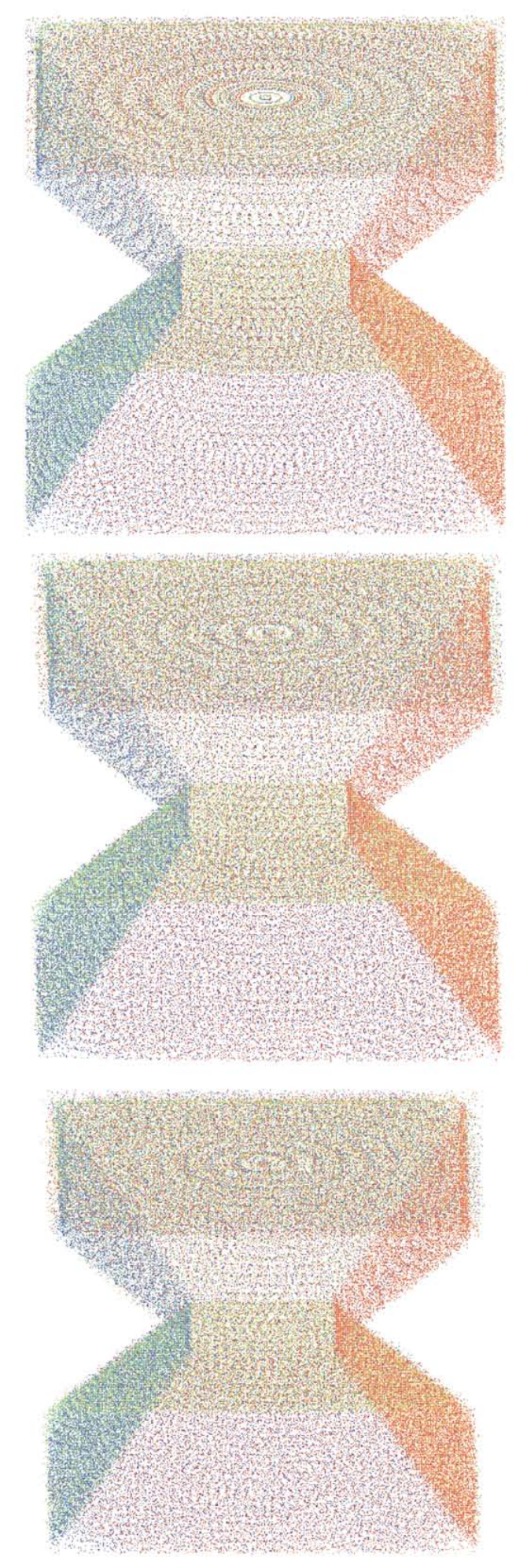} 
&
\includegraphics[width=0.2\textwidth , height = 10cm]{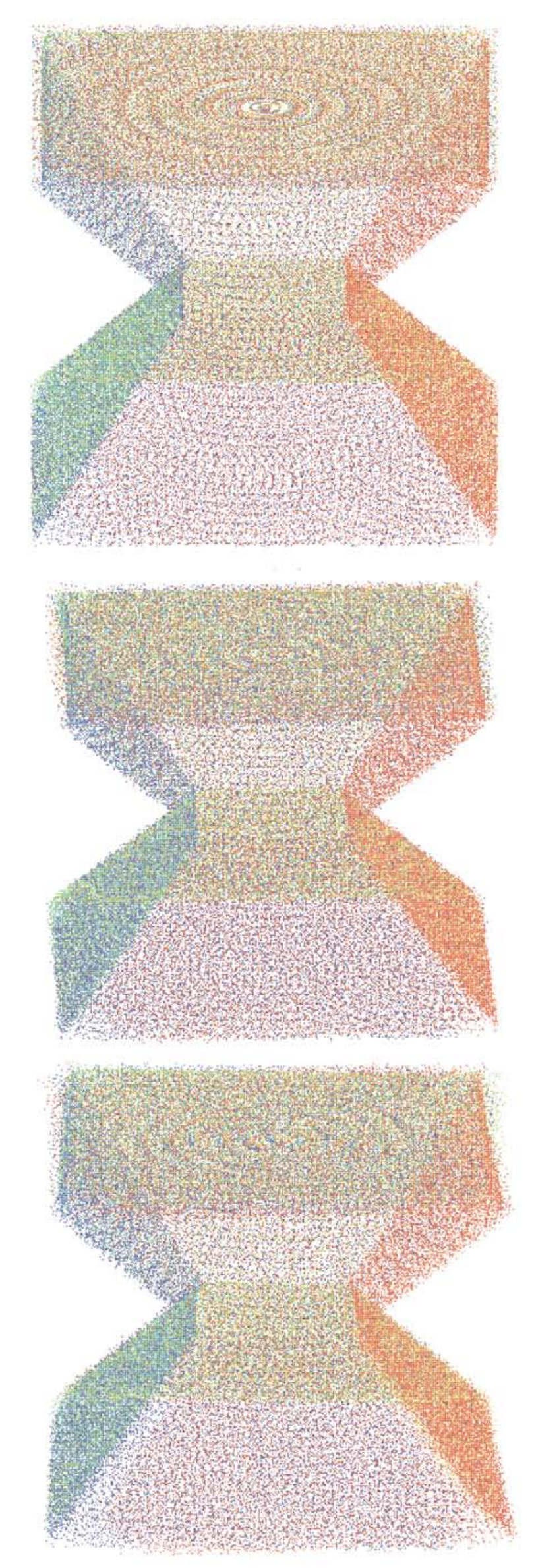}
&
\includegraphics[width=0.2\textwidth , height = 10cm]{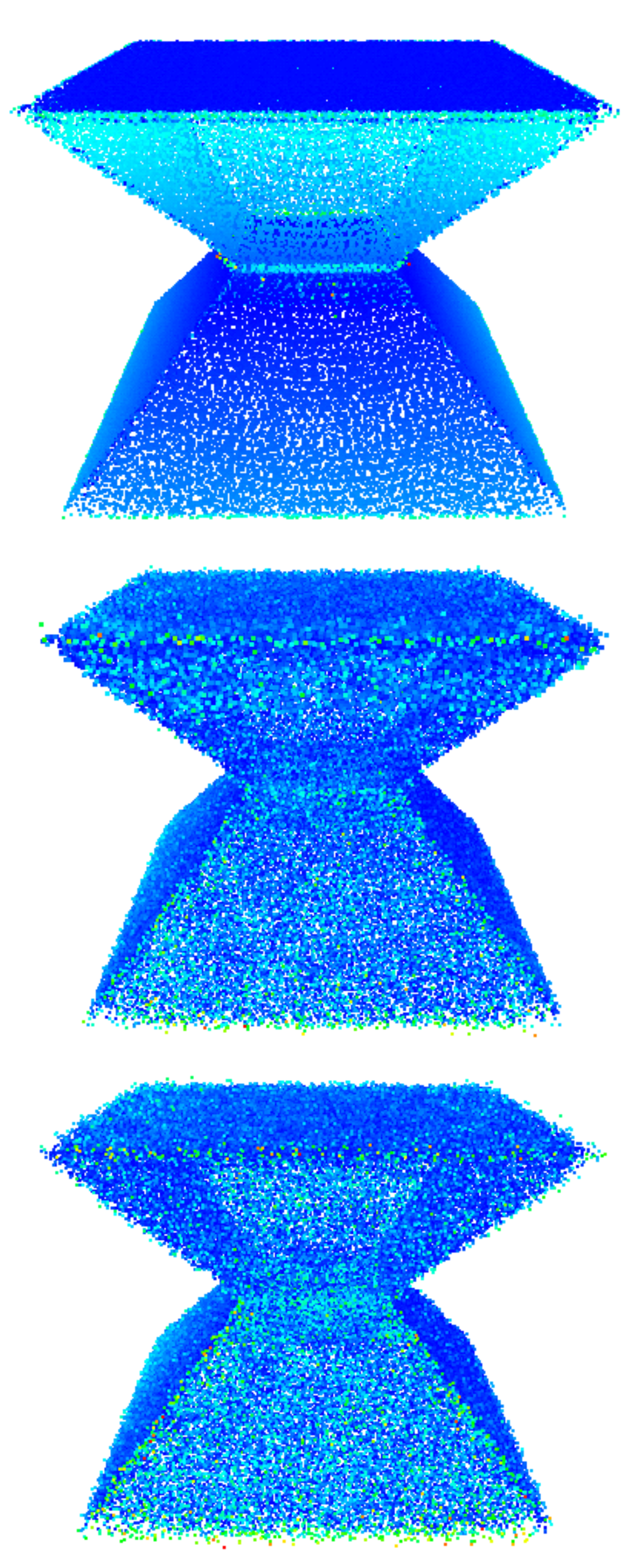} 
&
\includegraphics[width=0.2\textwidth , height = 10cm]{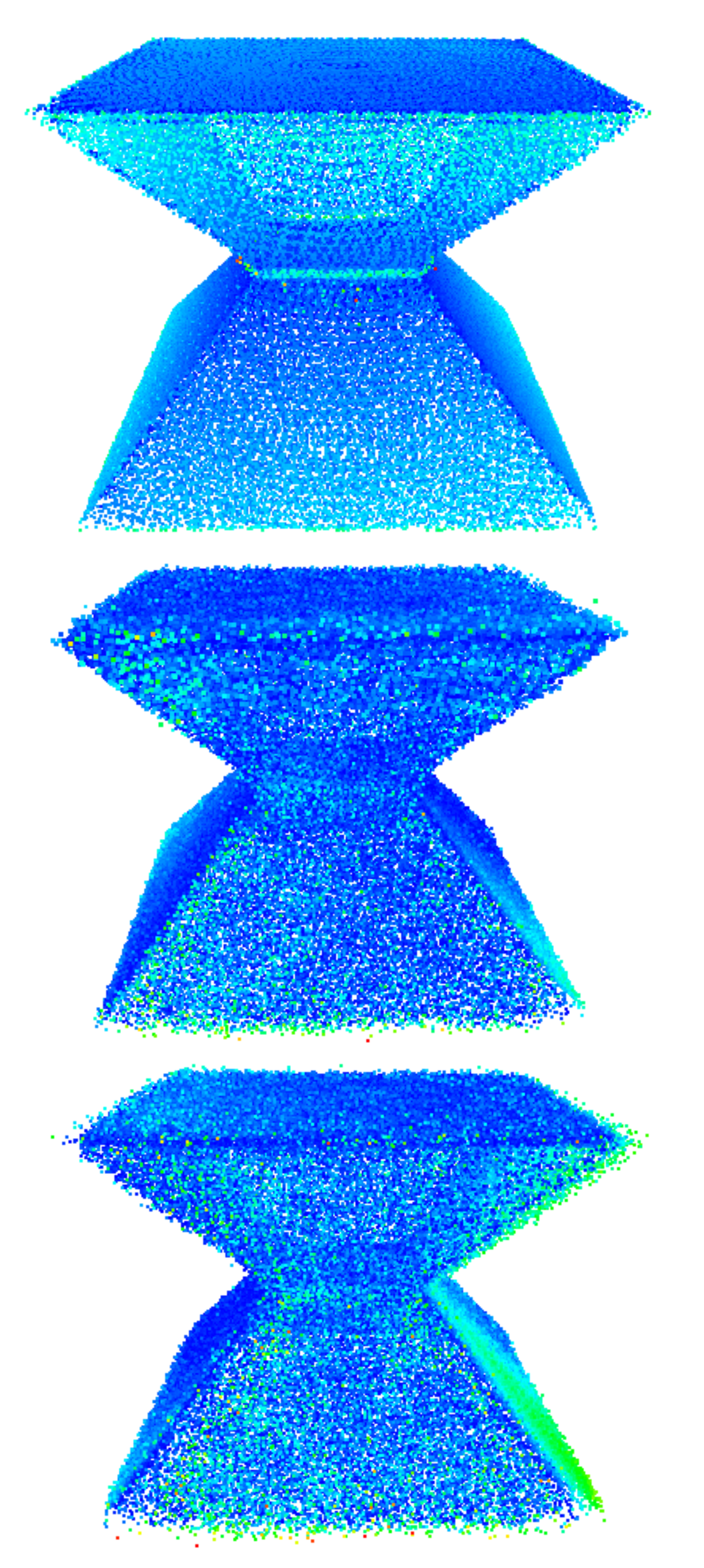}

\end{tabular}
\caption{Double pyramid results. The first column shows the proposed method, and the second the ICP. The rows have 0, $4\cdot10^{-6}$, and $6\cdot10^{-6}$ levels of noise ($\sigma$).}
\label{fig:exp:res_dpyr}
\end{figure}

The proposed algorithm is evaluated qualitatively by visual inspection Figures~\ref{fig:exp:res_cube},~\ref{fig:exp:res_pyr},~\ref{fig:exp:res_dpyr} and quantitatively by the Hausdorff distance. This measurement calculates the distance between two sets of space data. As we have the original model, we can apply this distance between each registration result against the original shape. It is necessary to be aware that all results have been finely aligned to the model in order to provide a faithful measurement. The Haussdorf evaluation allows to compare the proposed method and the ICP visually (columns C and D of the Figures~\ref{fig:exp:res_cube},~\ref{fig:exp:res_pyr},~\ref{fig:exp:res_dpyr}) and numerically (Table~\ref{tab:exp:res_hausdorff}).

Moreover, a comparison with the well-know ICP is also done due to this one of the most used registration methods for accurate registration. Several variants of this method are present in the literature. For this comparison, our ICP implementation uses point-to-plane distances, worst matches rejection and boundaries information as a variant conceived to deal with noisy data. 

Figures~\ref{fig:exp:res_cube}, \ref{fig:exp:res_pyr} and \ref{fig:exp:res_dpyr} show the registration result of the $\mu~-MAR$ proposed method (column A) and the ICP variant (column B) for different levels of noise. From the top to bottom, the levels of Gaussian noise are zero mean and sigma ($\sigma$) equal to 0, $4\cdot10^{-6}$ and $6\cdot10^{-6}$. Additionally, they include the Hausdorff distances for our method (column C) and for the ICP (column D). They are represented in colours where blues are the lowest distances and reds the largest. 

The cube has been registered in Figure~\ref{fig:exp:res_cube}. ICP clearly has wrong results because, when only two planes are registered (frontal and top), the registration tends to slide to the side. However, the $\mu$-MAR method properly registers the cubes for the different levels of noise. Figure~\ref{fig:exp:res_pyr} presents the pyramid registration, with a similar problem than before. The pyramid only has two visible sides, that is the reason for the wrong registration in ICP whereas the proposed method provides a high accurate registration. Lastly, Figure~\ref{fig:exp:res_dpyr} shows an interesting result. Due to the number of planes is higher, hence the geometry is more detailed, the registration of ICP works similar to the proposed method.

Table~\ref{tab:exp:res_hausdorff} contains the Hausdorff distance numerical values of the registration results (corresponding to the C and D columns in the Figures~\ref{fig:exp:res_cube}, \ref{fig:exp:res_pyr} and \ref{fig:exp:res_dpyr}). The values $Min$ and $Max$ correspond to the minimum and maximum values found between each registration result and the ground truth model. These are dependant on outliers apart from the registration result. The values used to study the performance are the $Mean$ and the Root Mean Square error ($RMS$) since they are global values instead of noise dependant values. As Table~\ref{tab:exp:res_hausdorff} shows, the proposed method achieves better results for all situations. 

\begin{figure}[!t]
\centering

\includegraphics[width=0.8\textwidth , height = 6cm]{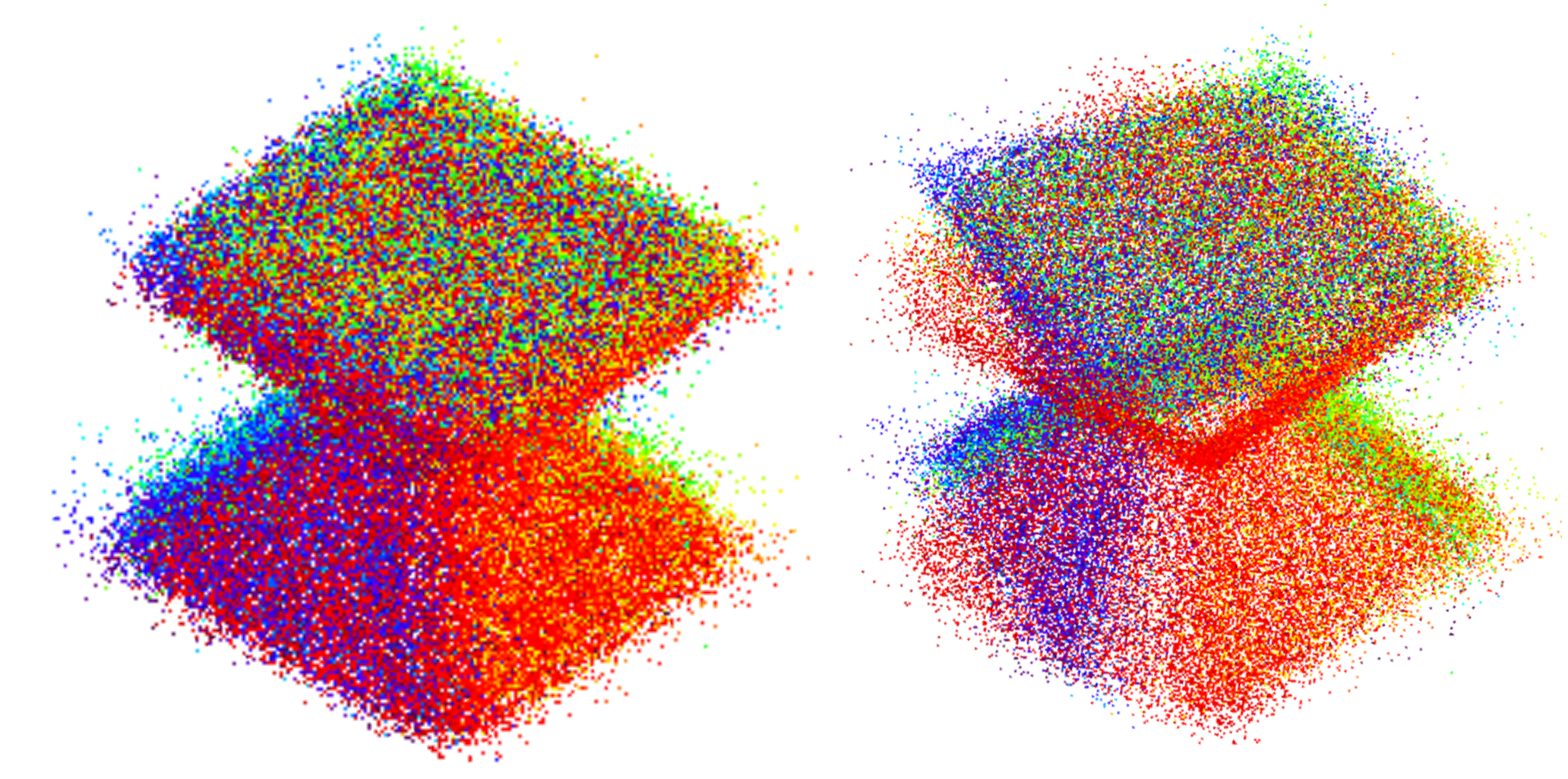}
\caption{Double pyramid result for $\sigma = 4\cdot10^{-6}$. In this situation the ICP (right column) cannot return a proper result while the propose method (left side) still has a good performance.}
\label{fig:exp:res_dpyr_break}
\end{figure}

The experiments presented in this section show the proper performing of the $\mu$-MAR obtaining a registration of the different views with high accuracy. The evaluation against the ICP variant conceived to deal with noise (point-to-plane, boundaries information and worst matching rejection) demonstrates that the proposal of registering using models reduce the effects of noise, even in situations of high noise. Moreover, a lack of geometry produces ICP to fail, whereas the proposed method still remains stable. However, even the double pyramid , which has a very varied geometry, ICP does not result in a proper registration, while the proposed method achieves good results. This effect is visually represented in a extreme situation in Figure~\ref{fig:exp:res_dpyr_break}. We evaluated the double pyramid with a higher level of noise. The ICP could not achieve a good result with sigma equal to $4\cdot10^{-5}$ (note that before it was $4\cdot10^{-6}$ and now is $4\cdot10^{-5}$). The proposed method still outperforms the ICP for this case. 

\begin{table}[!h]
  \centering
      \caption{Hausdorff distances for both methods. The values of Min and Max represent the minimum and maximum distance found between model and registration result. The RMS shows the Root Mean Square error.}
    \begin{tabular}{c | c | c | rrrr}
    \hline
    Method & Object & Noise($\sigma$) & Min   & Max   & Mean  & RMS \\
    \hline
      &                    & $0$             & 0       & 0,5912 & 0,1496 & 0,2210 \\
      &      Cube          & $4\cdot10^{-6}$ & 0       & 8,2982 & 0,7372 & 1,0040 \\
      &                    & $6\cdot10^{-6}$ & 0       & 11,1011 & 1,1635 & 1,5896 \\
      &                    & $0$             & 0,14179 & 0,4980 & 0,3214 & 0,3290 \\
    $\mu$-MAR              & Pyramid   & $4\cdot10^{-6}$ & 0       & 9,3453 & 0,7026 & 1,0153 \\
      &                    & $6\cdot10^{-6}$ & 0       & 7,1646 & 1,1254 & 1,5314 \\
      &                    & $0$             & 0       & 2,9296 & 0,2480 & 0,3435 \\
      &     DoublePyramid  & $4\cdot10^{-6}$ & 0       & 9,1900 & 0,6014 & 0,8427 \\
      &                    & $6\cdot10^{-6}$ & 0       & 9,9254 & 0,8513 & 1,1854 \\
      \hline
      &                    & $0$             & 0       & 68,1055 & 18,3129 & 25,7950 \\
      & Cube               & $4\cdot10^{-6}$ & 0       & 31,1023 & 7,0463 & 10,3379 \\
      &                    & $6\cdot10^{-6}$ & 0,00003 & 34,0622 & 7,8185 & 11,0435 \\
      &                    & $0$             & 0,02808 & 20,5320 & 14,4189 & 15,2855 \\
    ICP &  Pyramid         & $4\cdot10^{-6}$ & 0,00003 & 8,1125  & 2,3703 & 3,2943 \\
      &                    & $6\cdot10^{-6}$ & 0,00008 & 9,3999  & 3,2738 & 4,4372 \\
      &                    & $0$             & 0       & 2,9196  & 0,3329 & 0,3876 \\
      &  DoublePyramid     & $4\cdot10^{-6}$ & 0       & 9,8798  & 0,7739 & 1,0644 \\
      &                    & $6\cdot10^{-6}$ & 0       & 9,9841  & 1,0157 & 1,3933 \\
    \hline
    \end{tabular}%
  	\label{tab:exp:res_hausdorff}%
\end{table}%

\subsection{Real data}\label{sec:reg:exp:real}

In order to evaluate the $\mu$-MAR in real situations, a comparison between different methods has been carried out. The evaluation compares six methods including ICP, RANSAC-based method, a combination of ICP + RANSAC-based, RGBDemo, KinectFusion, and the proposed method.

ICP is highly dependant on the geometry of the object as it uses only the points in the space to register. RANSAC-based uses visual features (SIFT in this case) to establish correspondences and then estimate the best alignment between them. RGBDemo~\citep{Kramer2012} is a registration method for RGB-D sensors that uses ARToolkit markers for a coarse registration and then ICP for finely align the views. KinectFusion~\citep{Izadi2011, Newcombe2011} uses a ICP to estimate the transformation.

\begin{figure}[!h]
	\centering
	\caption{Objects used for experimentation. Taz toy (a), Cube (b), Tool (c) and Bomb toy (d).}
	\begin{subfigure}[b]{0.2\textwidth}
      	\includegraphics[width=\textwidth]{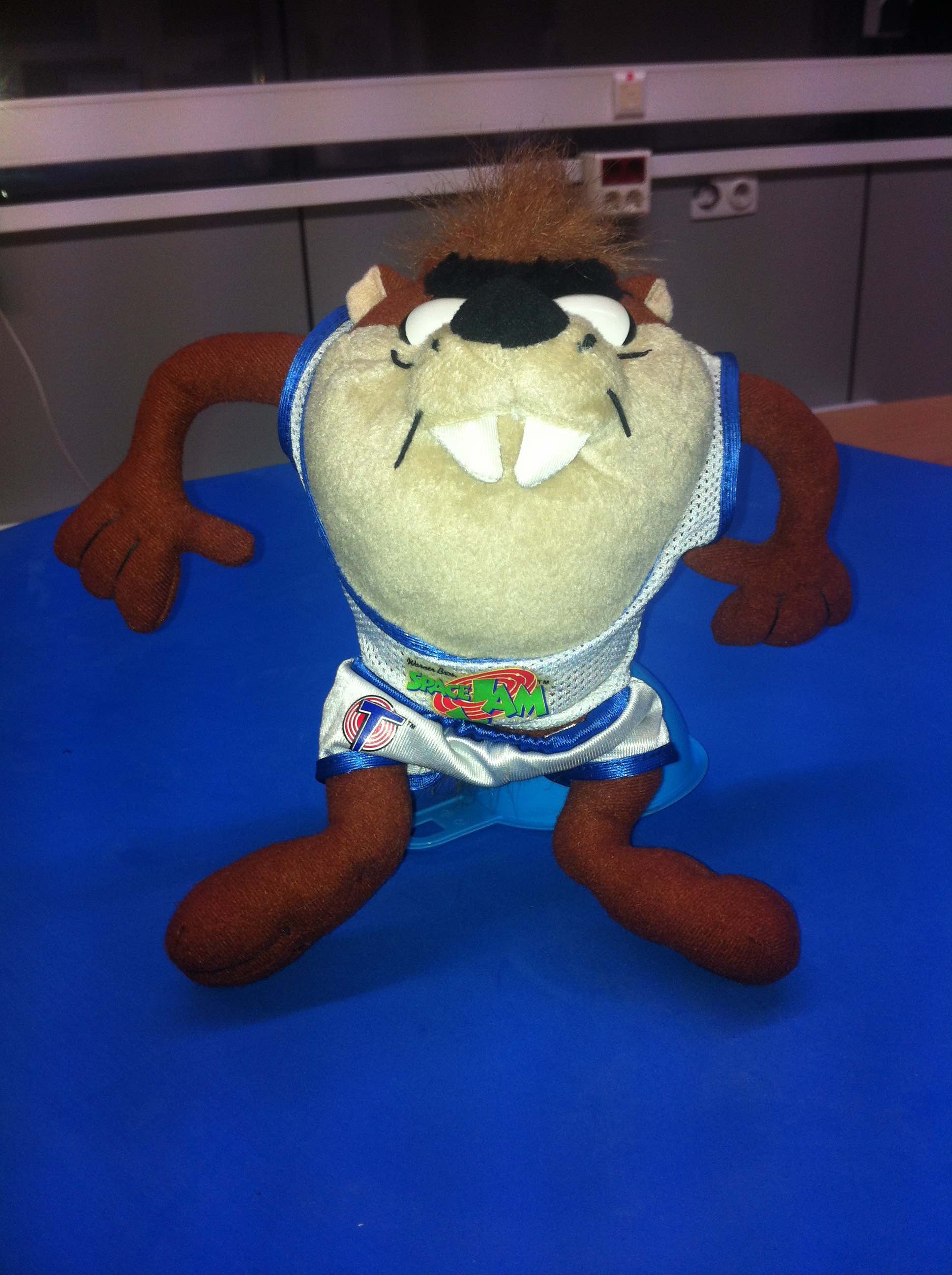} 
      	\caption{}
      	\label{fig:img:taz}
	\end{subfigure}
	\begin{subfigure}[b]{0.2\textwidth} 	
      	\includegraphics[width=\textwidth]{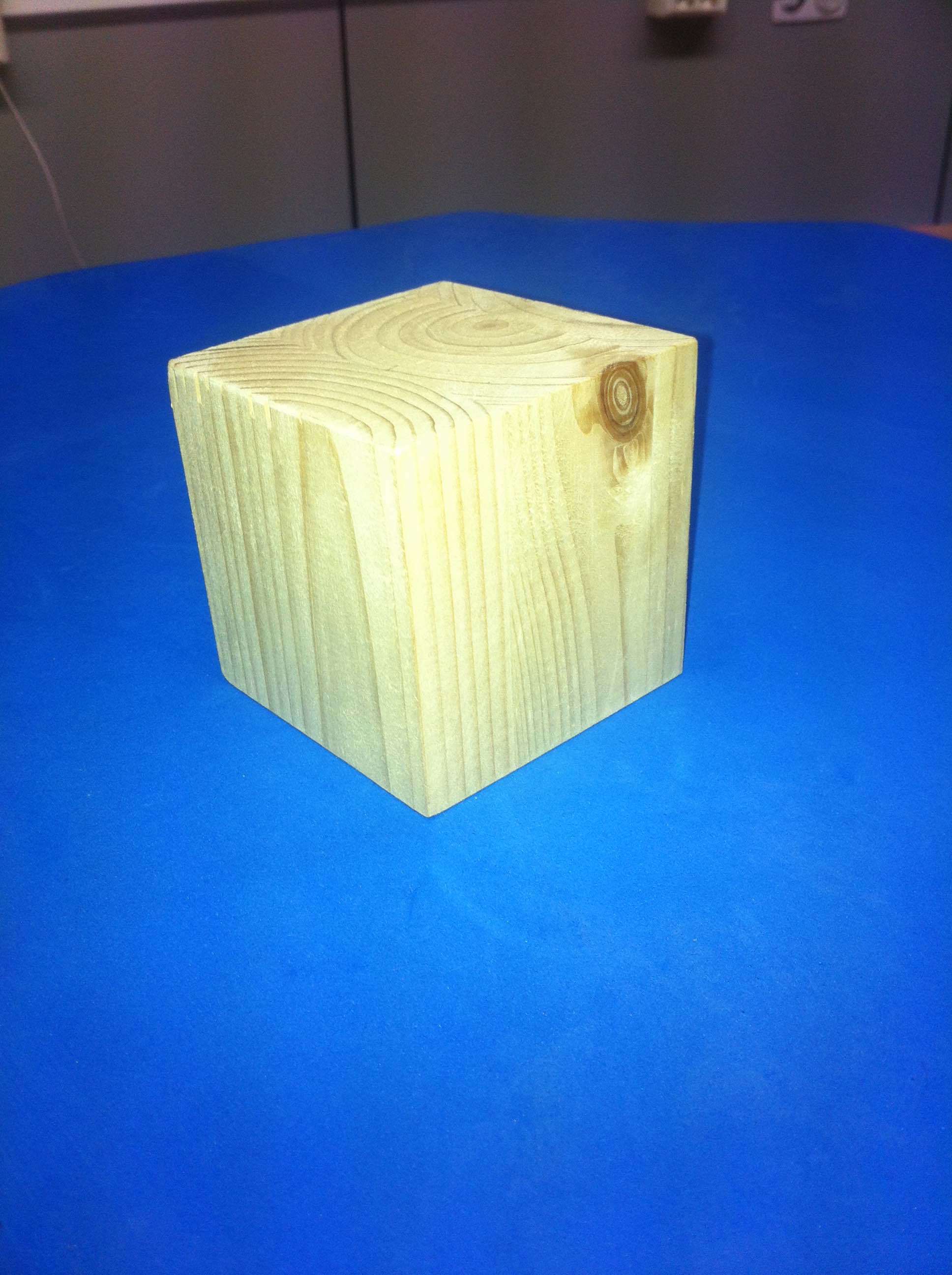}
      	\caption{}
      	\label{fig:img:cube}
	\end{subfigure}
	\begin{subfigure}[b]{0.2\textwidth} 	
      	\includegraphics[width=\textwidth]{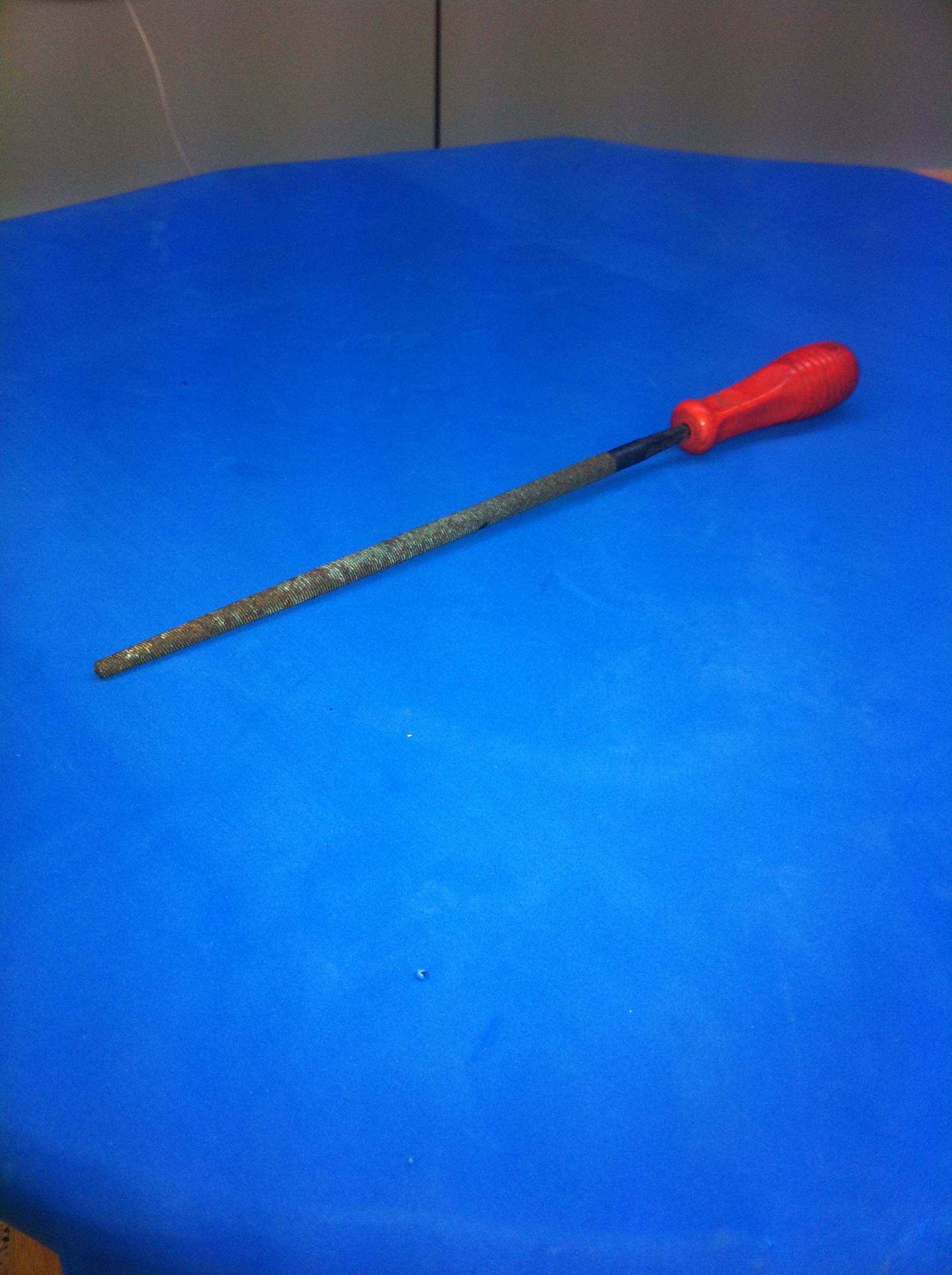}
      	\caption{}
      	\label{fig:img:tool}
	\end{subfigure}
	\begin{subfigure}[b]{0.2\textwidth} 	
      	\includegraphics[width=\textwidth]{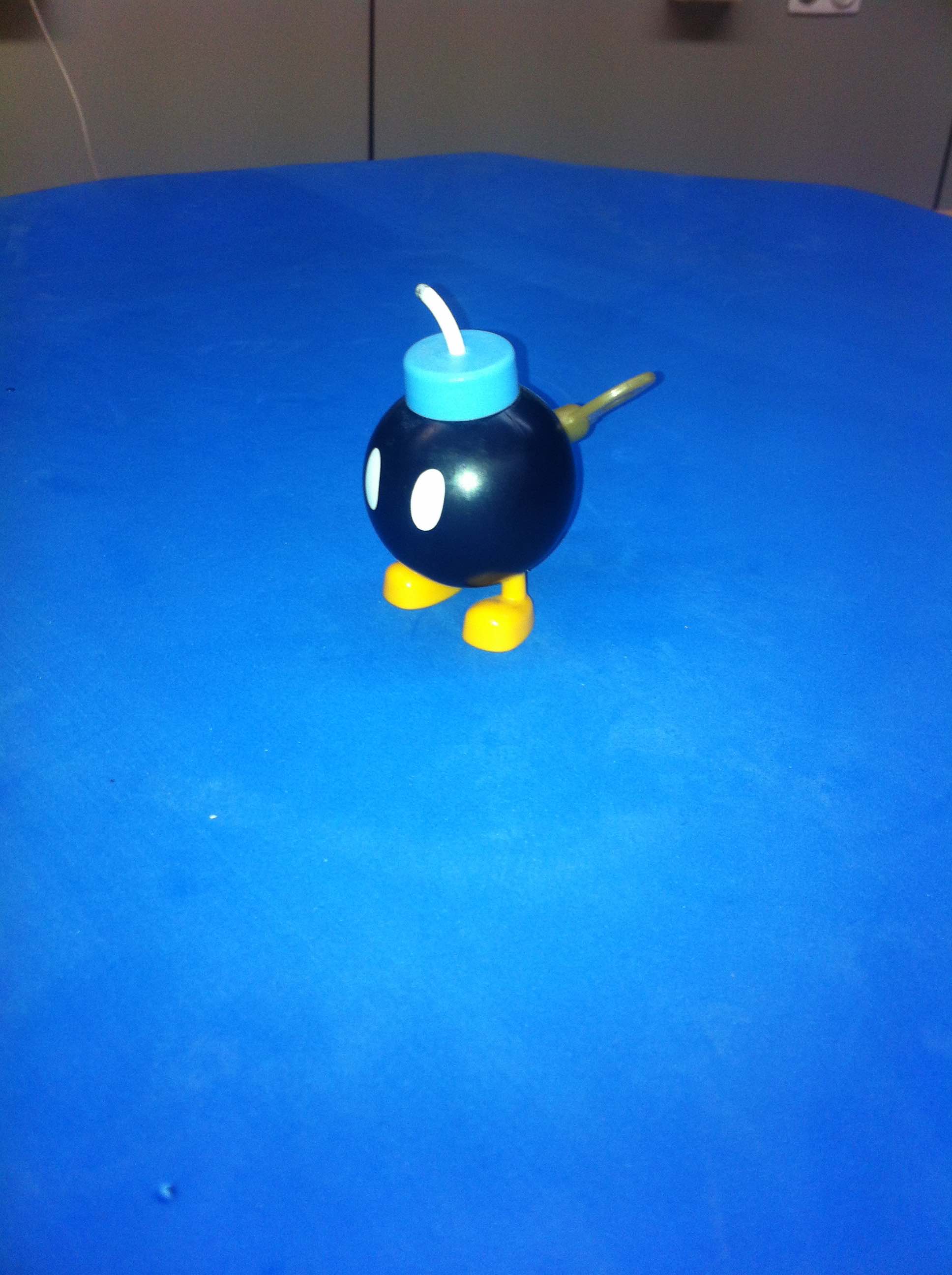}
      	\caption{}
      	\label{fig:img:bomb}
	\end{subfigure}	
	\label{fig:objects}
\end{figure}

Four different objects has been used for this comparison (Figure~\ref{fig:objects}). Each of them has specific features that will help to evaluate the specific characteristics of the methods.  The first one, Figure~\ref{fig:img:taz}, Taz toy is the largest object with variety of colours. The second object (Figure~\ref{fig:img:cube}) is a wooden cube. In the third column, a tool (Figure~\ref{fig:img:tool}) is presented. It is $30 cm$ long with different thinness. The last object is a Bomb toy shown in Figure~\ref{fig:img:bomb}. It has different colours, thin parts such as the white one on the top, the body part with smooth curve and the back part with a key attached. The same objects were used in our previous work in \citep{Morell2014}. This dataset was created because of there are no public datasets which include the markers we need for our proposal. 

They have been acquired using a Primesense Carmine on the turntable. In order to use the the fine registration of the ICP-based methods without a pre-alignment, 320 views were acquired of each element around them (about 1.13 degrees per step). Also, the camera is about a meter far from the target object, and placed with an angle to ensure to be able to view the markers of the RGBDemo.

\begin{figure}[H]
\centering   
\scalebox{0.8}{
    \begin{tabular}{c c c c c}
    \raisebox{2cm}{a)} &
    \includegraphics[width=0.2\textwidth]{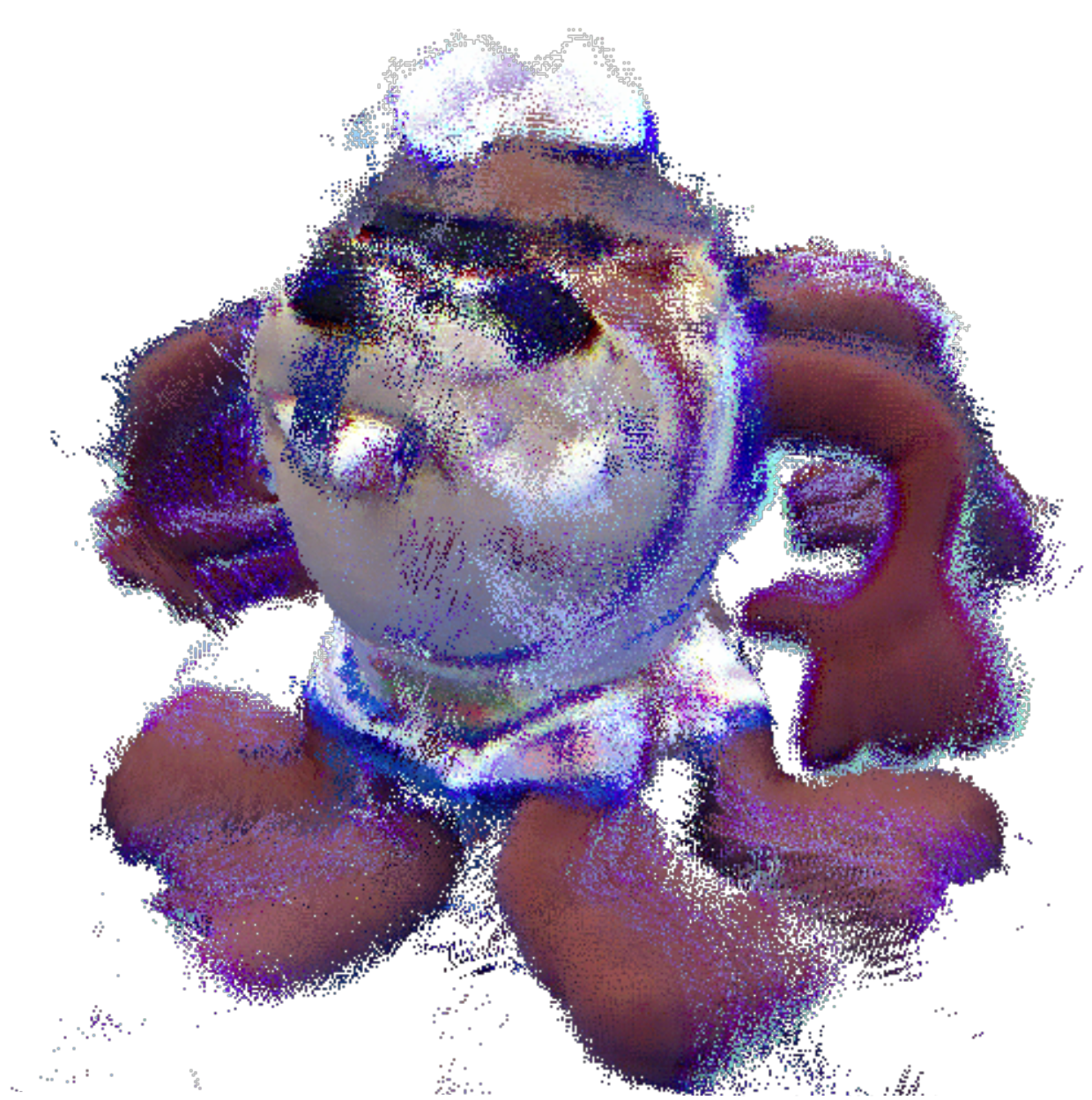} & \includegraphics[width=0.2\textwidth]{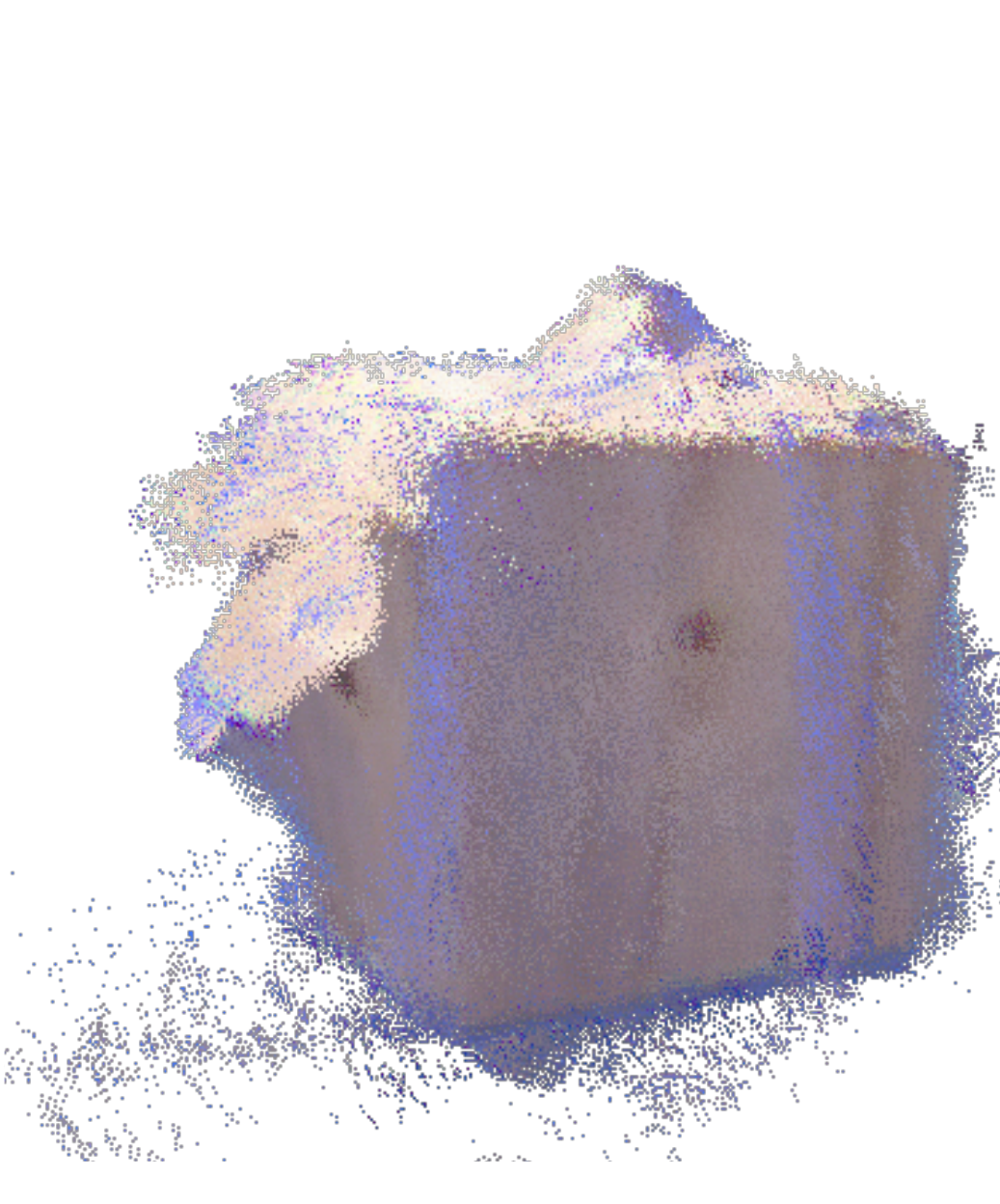} & \includegraphics[width=0.2\textwidth]{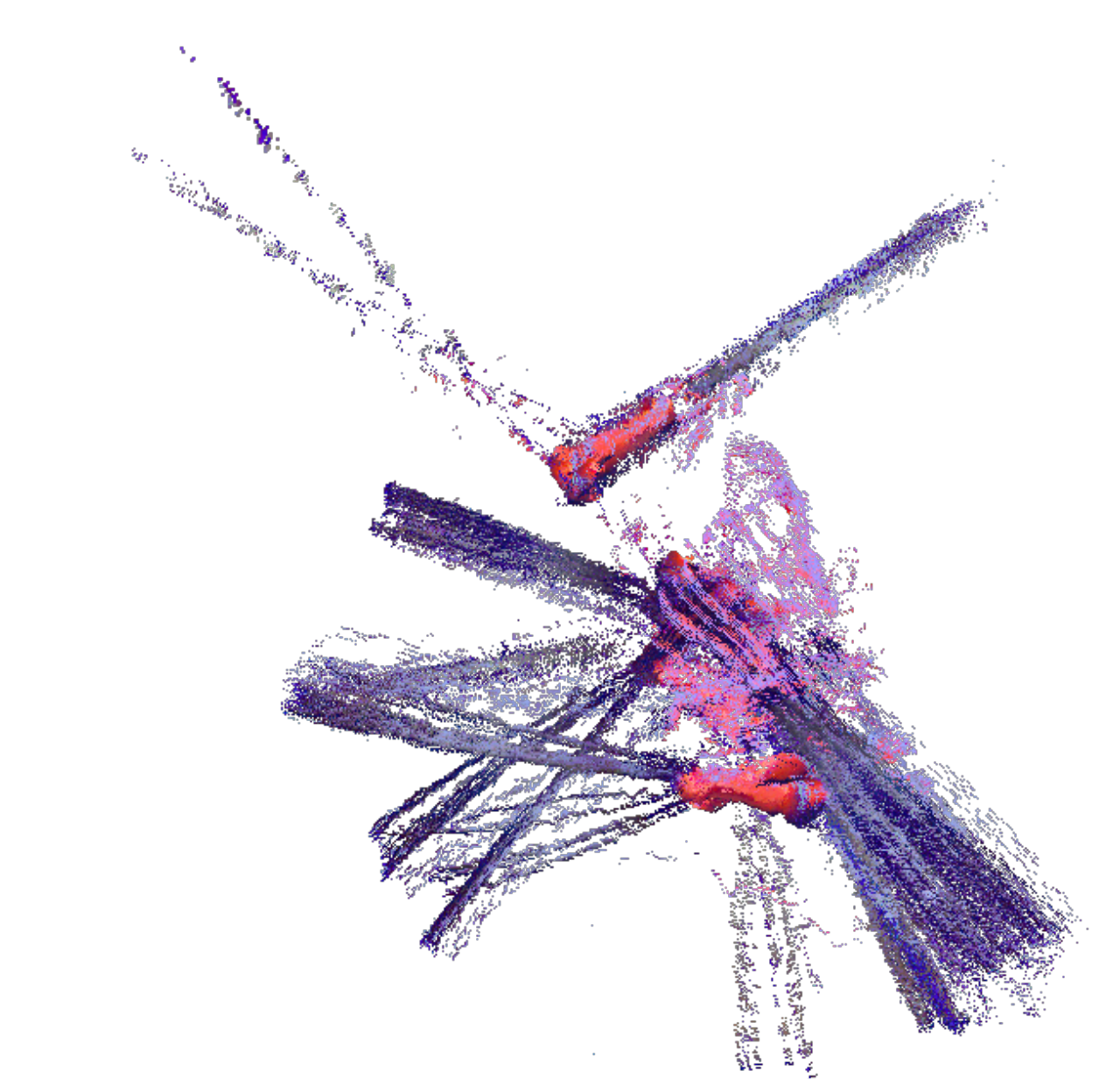} & \includegraphics[width=0.2\textwidth]{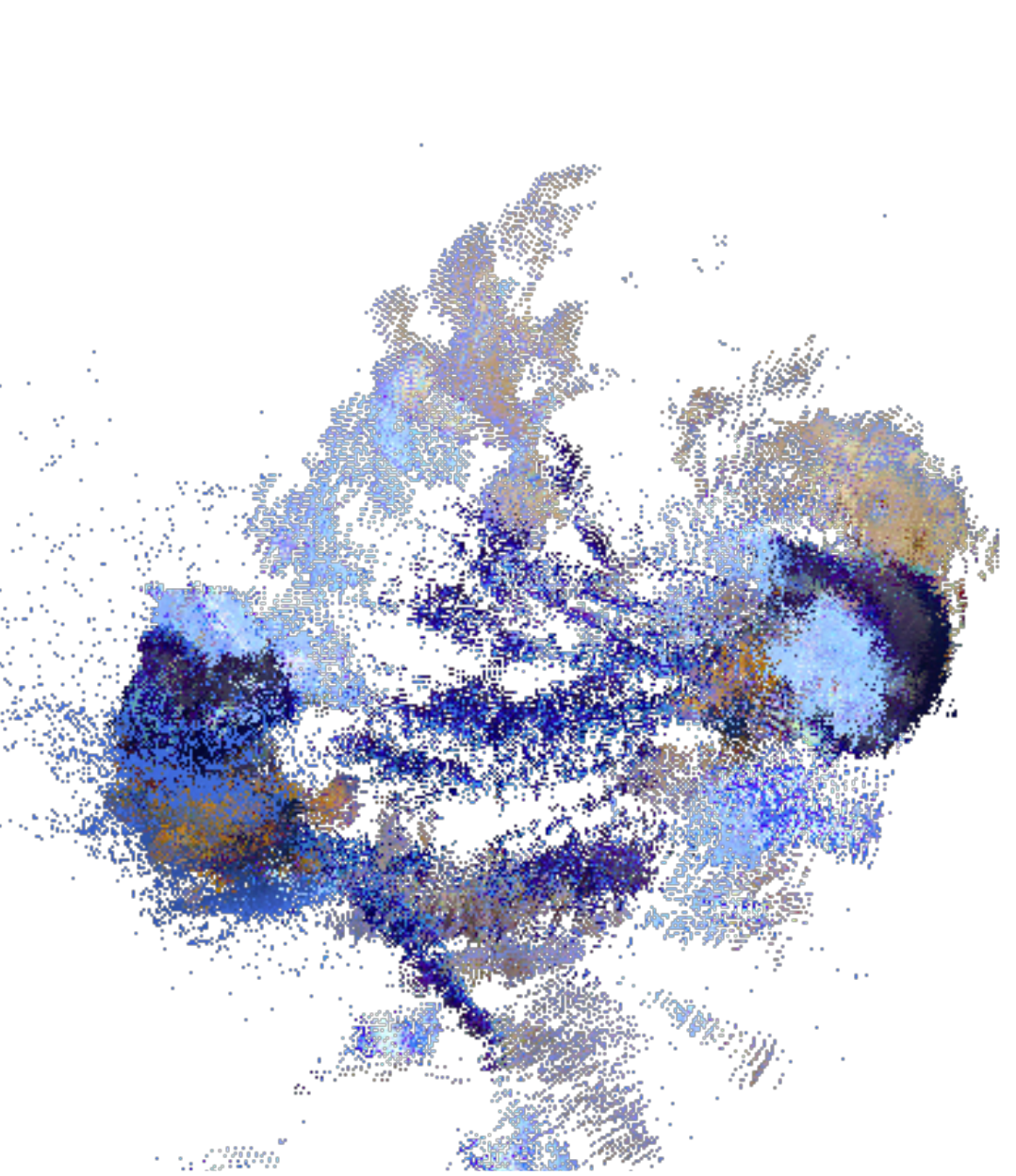}     
    \\  
    \raisebox{2cm}{b)} &  
    \includegraphics[width=0.2\textwidth]{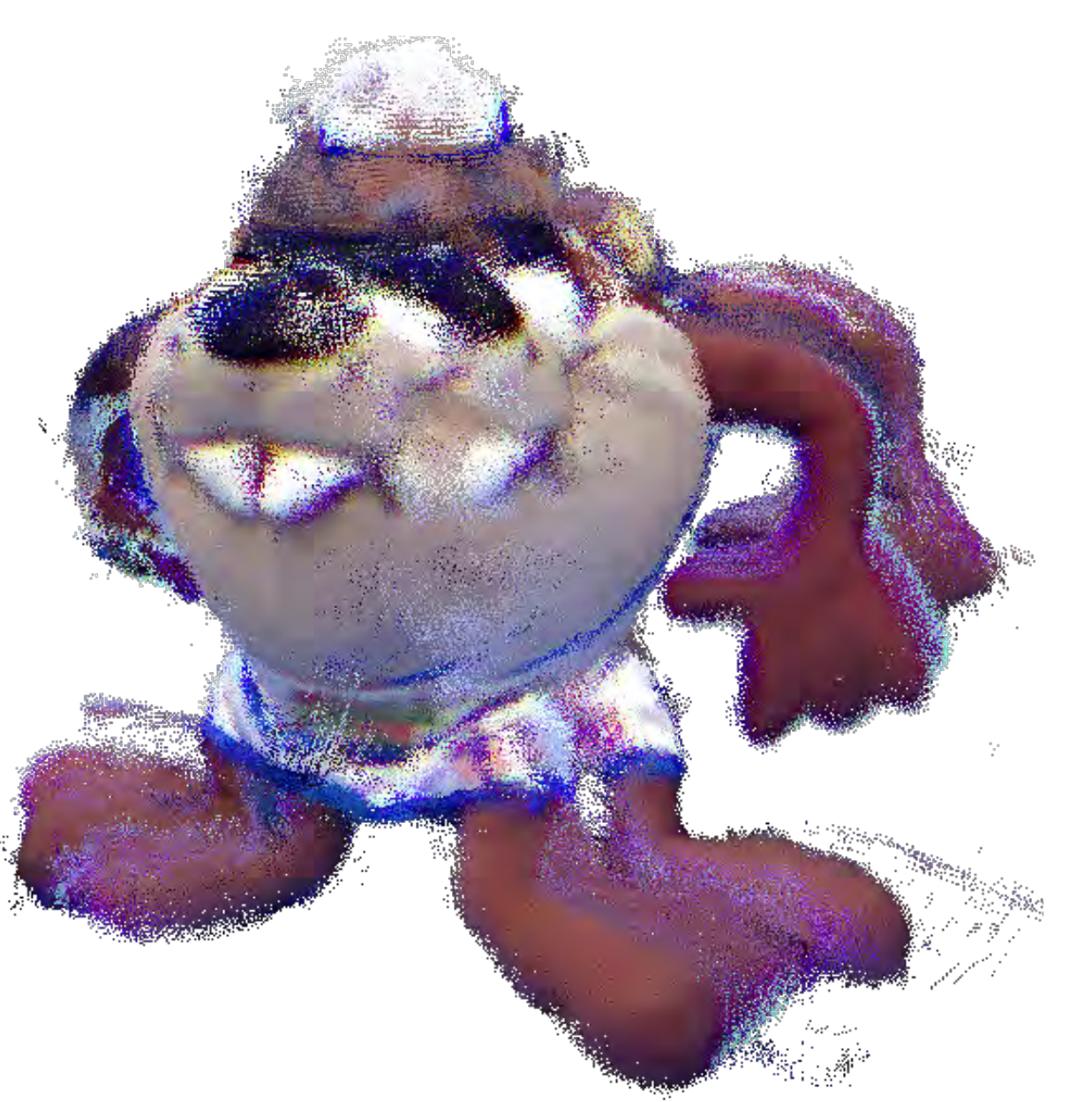} & \includegraphics[width=0.2\textwidth]{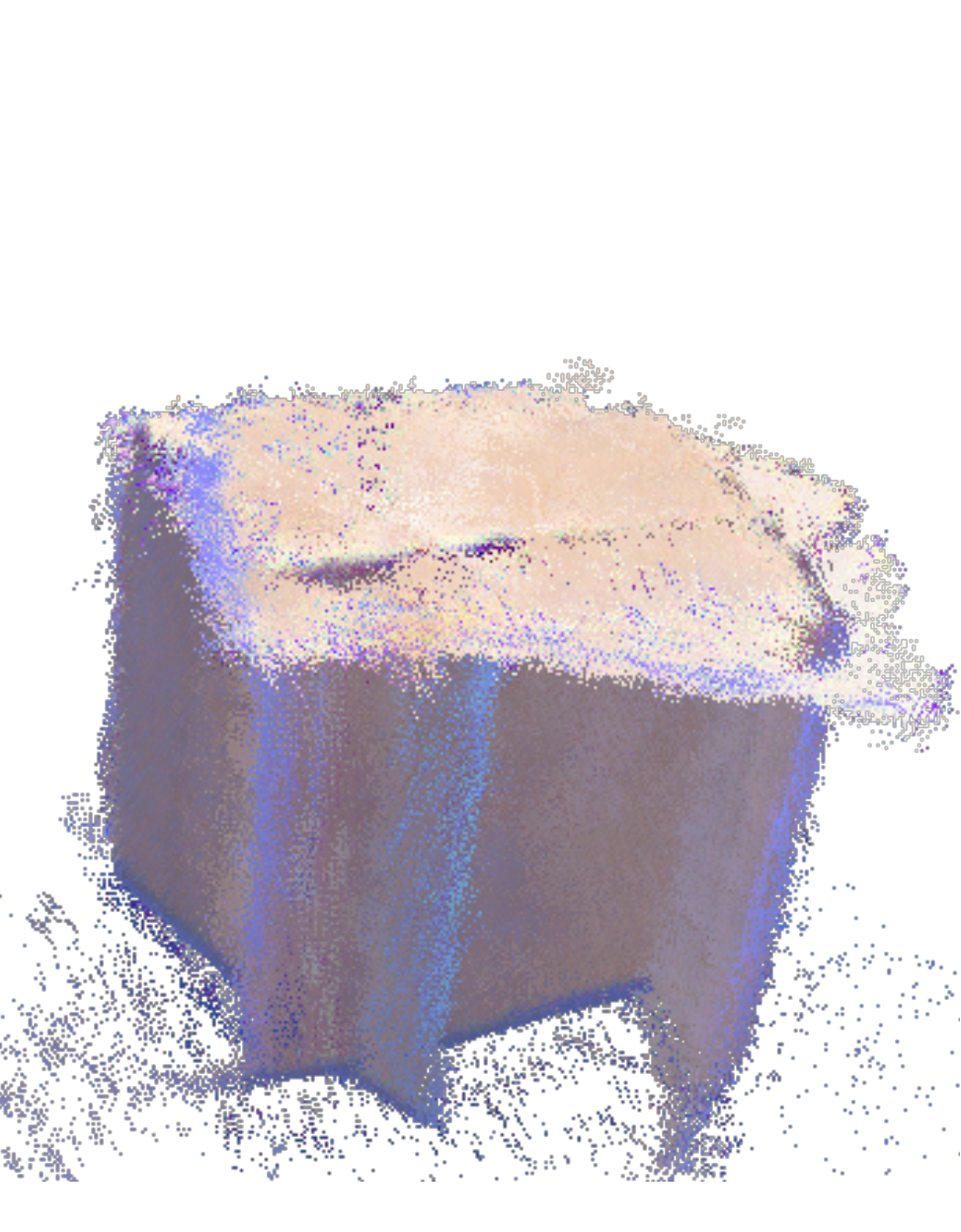} & \includegraphics[width=0.2\textwidth]{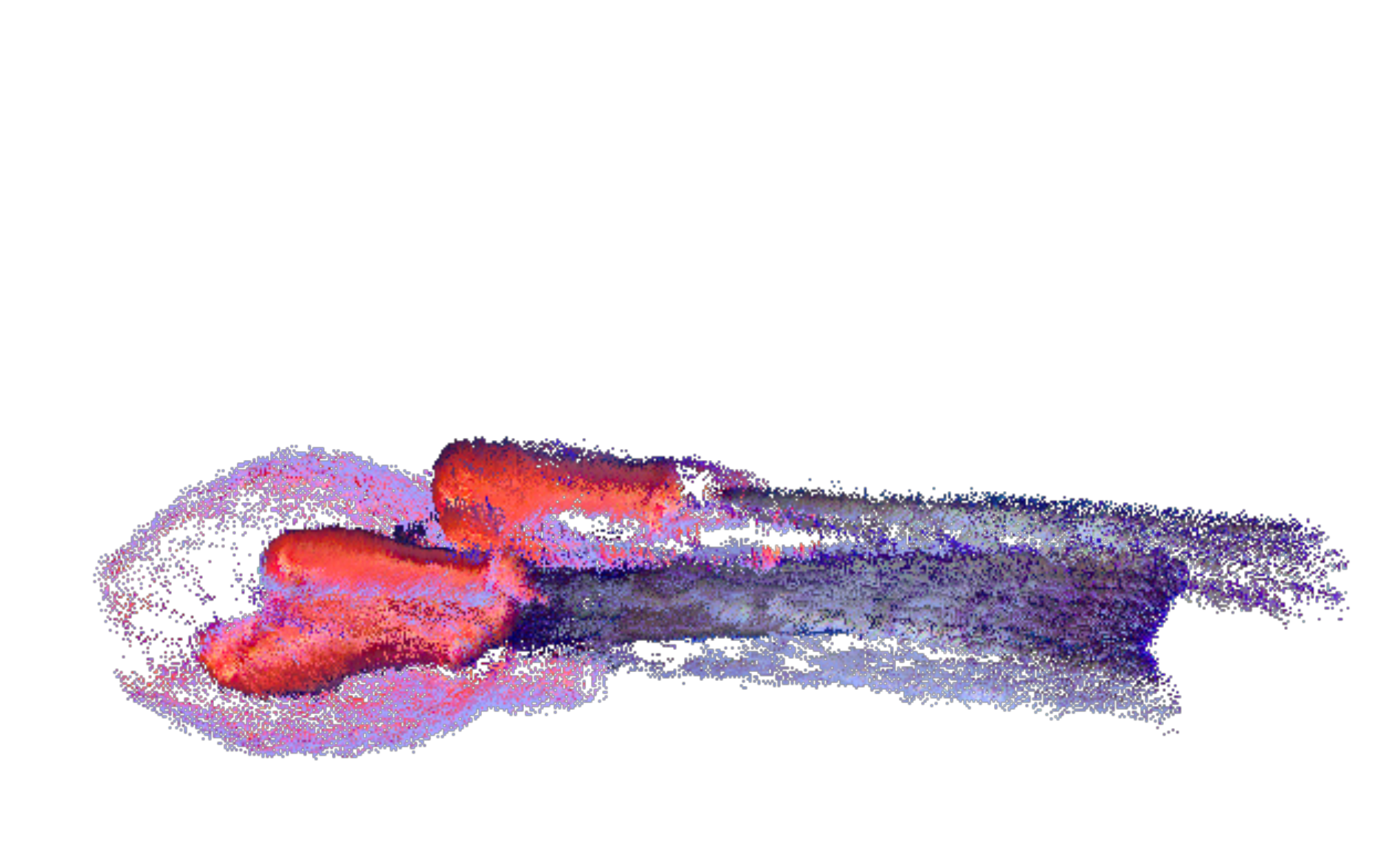} & \includegraphics[width=0.2\textwidth]{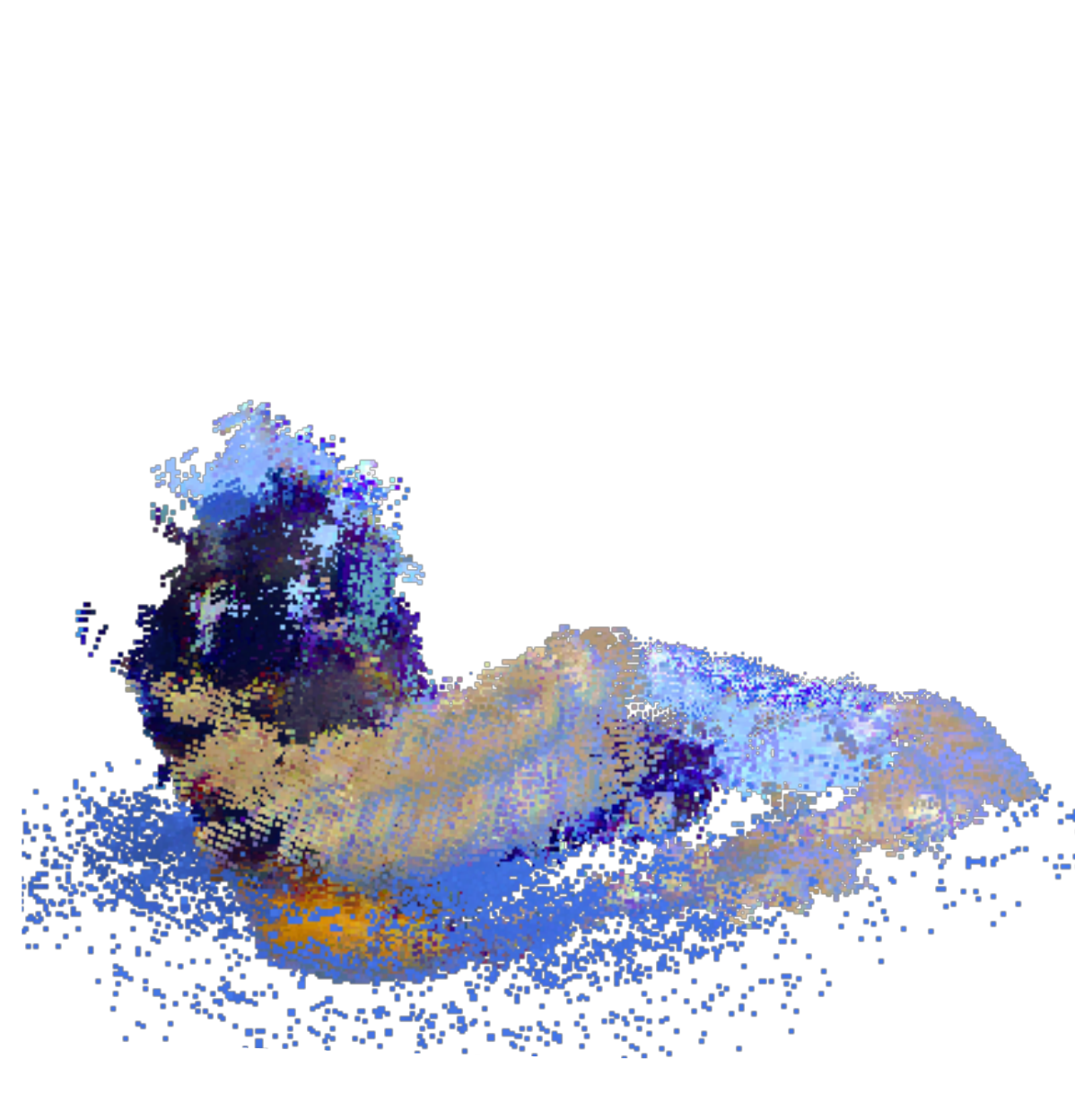}     
    \\   
   \raisebox{2cm}{c)} & 
    \includegraphics[width=0.2\textwidth]{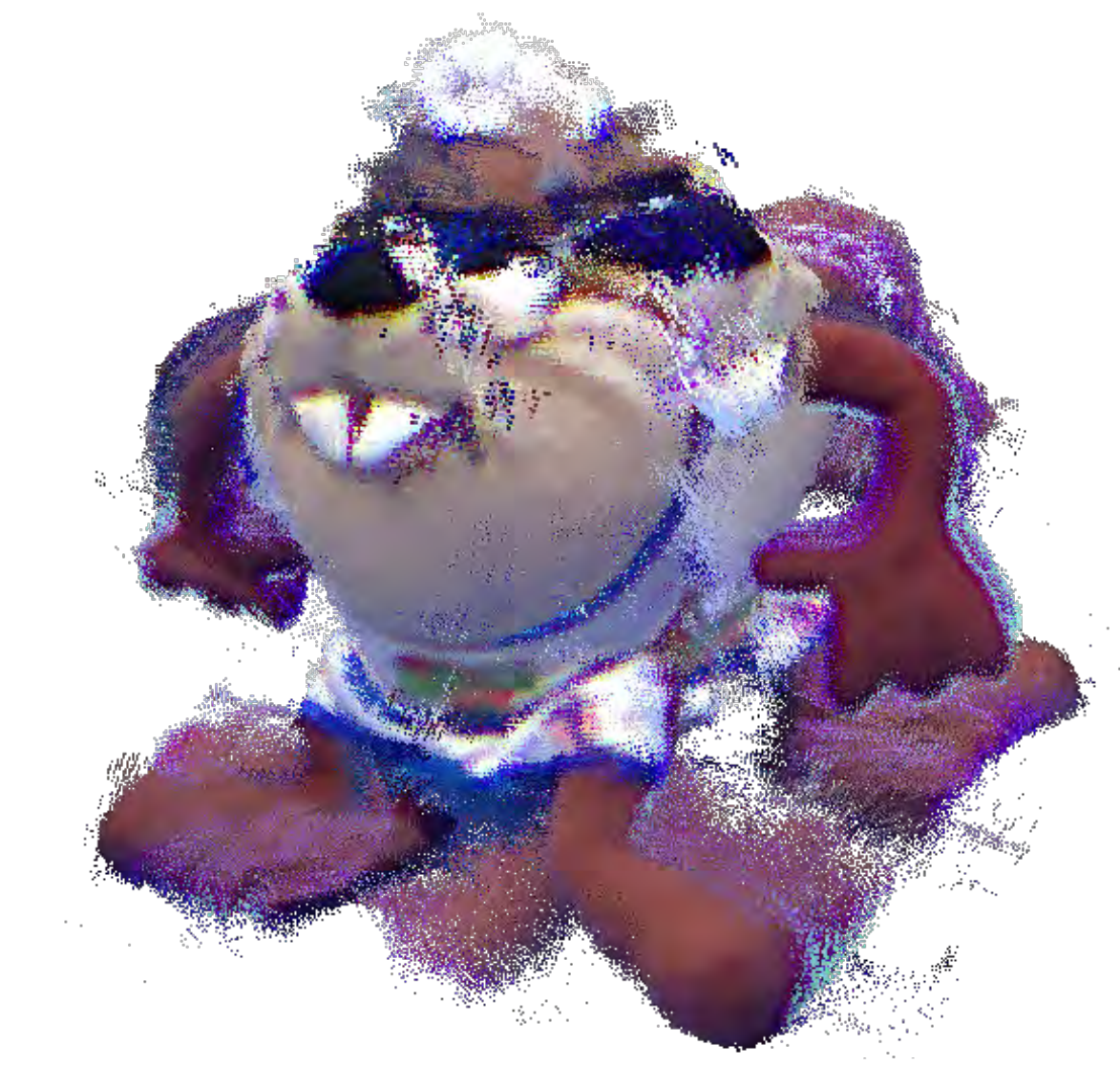} & \includegraphics[width=0.2\textwidth]{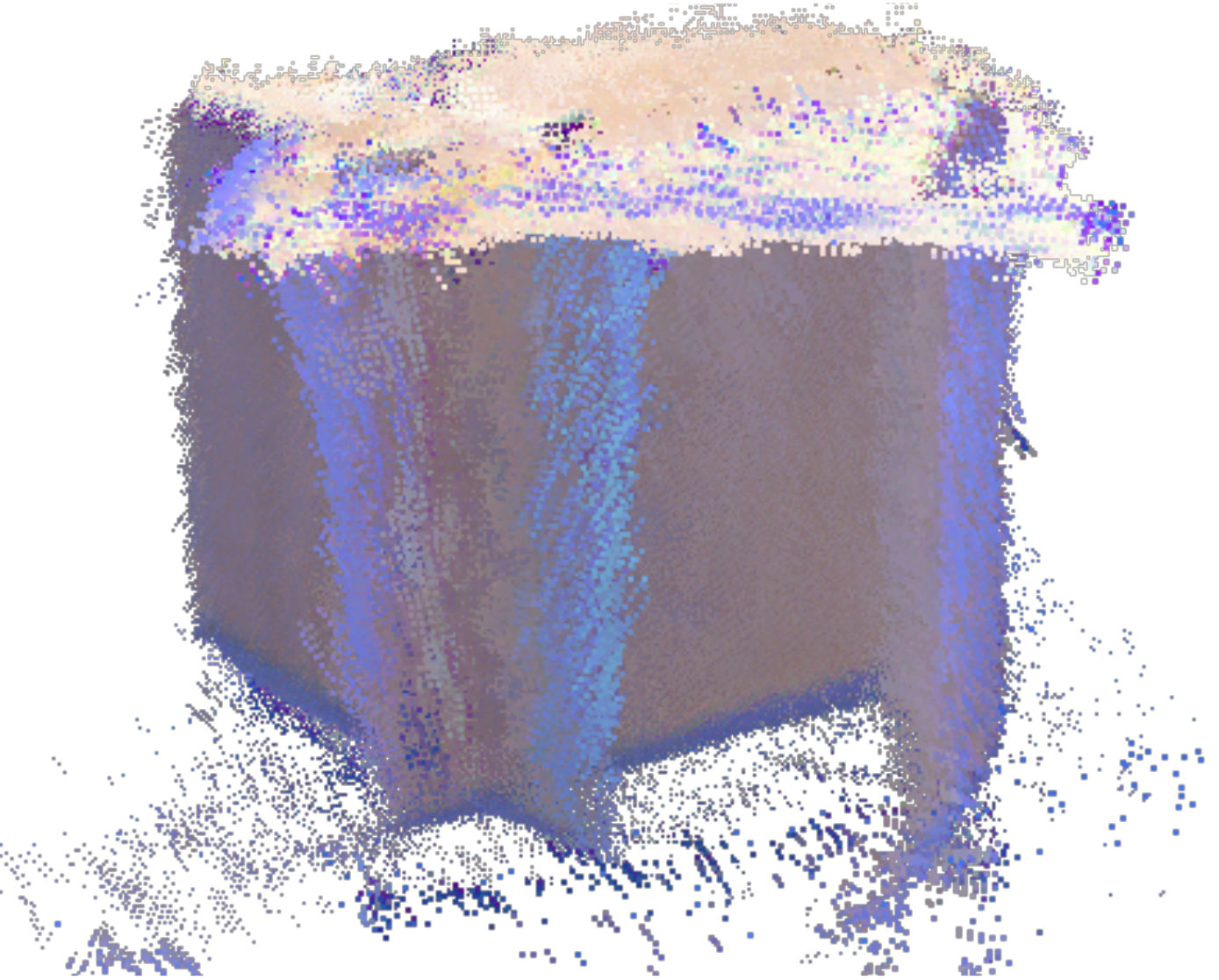} & \includegraphics[width=0.2\textwidth]{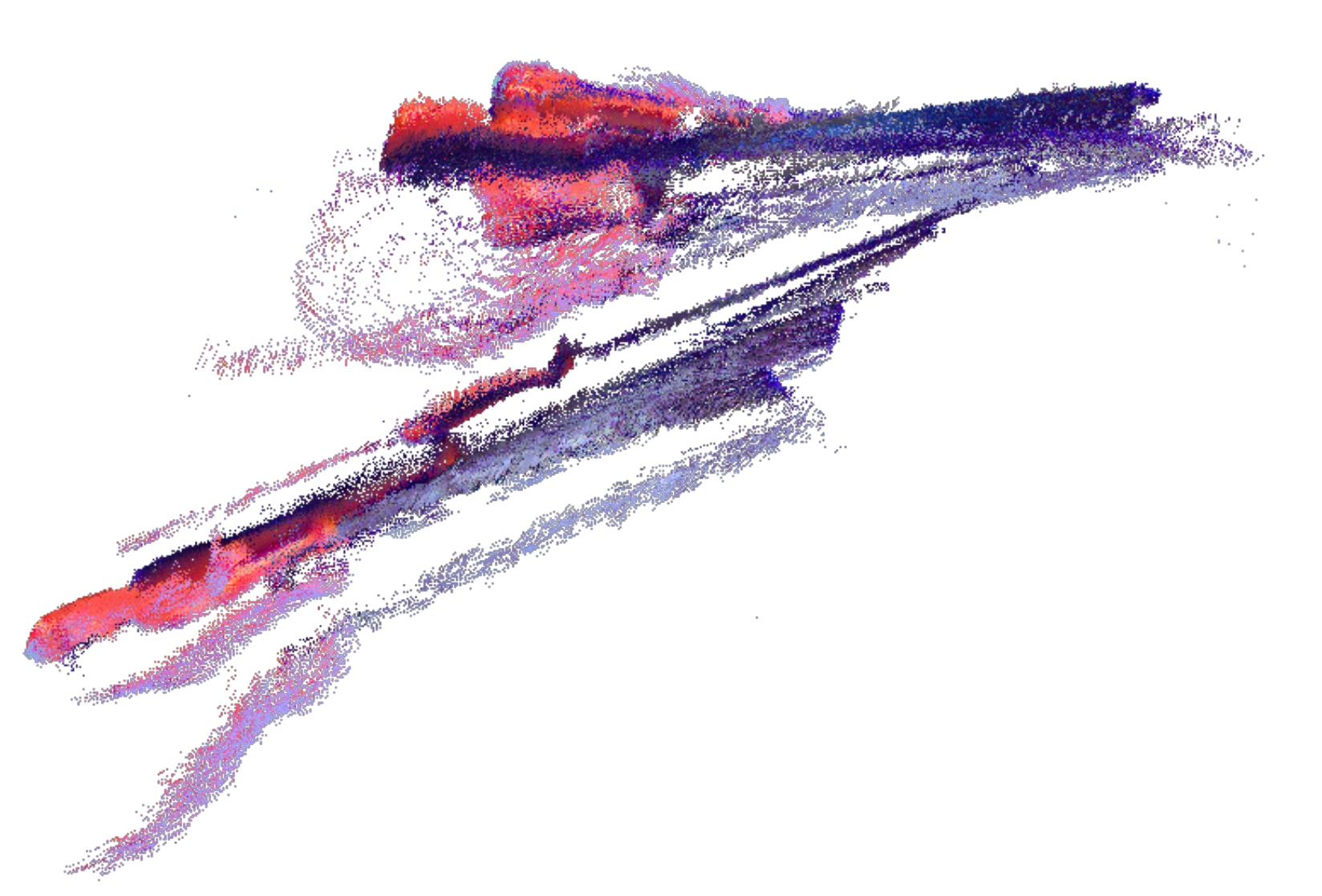} & \includegraphics[width=0.2\textwidth]{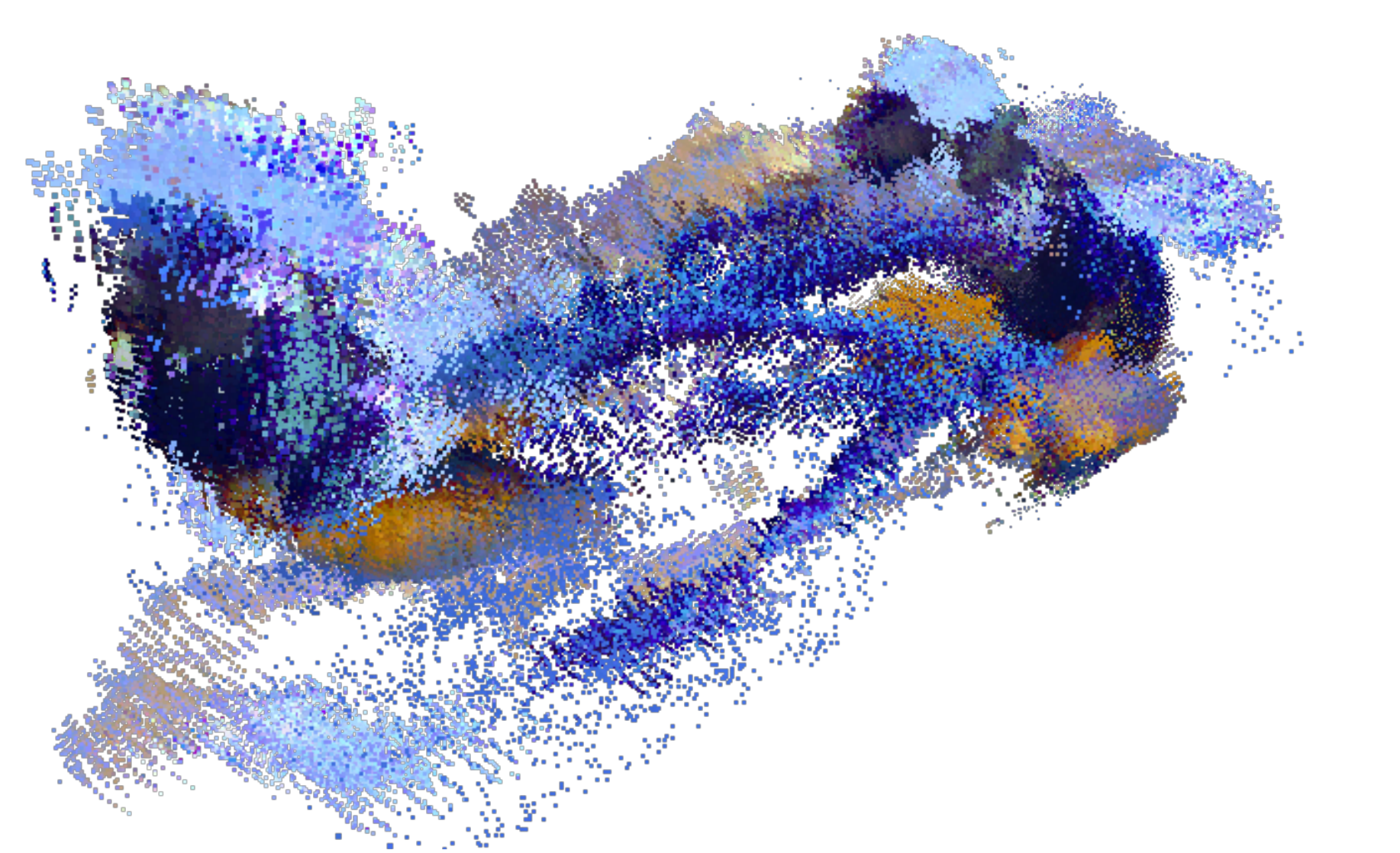}     
    \\ 
    \raisebox{2cm}{d)} &   
    \includegraphics[width=0.2\textwidth]{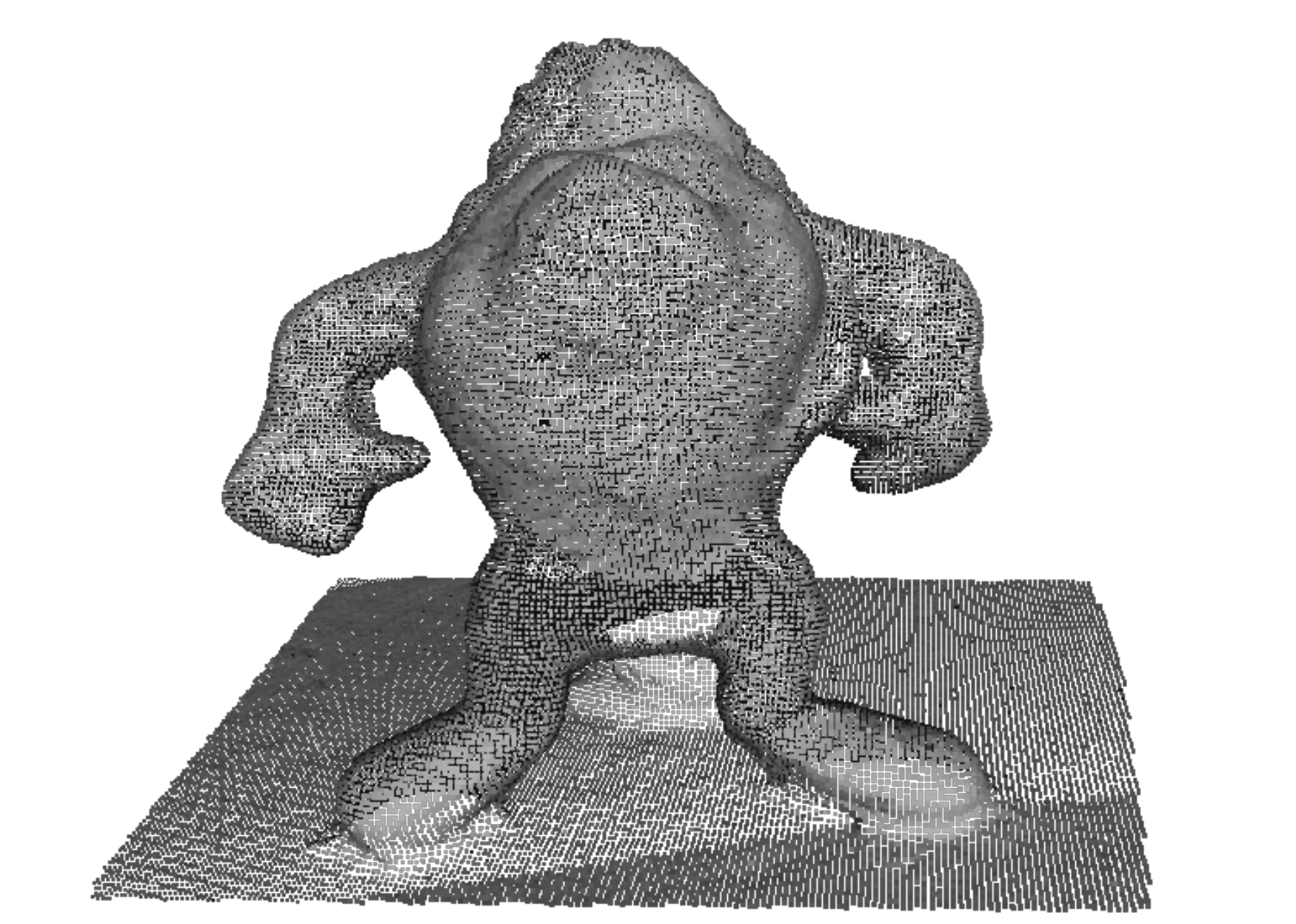} & \includegraphics[width=0.2\textwidth]{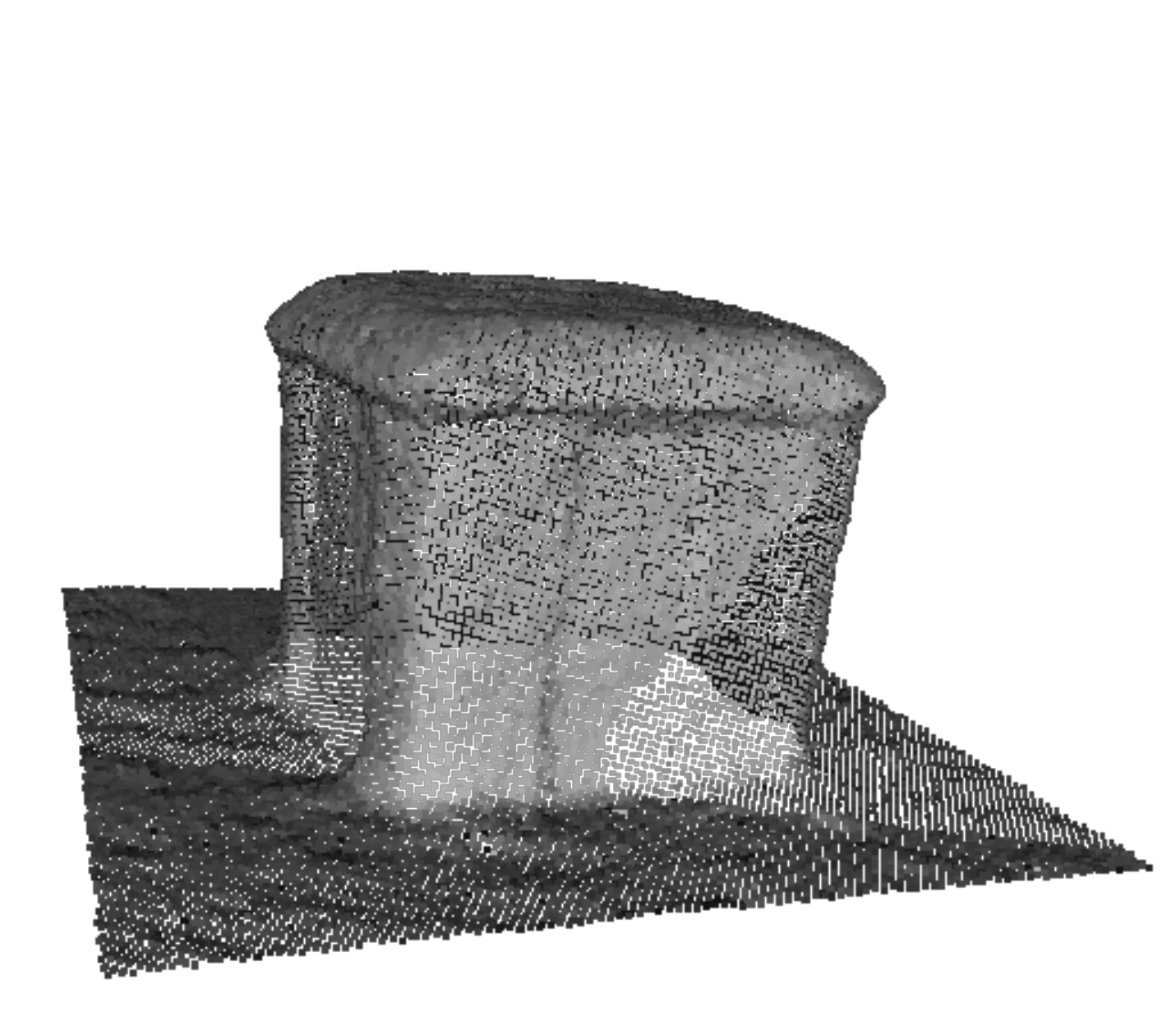} & \includegraphics[width=0.2\textwidth]{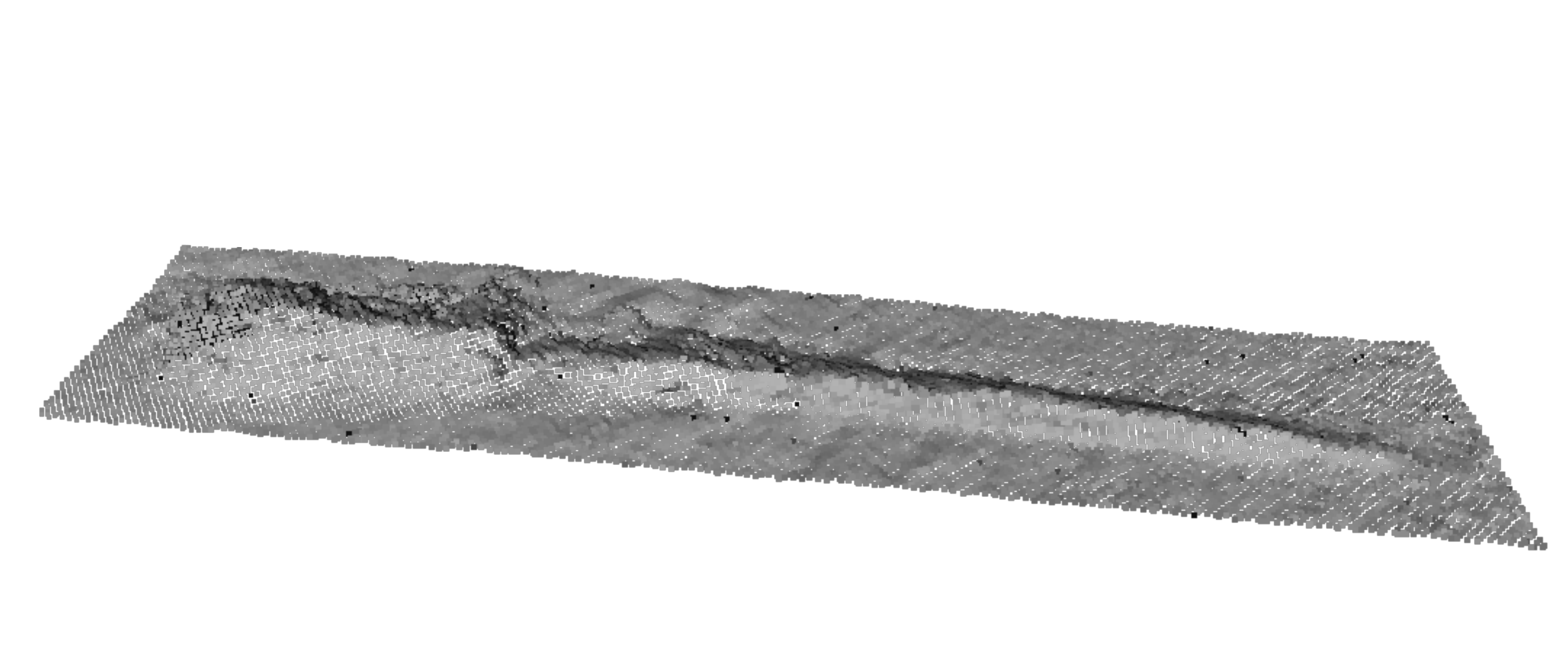} & \includegraphics[width=0.2\textwidth]{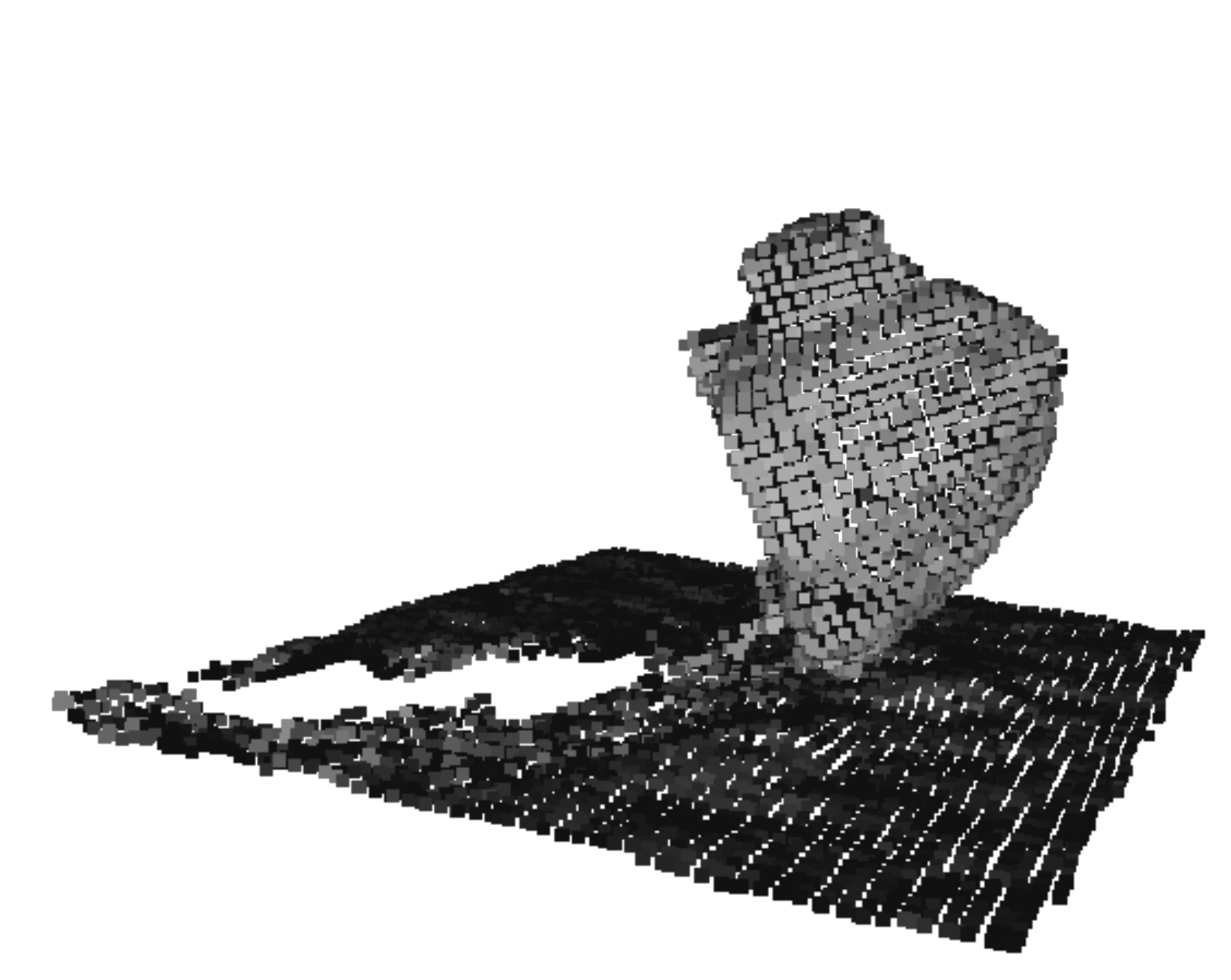} 
    \\
    \raisebox{2cm}{e)} &
    \includegraphics[width=0.2\textwidth]{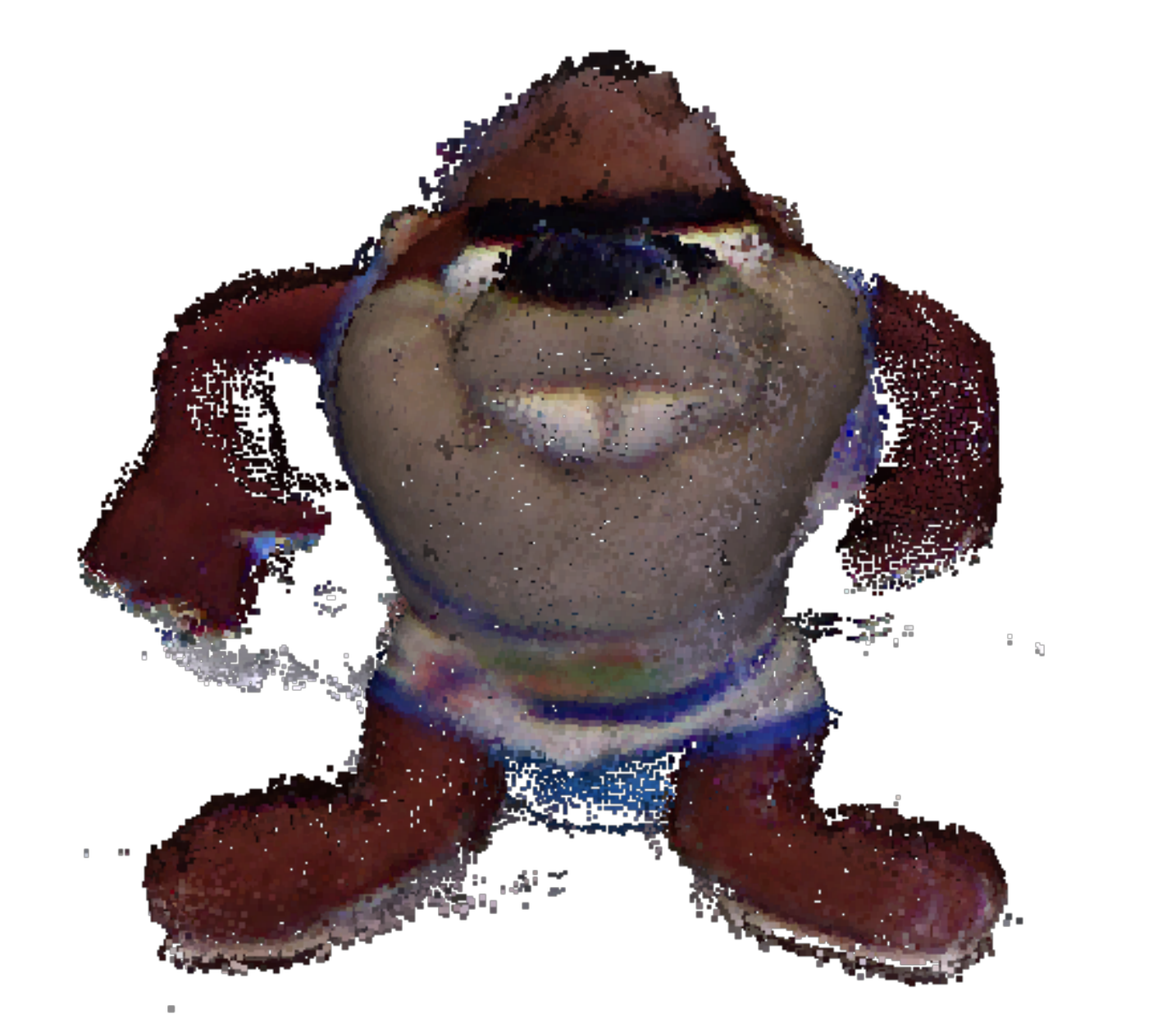} & \includegraphics[width=0.2\textwidth]{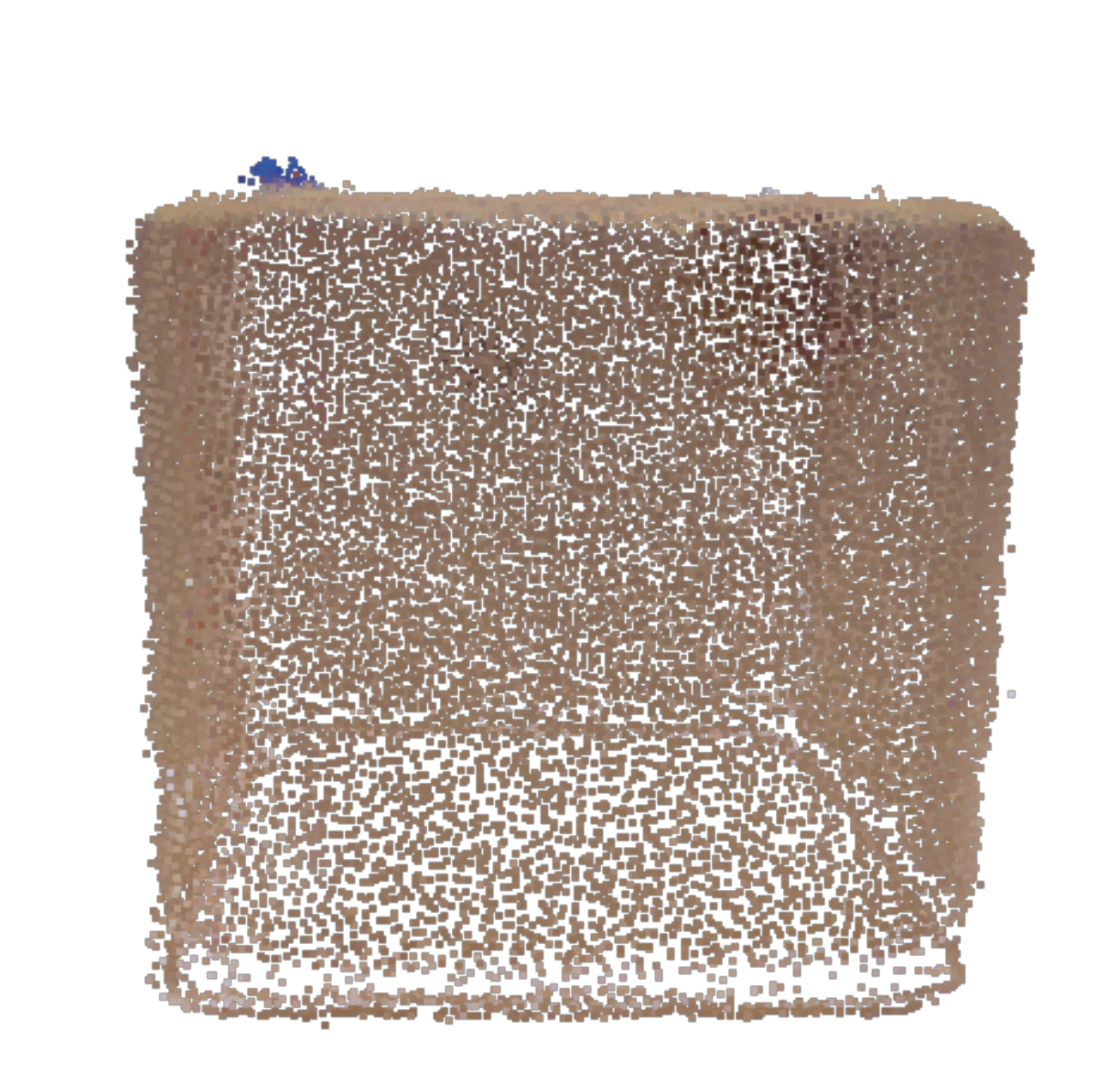} & \includegraphics[width=0.2\textwidth]{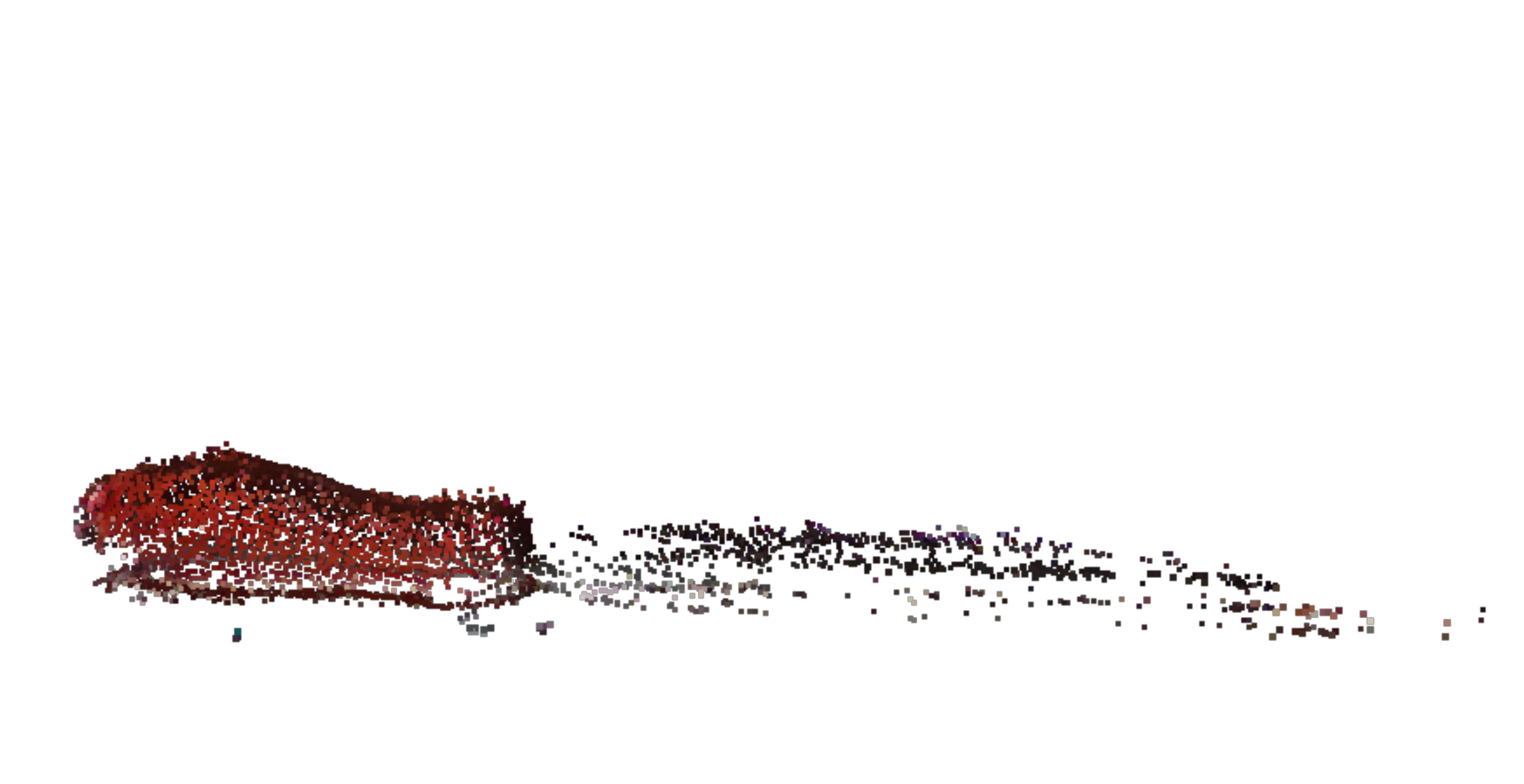} & \includegraphics[width=0.2\textwidth]{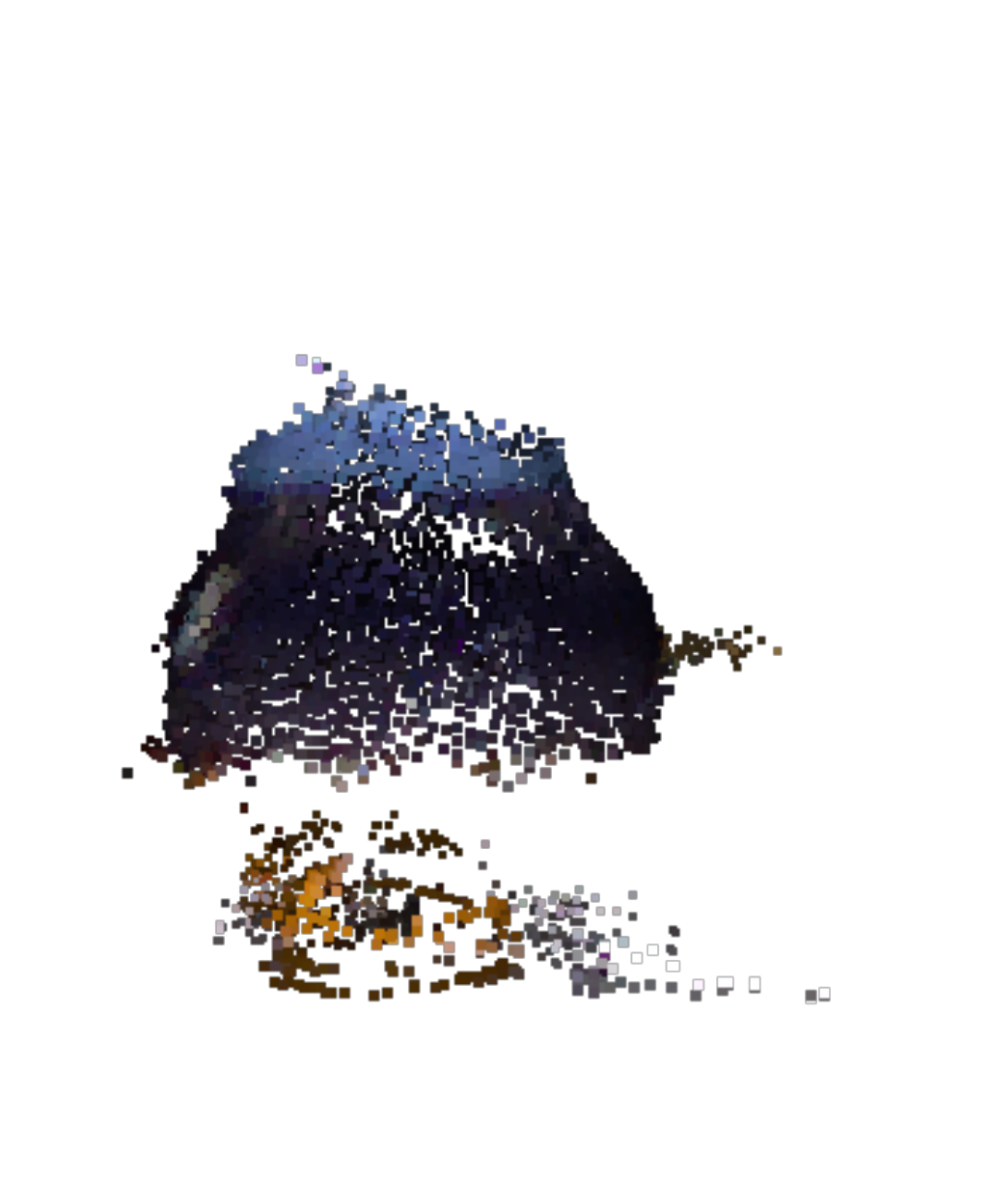}    
    \\
    \raisebox{2cm}{f)} &
    \includegraphics[width=0.2\textwidth]{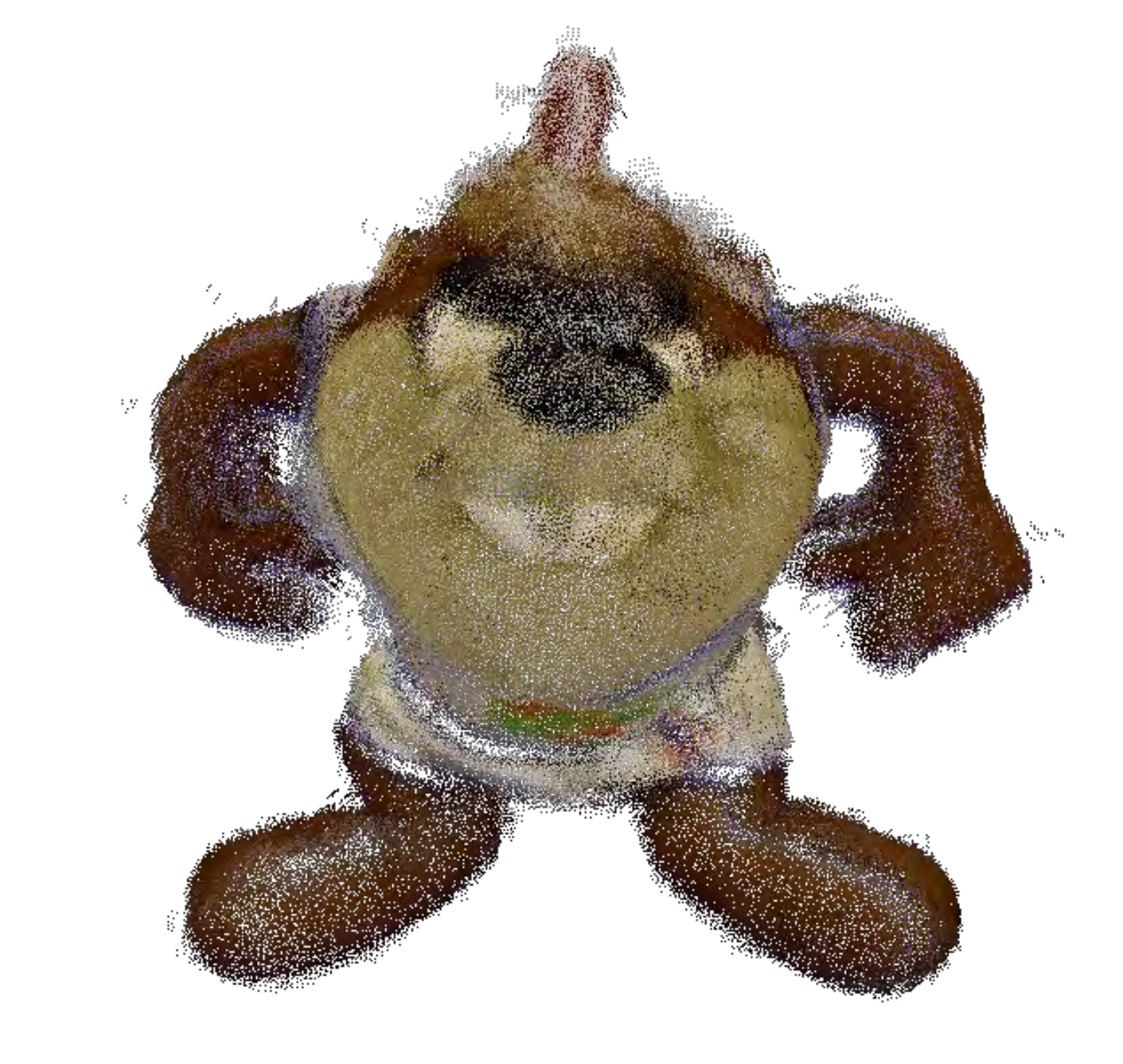} & \includegraphics[width=0.2\textwidth]{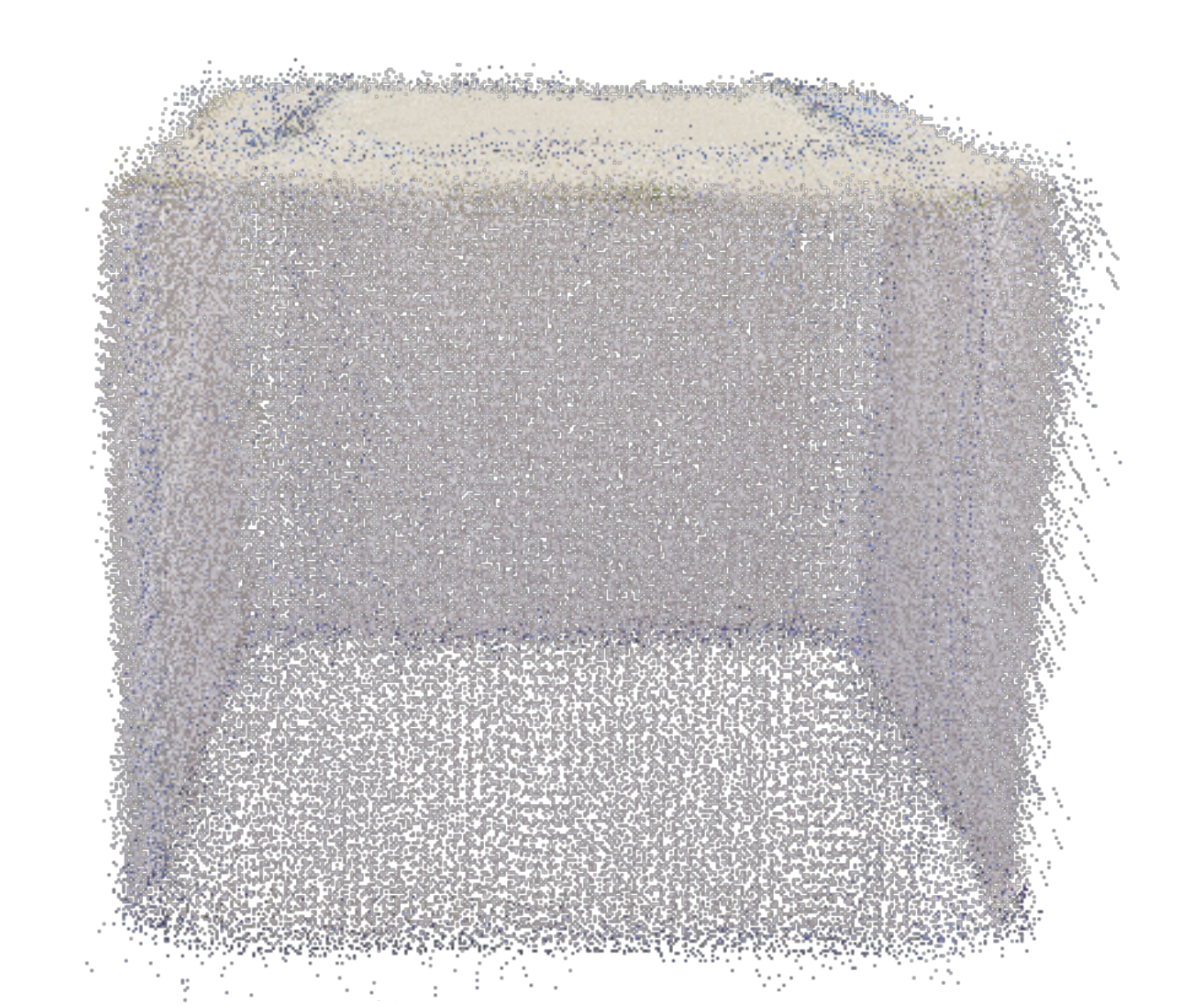} & \includegraphics[width=0.2\textwidth]{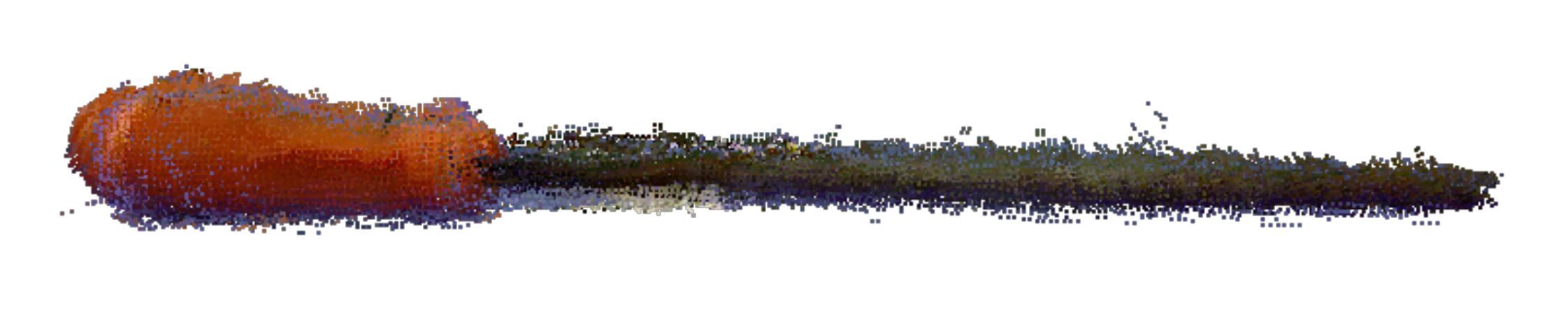} & \includegraphics[width=0.2\textwidth]{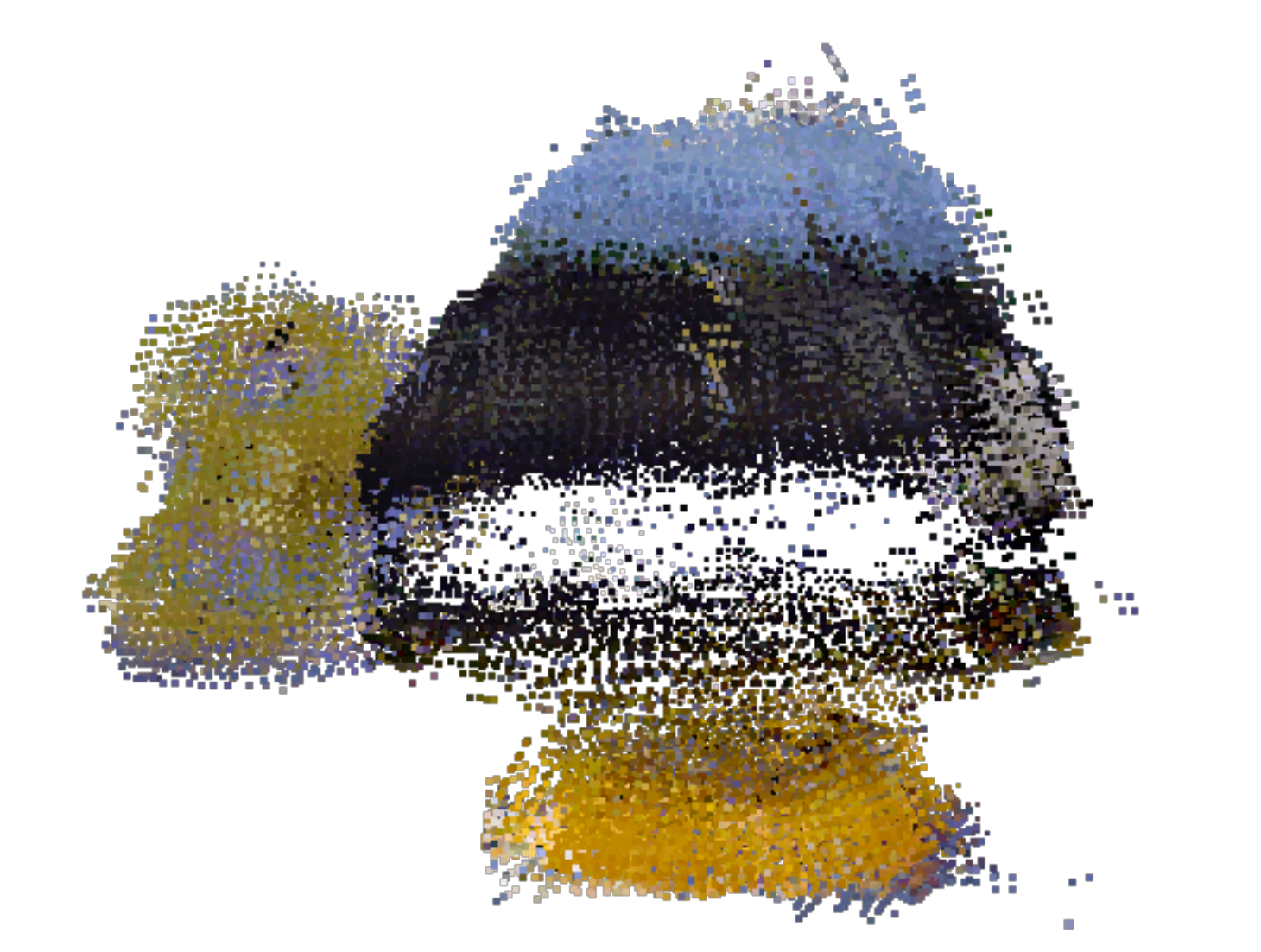}
    \end{tabular}}
   
    \caption{Objects registration results of the tested methods. Row a) presents RANSAC with SIFT features registration. Second row ICP registration.Third one shows the result of RANSAC and ICP combination. In row d) presents the results of KinectFusion. RGBDemo results are shown in the e) row. Finally, the last row (f)) shows the results of the proposed method.}
     \label{fig:objects_res}
\end{figure}

Figure~\ref{fig:objects_res} shows the result of the different registration algorithms. The row a) has the RANSAC-based registration results. From this first row it is concluded that Taz has been well registered in the front part due to the texture, but the cumulative error produces a wrong final registration. The rest of the objects have less texture being harder to find matches. Hence an inaccurate registration is resulted. Figure~\ref{fig:objects_res} b) presents the ICP results. The results are better than in a). However, in the Cube it is possible to appreciate how when two planes are visible the registration drifts to a wrong result. In c), a combination of RANSAC-based as pre-alignment and ICP for fine registration is shown. The results are dependant on both error produced in a) and b). Figure~\ref{fig:objects_res} d) presents KinectFusion registration. This method only uses geometry (i.e. ICP for registration, and smoothing techniques). The results are very rounded and the details disappear. Row e) presents the RGBDemo results. This method has been initially developed for RGB-D cameras and uses markers for coarse registration. The results are better than in previous presented methods. However in the Tool and the Bomb, due to the large down-sampling used for the ICP final refinement, many details are removed. 

The last row, Figure~\ref{fig:objects_res} f), shows the registration results of the $\mu$-MAR. The objects are accurately registered, outperforming the previous methods. Moreover, as no down-sampling is applied, all the points are registered. This is easily appreciated in the Tool (third column) compared to RGBDemo (row e) in the same object. The Bomb is also more complete having the feet of the toy. Since the Bomb is a specular object, it is necessary more data because of many points cannot be well extracted during the acquisition. Due to our method preserve all data points as it uses the external markers for the fine registration, a more dense point cloud is returned. The advantage of this, is that having all data allows a posterior down-sampling instead of an on-line reduction that is not possible to recover. 

\section{Discussion}\label{sec:reg:discussion}
Computer vision systems have to deal in many situations with noisy and low quality data because the resolution and sensitivity of the sensor is not appropriate for the problem to be solved. For example, the new low-cost RGB-D sensors are very interesting devices because the have a balanced sensitivity for a wide range of vision problems. For example, the sensitivity of the camera is enough to roughly reconstruct a human body in order to interact with a game
if the player is about 3 meters from the camera. However, at that distance, the sensitivity is not enough to detect millimetre features as for example face wrinkles. These sensors produce noise in the depth information (about millimetre orders) with a Signal-to-Noise ratio that difficult the perception of small shape features, which is critical in the registration process when a high accuracy is required in the application. The accuracy has to be understood here as the degree of closeness of the alignment provided by the registration process and the true alignment. An alignment will be more accurate when it offers a smaller distance to the real one.

In registration problems where several views have to be aligned with high accuracy into a common coordinate system, usually it is necessary to infer the transformations between views using an expert and intelligent systems approach. Most methods rely on colour or geometrical salient points, as features from the views, to estimate the correspondences among views in order to infer the transformations. However, the data provided by the sensors may not be reliable to match the necessary features, neither in colour nor in three-dimensional space. In the literature, it has been proposed specific methods for registering data from consumer RGB-D sensors, including RGBDemo \citep{Kramer2012,burrus2011} and KinectFusion \citep{Newcombe2011,Izadi2011}. The former uses colour 2D markers extracted from the colour camera (RGB), which depends on the two-dimensional image quality to estimate the shape (e.g. corners, edges). The point of view is another important drawback, since the perception of the markers could be difficult if the viewpoint is close to the plane in which the marker is printed. KinectFusion, on the other hand, uses a smoothing technique to estimate the incremental 3D reconstructed model and relies the kernel of registration in a traditional ICP method. These facts make the method not able to register objects with small details. The $\mu$-MAR method presented in this paper is able to reconstruct an object making use of the transformations calculated to align 3D markers in the scene using an expert system approach. The method is able to align multiple planes simultaneously which correspond to the faces of external 3D markers to apply them to the object to be reconstructed. The registration method uses prior knowledge about the marker to calculate the correspondences among views and infer the transformations with high accuracy. Since the method is focused on the marker registration but not on the object of interest to be reconstructed, the method is able to register any object irrespectively specific characteristics of it including colour, optical characteristics of the surface or shape. In order to use this method, it is only necessary to segment properly the 3D makers in the scene and provide the planar models of each of their faces. With these planes, the registration is easily performed from a computational point of view allowing future variants to improve the generality and implementation in embedded architectures to reduce time performance. This proposal is not only a contribution for the scientific community, but also in practical problems due to the affordability of these sensors. Many practical solutions may use them. In the case of reconstructing objects, the proposal is a solution when small or detailed objects are the aim of interest. Moreover, since this method is not only focused on RGB-D sensors, $\mu$-MAR could be used even when the sensor may not provide the data of the object with high accuracy for traditional methods. 

The use of low-cost RGB-D sensors is very interesting in many applications but for object reconstruction they remain certain problems. The resolution of these sensors is low to perceive the details of the object to be reconstructed. Therefore, despite the use of the registration method which uses external elements to estimate the transformations, the small details may not be accurately reconstructed due to the sensor does not provide the correct data. In this case, multiple views of the same part of the object have to be acquired in order to obtain as much information as possible to posteriorly fuse it. Moreover, after various experimental evaluations, it has been detected that the default calibration of these sensors do not provide correct data in the extremes of the image due to the lens distortion. Then, if a calibration steps could be perform, the registration accuracy improves. However, if a calibration is not possible to be applied, it is recommended to locate the camera and markers that the latter are not in the extreme of the images. Finally, the main drawback of our method is that the markers, at least one, have to be seen from the camera in order to reconstruct the object.

Experimental results show the performance of $\mu$-MAR to register objects in presence of noise. The synthetic data experiments validate the proposed methodology to provide an accurate alignment of a set of views. An objective evaluation using the Hausdorff distance shows high registration accuracy of the proposed method. Concretely, the Hausdorff distance error mean in average (using mean distances to calculate the Hausdorff distance) is 0.6556 for $\mu$-MAR and 6.1515 for ICP. In the case of RMS (using RMS to calculate the Hausdorff distance), the proposed method achieves an error of 0.8958 in average and 8.1154 for ICP, concluding that $\mu$-MAR is close to 10 times better in average than ICP. The real data experimentation validates the proposal in real situations using a general purpose RGB-D sensor data. The method is compared to state-of-the-art methods including ICP, RANSAC-based, KinectFusion and RGBDemo. The visual inspection evaluation shows how the ($\mu$-MAR) outperforms the rest of the tested methods.

\section{Conclusions}\label{sec:reg:conclusions}

In this paper, a novel MUltiplane 3D MArker based Registration method ($\mu$-MAR) is proposed. The proposal is able to transform a set of views in a common coordinate system minimizing the effects of noise (very common in data from RGB-D cameras). $\mu$-MAR method uses a multi-view registration variant for subsets of views using planar models as 3D markers. Planes are a geometric model presented almost in any scenario. The planes are simultaneously registered using the normals for the rotation estimation, and the centroids for the translation finding. The method is able to register the views without any previous coarse registration, providing a fine and high accurate alignment.


The two main contributions of this paper are: the use of 3D markers to perform a model-based registration of them avoiding the noise associated to general purpose cameras; and the iteratively variant of the multi-view registration using subsets to finely match same regions of the marker to increase accuracy. Since the method uses the same transformations to register the whole scene, the objects could be reconstructed with high accuracy.

As future works, now I am working in extend the proposal for general scenes where any object composed by planes could be present in the scene. For mapping applications where several planes are in the scene, commonly with a known angle geometry (usually 90 degrees between the walls and roof or floor). Moreover, a evaluation of the effects of mixing different shapes (e.g.: pyramids, cubes,...) as 3D markers is going to be carried out. It is very plausible that the use of a double pyramid could be the suitable marker to reconstruct the object from different view points, not only around the object but also upper and lower, improving the self-occlusions recovery. Finally, we plan to provide additional prior knowledge including expected accuracy of the reconstruction and complexity of the shape, to the registration method. The objective is the registration method could decide by itself the necessary viewpoints the system should capture to properly reconstruct the object according to the required accuracy of the problem to be solved.

%
\section*{Acknowledgments}
This work has been supported by grant University of Alicante projects GRE11-01 and grant Valencian Government GV/2013/005.
%
%
%
%
%


%

\section{References}
\bibliographystyle{elsarticle-harv}
\bibliography{references}

\end{document}